%% file: arxiv.tex
\documentclass[10pt,journal,compsoc]{IEEEtran}
\usepackage{color} 
\usepackage{xcolor}
\usepackage{xspace} 
\usepackage{ifthen} 

\usepackage{floatrow} 
\usepackage{booktabs} 
\usepackage{multirow} 

\ifCLASSOPTIONcompsoc
  \usepackage[nocompress]{cite}
\else
  \usepackage{cite}
\fi
\ifCLASSINFOpdf
  \usepackage[pdftex]{graphicx}
\else
\fi
\usepackage{subcaption}
\usepackage{adjustbox}

\usepackage{amsmath}
\usepackage{amssymb}
\usepackage{amsfonts}
\usepackage{dsfont}
\usepackage{mathtools}
\definecolor{applegreen}{rgb}{0.55, 0.71, 0.0}
\newboolean{revising}
\setboolean{revising}{true}
\ifthenelse{\boolean{revising}}
{

\newcommand{\q}[1]{\textcolor{red}{Question: #1}}
\newcommand{\todo}[1]{\textcolor{red}{TODO: #1}}
}{

\newcommand{\q}[1]{}
\newcommand{\todo}[1]{}
}

\makeatletter
\DeclareRobustCommand\onedot{\futurelet\@let@token\@onedot}
\def\@onedot{\ifx\@let@token.\else.\null\fi\xspace}

\def\eg{\emph{e.g}\onedot} 
\def\ie{\emph{i.e}\onedot}

\def\etal{\emph{et al}\onedot}
\makeatother

\newcommand\subfix[1]{\mathtt{#1}}

\newcommand\minisection[1]{\vspace{1mm}\noindent \textbf{#1}}
\newcommand{\figcaption}[2]{\caption{\textbf{#1} #2}}

\newcommand{\tabref}{Table~\ref}
\newcommand{\figref}{Figure~\ref}
\newcommand{\secref}{Section~\ref}
\newcommand{\eqnref}{Equation~\ref}

\begin{document}
%
\title{Monocular Quasi-Dense 3D Object Tracking}
%
%
%
%

\author{
  Hou-Ning~Hu, 
  Yung-Hsu~Yang, 
  Tobias Fischer, 
  Trevor Darrell, 
  Fisher~Yu, 
  and~Min~Sun
  \IEEEcompsocitemizethanks{
    \IEEEcompsocthanksitem H.-N. Hu, Y.-H. Yang and M. Sun are with the Department
    of Electrical Engineering, National Tsing Hua University, Hsinchu, Taiwan.\protect\\
    E-mail: eborboihuc@gapp.nthu.edu.tw
    \IEEEcompsocthanksitem T. Darrell is with the Department of EECS Department, UC Berkeley.
    \IEEEcompsocthanksitem T. Fischer and F. Yu are with the Department of Information Technology and Electrical Engineering, ETH Zurich.}
}

%
%

\markboth{}%
{}
%



\IEEEtitleabstractindextext{%
  \begin{abstract}
    
    A reliable and accurate 3D tracking framework is essential for predicting future locations of surrounding objects and planning the observer's actions in numerous applications such as autonomous driving. We propose a framework that can effectively associate moving objects over time and estimate their full 3D bounding box information from a sequence of 2D images captured on a moving platform. The object association leverages quasi-dense similarity learning to identify objects in various poses and viewpoints with appearance cues only. After initial 2D association, we further utilize 3D bounding boxes depth-ordering heuristics for robust instance association and motion-based 3D trajectory prediction for re-identification of occluded vehicles. In the end, an LSTM-based object velocity learning module aggregates the long-term trajectory information for more accurate motion extrapolation. Experiments on our proposed simulation data and real-world benchmarks, including KITTI, nuScenes, and Waymo datasets, show that our tracking framework offers robust object association and tracking on urban-driving scenarios. On the Waymo Open benchmark, we establish the first camera-only baseline in the 3D tracking and 3D detection challenges. Our quasi-dense 3D tracking pipeline achieves impressive improvements on the nuScenes 3D tracking benchmark with near five times tracking accuracy of the best vision-only submission among all published methods. Our code, data and trained models are available at https://github.com/SysCV/qd-3dt.
  \end{abstract}

   \begin{IEEEkeywords}
   Monocular 3D Detection, Monocular 3D Tracking, Multiple Object Tracking, Quasi-Dense Similarity Learning
   \end{IEEEkeywords}
}

\maketitle

\IEEEdisplaynontitleabstractindextext

%
\IEEEpeerreviewmaketitle

\input{01Intro}

\input{02RelatedWorks}
\input{03Technical}
\input{04Dataset}
\input{05Experiments}

\input{06Conclusion}

\input{08Acknowledgements}
\clearpage
\input{07Appendix}
\clearpage
\input{09References}

\end{document}

%% file: 01Intro.tex
\IEEEraisesectionheading{\section{Introduction}\label{sec:introduction}}

%
%
%
%
\IEEEPARstart{A}{}utonomous driving motivates much of contemporary visual deep learning research.
However, many commercially successful approaches to autonomous driving control rely on a wide array of views and sensors, reconstructing 3D point clouds of the surroundings before inferring object instance trajectories in 3D.
In contrast, human observers have no difficulty perceiving the 3D world in space and time from simple sequences of 2D images rather than 3D point clouds, even though human stereo vision only reaches several meters.
Recent progress in monocular object detection and scene segmentation offers the promise to make low-cost mobility widely available.
In this paper, we focus on monocular 3D detection and tracking so that this low-cost system can be robust and reason in 3D even without 3D sensors.

\begin{figure}[t]
	\begin{center}
		\includegraphics[width=1.0\linewidth]{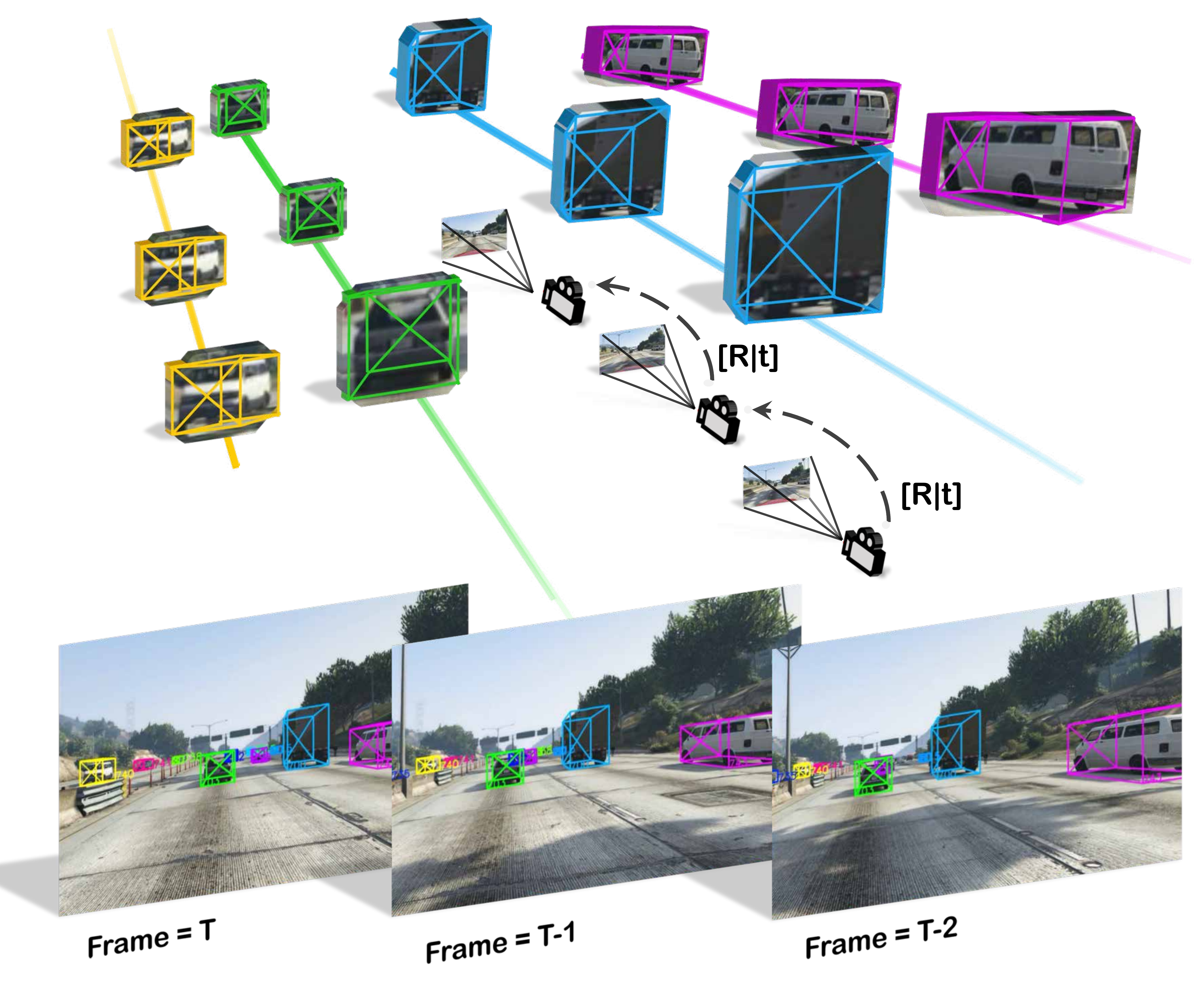}
	\end{center}
	\figcaption{Monocular quasi-dense detection and tracking in 3D.}{
		Our dynamic 3D tracking pipeline predicts 3D bounding box association of observed target from quasi-dense object proposals in image sequences captured by a monocular camera with an ego-motion sensor.}
	\label{fig:teaser}
\end{figure}

Monocular 3D detection and tracking are inherently intertwined. 3D detection is challenging by itself in the absence of depth measurements or strong priors given a single image. However, current 3D object detectors based on deep learning show promise in capturing geometric priors. This detection basis makes short-term 3D tracking more robust and makes long-term tracking possible. At the same time, 3D tracking information across multiple frames can assist 3D detection in future frames. Note that 3D tracking in the world coordinate can be achieved by projecting object 3D extents in each frame to the world coordinate so that a still object stays still although it moves in 2D observations. This 3D transformation makes both 3D detection and tracking robust as ego-motion of the sensor platform is factored out.


We propose an online 3D tracking framework to detect and track 3D objects in world coordinates from a series of monocular images. \figref{fig:teaser} provides an overview of our 3D detection and tracking task. After generating object proposals from a monocular image, we estimate their 3D properties, \ie, 3D bounding box center, depth, dimension, and orientation with a multi-head network. Our model also learns an instance descriptor represented as a vector in a latent space via quasi-dense similarity learning. The quasi-dense mechanism leverages densely populated object proposals, instead of scarce ground truth bounding boxes, to learn instance similarity mappings on a high-dimensional latent space and decide if two object detections are the same object. We utilize the instance feature embedding, estimated 3D information, and object motion from lifted 3D object instance poses to associate instances over time. Notably, we leverage novel motion-aware association and depth-ordering matching algorithms to overcome the occlusion and reappearance problems in tracking. Finally, our model captures the movement of instances in a world coordinate system and updates their 3D poses using velocity-based LSTM (VeloLSTM) to estimate motion along a trajectory, integrating single-frame observations associated with an object instance over time.

Like any deep network, our model is data-hungry.
The more data we feed it, the better it performs.
However, early public datasets are either limited to static scenes~\cite{wen2015ua}, lack the required ground truth trajectories~\cite{milan2016mot16}, or are too small to train contemporary deep models~\cite{geiger2012kitti}.
To bridge this gap, we resort to realistic video games.
We use a new pipeline to collect large-scale 3D trajectories from a realistic synthetic driving environment, augmented with dynamic meta-data associated with each observed scene and object instance.
Thanks to the hype of autonomous driving, more and more large-scale datasets, and benchmarks have surged onto the desk.
However, most of the datasets, collected in days to months and with costly human resources to reach a reasonable scale for a data-hungry network, are licensed for only non-commercial use.
Our 3D vehicle tracking simulation dataset democratizes the autonomous driving field that our simulation dataset is open to both the research community and industrial side. As a result, we believe the presented synthetic driving dataset would foster the community in discovering the power of a deep network as an entry point, especially in combination with sim-to-real approaches, for the large dataset and benchmarks.


In summary, we propose a new pipeline while revising each module building upon the findings of our initial work~\cite{Hu2019Mono3DT}. The key contributions and extended results are highlighted below.

\minisection{Quasi-Dense 3D Object Tracking Pipeline.}
We integrate Quasi-Dense Similarity Learning~\cite{pang2020quasidense} into our tracking framework. It learns to match objects from dense object proposals, instead of only a few annotated foreground regions.
Our quasi-dense 3D tracking pipeline outperforms the camera-based state of the art by near $500\%$ while bridging the gap to LiDAR-based methods on the nuScenes 3D tracking benchmark.
Please refer to~\secref{sec:quasi_dense_similarity_feature} for the method details.

\minisection{3D Confidence, Motion-based Similarity and VeloLSTM.}
In addition to the 2D tracking upgrade, we revised our detection and tracking method from all aspects, \ie, detector, tracker, and motion model.
We introduce a 3D confidence in~\secref{sec:3d_estimation}, aggregating 2D bounding box confidence and depth confidence, that benefits the bounding box filtering. 
In~\secref{sec:3d_tracking}, we improve our centroid-based data association scheme with a motion-based similarity so that the tracker is encouraged to match reappeared candidates in trajectories. 
Lastly, in~\secref{sec:lstm_motion}, we enhance our VeloLSTM module to model not only object velocity, but heading angle and dimension. 
With all the new techniques, our 3D tracking method has improved the rank on the KITTI 2D tracking benchmarks outperforming existing online approaches.

\minisection{Detailed Experiments on large-scale Datasets.}
Apart from the KITTI and our simulated dataset, we applied our extended model to experiment on urban-driving situations with recently collected large-scale datasets, \ie, nuScenes and Waymo Open.
On nuScenes, we focus on detailed module design comparison and hyper-parameter tuning, while on Waymo Open, we discuss the advantages and limitations of a monocular-based method.

%% file: 02RelatedWorks.tex
\section{Related Works}

\minisection{2D Object Detection.} 
Object detection reaped many of the benefits from the success of convolutional representation.
There are two mainstream deep detection frameworks:
1) two-step detectors: R-CNN~\cite{girshick2014rich}, Fast R-CNN~\cite{girshick2015fast}, and Faster R-CNN~\cite{ren2015faster}.
2) one-step detectors: SSD~\cite{liu2016ssd}, YOLO9000~\cite{redmon2017yolo9000} and RetinaNet~\cite{lin2017focal}. While these methods rely on so-called anchor boxes to predict offsets for bounding box estimation which are usually derived from dataset depended statistics, more recent works study anchor-free object detection by predicting keypoints such as centers or corners~\cite{zhou2019objects, law2018cornernet, tian2019fcos}. We utilize Faster R-CNN, one of the most popular object detectors, as our detection model basis.

\minisection{Image-based 3D Object Detection.}
A large body of research \cite{chen20153d, mono3d, mousavian2017deep3dbox, m3drpn, monodis, pseudolidar, am3d, 3drcnn, deepmanta} has focused on 3D object detection rather than detecting objects on the image plane since understanding the environment in 3D is of fundamental importance for many vision applications such as robotic navigation and autonomous driving.
Over the last years, these methods predominantly regressed the 3D bounding box parameters directly from the image domain using 2D CNNs \cite{chen20153d, mono3d, mousavian2017deep3dbox, m3drpn, monodis}. More recent works \cite{pseudolidar, am3d} propose to transform the data representation to detect objects by utilizing recent advances in point cloud based 3D object detection \cite{zhou2018voxelnet, shi2018pointrcnn, lang2019pointpillars, chen2017multi} where methods showed impressive performance on point clouds retrieved from LiDAR sensors. Other works leverage shape prior cues to enhance the performance of image-based 3D object detection \cite{3drcnn, deepmanta}.
In contrast, our work extends beyond the single-frame domain, leveraging temporal information to enhance monocular 3D detection by utilizing instance-level uncertainty estimation in combination with our LSTM-based motion module.

\minisection{2D Object Tracking.}
Object tracking in the image-domain has been explored extensively in the last decade~\cite{yilmaz2006object,salti2012adaptive,smeulders2014visual}. 
Early methods~\cite{bolme2010visual,gaidon2016virtual,kristan2015visual} track objects based on correlation filters. Recent ConvNet-based methods typically leverage pre-trained object recognition networks. Some generic object trackers are trained entirely online, starting from the first frame of a given video~\cite{hare2016struck,babenko2009visual,kalal2012tracking}. 
While object tracking has been tackled by many different paradigms~\cite{luo2017motreview}, including tracking by object verification~\cite{tao2016siamese}, tracking by correlation~\cite{bertinetto2016fully} and tracking by detection~\cite{feichtenhofer2017detectandtrack}, we build upon on the recent successes of tracking by detection methods~\cite{bergmann2019tracktor, mykheievskyi2020odesa, zhang2019mmMOT}. These are typically distinguished by their data association mechanism, for which many different algorithms have been explored, such as network flow~\cite{zhang2008global}, conditional random fields~\cite{choi2015near}, multi-hypothesis tracking~\cite{kim2015multiple} and quadratic pseudo boolean optimization~\cite{ess2008mobile}. These association mechanisms are fueled by different appearance and location based cues such as consistency of 2D and 3D motion~\cite{choi2015near, bergmann2019tracktor, Scheidegger2018pmbm} as well as visual appearance similarity~\cite{voigtlaender2019mots, bergmann2019tracktor}. Our method combines the strengths of different similarity cues for object tracking.

\minisection{3D Object Tracking.}
The previously discussed methods for object tracking in the image-domain usually only take 2D visual features into consideration, where the search space is restricted near the original position of the object. This works well for a static observer, but fails in a dynamic 3D environment. Therefore, various research works have proposed to further leverage 3D information to narrow down the search space and stabilize the trajectory of target objects \cite{MOTBeyondPixels, luiten2020track, Scheidegger2018pmbm, Li_2018_ECCV, Osep17ICRAciwt}. Scheidegger~\etal.~\cite{Scheidegger2018pmbm} estimate 3D object positions from single images and add a 3D Kalman filter on the 3D positions to get more consistent localization results and thus improve association. Another approach by Luiten~\etal.~\cite{luiten2020track} proposes to use dynamic 3D reconstruction for tracklet association in 3D space to improve long-term tracking. Osep ~\etal.~\cite{Osep17ICRAciwt} and Li~\etal.~\cite{Li_2018_ECCV} study the extension of this paradigm to 3D bounding box tracking using 3D information obtained from stereo cameras. Recent work by Weng~\etal.~\cite{weng2020ab3dmot} proposes specific evaluation metrics for this task and provides a baseline relying on 3D detections from LiDAR. Our work is in line with the recent developments in this field and aims to improve data association by leveraging 3D information, but goes beyond this by integrating both visual appearance and 3D localization information. Moreover, we utilize only a monocular camera and GPS information to track objects in 3D.

\minisection{Joint Detection and Tracking.}
While accurate object detection is a crucial component for tracking, i.e. to track an object throughout a video we must first detect it in every frame it is present, information from tracking can also provide strong priors for object detection. Therefore, many works have studied ways of integrating techniques from object detection with tracking algorithms. Feichtenhofer~\etal.~\cite{feichtenhofer2017detectandtrack} compute correlation maps between a pair of frames which are subsequently used to predict the 2D bounding box deformation of each instance between the two frames. In \cite{bergmann2019tracktor}, the authors utilize a single-frame Faster-RCNN model to predict the 2D bounding box deformation from the second stage refinement module alone. Recently, Lu~\etal.~\cite{Lu2020RetinaTrack} extend RetinaNet \cite{lin2017focal} to learn track-level instance embeddings via a triplet loss in a joint detection and tracking model. In contrast to the popular triplet loss training scheme, Pang~\etal.~\cite{pang2020quasidense} propose to exploit given object proposals within the Faster-RCNN framework to learn instance embedding similarity using densely-connected contrastive pairs and show impressive improvement to instance embedding quality. Our work leverages these findings within our 3D detection and tracking framework, extending the work of Pang~\etal. beyond the image domain.
In the 3D tracking domain, the paradigm of joint detection and tracking also caught researcher's attention recently, with Yin~\etal.~\cite{yin2020center} combining 3D LiDAR detection with the center-based association paradigm from CenterTrack \cite{zhou2020centertrack} to perform joint 3D detection and tracking. However, because the depth can be perceived directly in LiDAR data, the task is much easier. Our work in contrast is able to perform joint 3D detection and tracking solely from 2D image information, estimating the 3D properties of the visible objects from the image information as well as associating these objects over time.

\minisection{Autonomous Driving Datasets.} 
Driving datasets have comprised some of the most popular benchmarks for computer vision algorithms in the last decade. Benchmarks like KITTI~\cite{geiger2012kitti}, UA-DETRAC~\cite{wen2015ua}, Cityscapes~\cite{cordts2016cityscapes} and Oxford RobotCar~\cite{maddern2017robotcar} provide well annotated ground truth for visual odometry, stereo reconstruction, optical flow, scene flow, object detection and tracking as well as semantic segmentation. However, due to the high effort that the annotation of these datasets requires, these benchmarks have been limited in scale. In recent years, the topic of autonomous driving catched on more and more in the industry, providing the resources for new, large-scale driving benchmarks for computer vision, providing annotations for 3D computer vision tasks like 3D object detection and tracking at an unprecedented scale. Therefore, benchmarks like BDD100K~\cite{yu2018bdd100k}, NuScenes~\cite{caesar2019nuscenes}, Argoverse~\cite{Chang2019argoverse} and Waymo Open \cite{waymo} have attracted a lot of attention by the research community.
Still, accurate 3D annotations are challenging to obtain and expensive to measure with 3D sensors like LiDAR. To overcome this difficulty, there has been significant work on virtual driving datasets: virtual KITTI~\cite{gaidon2016virtual}, SYNTHIA~\cite{ros2016synthia}, GTA5~\cite{richter2016playing}, VIPER~\cite{richter2017playing}, CARLA~\cite{dosovitskiy2017carla}, and Free Supervision from Video Games (FSV)~\cite{pk-fsvg-2018}. These datasets have the potential to lower the costs of training accurate deep learning models for applications like autonomous driving tremendously, thus opening up the capabilities of these to a broader audience.
The closest dataset to ours is VIPER~\cite{richter2017playing}, which provides a suite of videos and annotations for various computer vision problems while we focus on object tracking.
We extend FSV~\cite{pk-fsvg-2018} to include object tracking in both 2D and 3D, as well as fine-grained object attributes, control signals from driver actions.

In the next section, we describe how to generate 3D object trajectories from 2D dash-cam videos.
Considering the practical requirement of autonomous driving, we primarily focus on online tracking systems, where only the current and constant number of past frames are accessible to a tracker.

%% file: 03Technical.tex
\begin{figure*}[htb!]
	\includegraphics[width=1.0\linewidth]{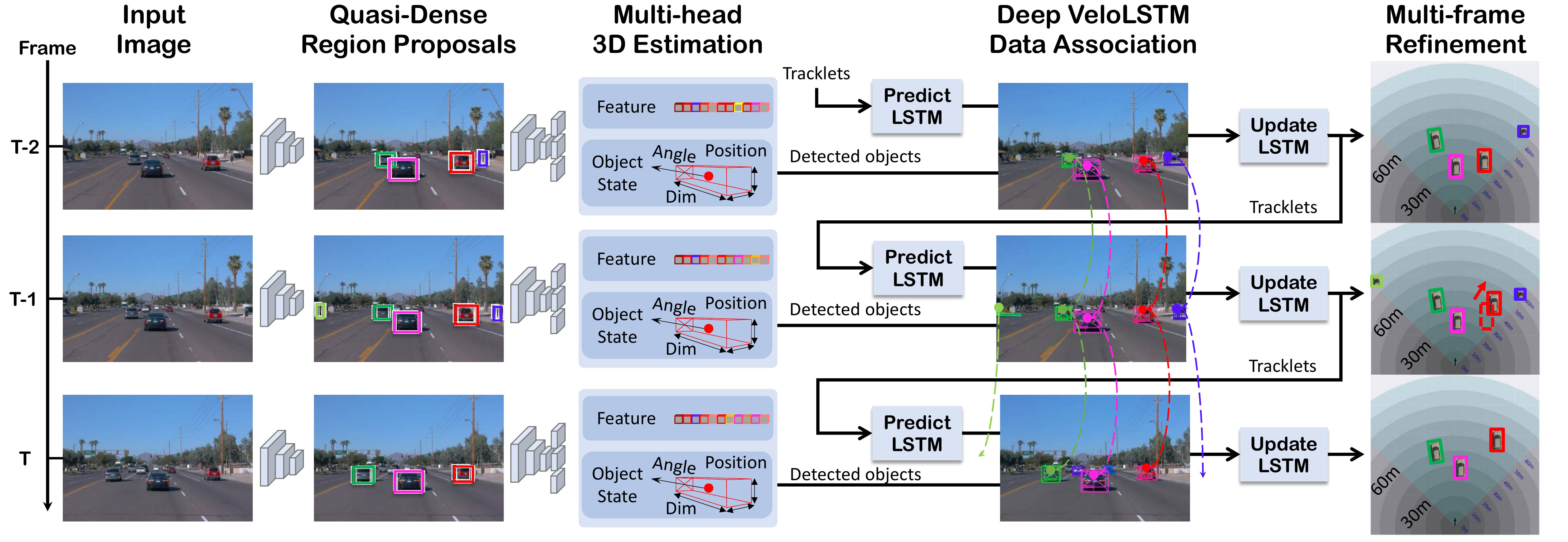}
	$\phantom{\hspace{0.030\linewidth}}%
	\underbracket{\hspace{0.285\linewidth}}_{\substack{\colorbox{white}{(a) }}}%
	\underbracket{\hspace{0.200\linewidth}}_{\substack{\colorbox{white}{(b) }}}%
	\underbracket{\hspace{0.340\linewidth}}_{\substack{\colorbox{white}{(c) }}}%
	\underbracket{\hspace{0.110\linewidth}}_{\substack{\colorbox{white}{(d) }}}$
	\figcaption{Overview of our monocular quasi-dense 3D tracking framework.}{
		Our online approach processes monocular frames to estimate and track regions of interest (RoIs) in 3D (a).
		For each RoI, we learn the 3D layout (\ie, depth, orientation, dimensions, a projection of 3D center) estimation and instance-level feature embedding (b).
		With the 3D layout, our VeloLSTM helps to predict object states, and our 3D tracker produces robust linking across frames leveraging motion-aware association and depth-ordering matching (c). VeloLSTM further refines the 3D estimation by fusing object motion features of the previous frames (d).
	}
	\label{fig:overview}
\end{figure*}

\section{Joint 3D Detection and Tracking}
\label{sec:joint_3d_detection_tracking}
Our goal is to jointly track objects across frames and infer the full 3D information of all tracks from a single monocular video stream and a GPS sensor. The 3D information includes the position, orientation, and dimensions of each object instance. 
\figref{fig:overview} shows an overview of our system.
Images are first passed through a backbone network and Region Proposal Network (RPN) to generate 2D object proposals (\secref{sec:object_detection}).
These 2D proposals are then fed into two lightweight multi-head networks to infer per-instance similarity feature embedding (\secref{sec:quasi_dense_similarity_feature}) and 3D information (\secref{sec:3d_estimation}).
Using both feature embedding and 3D information to generate similarity metrics between all trajectories and detected proposals, we leverage estimated 3D object instance of current trajectories to track them through time (\secref{sec:3d_tracking}).
We introduce motion-aware data association and depth-ordering matching to solve the occlusion problem in tracking.
Finally, we refine the 3D location of objects using the motion model through the newly matched trajectory (\secref{sec:lstm_motion}).

\subsection{Problem Formulation}
\label{sec:problem_formulation}
We phrase the 3D tracking problem as a supervised learning problem.
The goal is to find $N$ trajectories $\mathbb{T} = \{\mathbf{\tau}^1,\ldots,\mathbf{\tau}^N\}$ that match the ground truth trajectories in a video. Each trajectory $\mathbf{\tau}^i_{a,b}$ links to a sequence of detected object states $\langle \mathbf{s}^{(i)}_a,\mathbf{s}^{(i)}_{a+1},\ldots,\mathbf{s}^{(i)}_b \rangle$ starting at the first visible frame $a$ and ending at the last visible frame $b$.
The state of an object $i$ at frame $a$ is given by $\mathbf{s}^i_a = [P, O, D, F, \Delta P] \in \mathbb{R}^{10}$
where $P$ defines the 3D world location $(x, y, z)$ of the object center, $O$ for object orientation $\theta$, $D$ for dimensions ($l, w, h$), $F$ for appearance feature $f_\subfix{app}$, and $\Delta P$ stands for its velocity $(\dot{x}, \dot{y}, \dot{z})$.
The 3D object states in the world coordinates enable the use of our depth-ordering matching and motion-aware association.

On a moving platform with a pose $[\mathbf{R}|\mathbf{t}]$, the mounted camera captures objects found in the 3D world.
We describe an object in the form of a 3D bounding box $\mathbf{X} \in \mathbb{R}^{3\times8}$ in the world coordinates to avoid the changing reference axes introduced by ego-motion.
Each 3D bounding box $\mathbf{X}$ can be viewed as the composition of object location $P$, dimensions $D$, and orientation $O$.
The location $P$ can be projected onto the camera plane using camera pose $[\mathbf{R}|\mathbf{t}]$ as a point $P_\subfix{cam}=(x_\subfix{cam}, y_\subfix{cam}, d)$ where $d$ is the depth from camera.
The $P_\subfix{cam}$ can be further projected to image plane using camera intrinsics $\mathbf{K}$ as the projection of the 3D bounding box center $C=({u_c}, {v_c})$.
The intrinsic parameter $\mathbf{K}$ can be obtained from camera calibration.
Therefore, we can ``lift'' an object center $C$ in the image plane with a depth $d$ using camera parameters $\mathbf{M}=\mathbf{K}[\mathbf{R}|\mathbf{t}]$ to a object location $P$ in the world.
Given the mapping from 3D points in the 3D world and 2D points in the image is known, we enclose a 2D bounding box $B=(u_\subfix{min},v_\subfix{min},u_\subfix{max},v_\subfix{max})$ over the projected 3D bounding box $\mathbf{M} \times \mathbf{X}$ as the 2D annotation for object detectors.
The extrinsic parameter $[\mathbf{R}|\mathbf{t}]$ can be retrieved from a commonly equipped GPS or IMU sensor.
The recorded extrinsic parameter will be used later in the 3D tracking phase to cancel out the ego-motion of the moving platform.

The whole system is powered by an end-to-end multi-head convolutional network trained on a considerable amount of ground truth supervision.
Next, we discuss each component in more detail.

\subsection{Candidate Box Detection}
\label{sec:object_detection}
We employ Faster R-CNN~\cite{ren2015faster} trained on our dataset to provide object candidates in the form of bounding boxes.
As a two-stage detector, Faster R-CNN first generates regions of interest (RoIs) using a multiple-scale Region Proposal Network (RPN) and then utilizes a region convolutional neural network (R-CNN) to classify the RoI as well as to refine its localization.
The multiple-scale RPN generates RoIs from the image feature and obtains its regional feature maps at different levels as the input of our multi-head detector, 3D estimator, and quasi-dense feature extractor networks.
RoI align~\cite{he2017mask} is used instead of RoI pool to obtain the regional representation, which reduces the misalignment of two-step quantization.
Each object proposal (\figref{fig:overview}(a)) corresponds to a 2D bounding box $B$ as well as an estimated projection of the 3D bounding box center $\hat{C}$.
The proposals are used to locate the candidate objects and extract their appearance features.
However, the center of the objects' 3D bounding box usually does not project directly to the center of its 2D counterparts.
Hence, we also provide an estimation of the 3D bounding box center for better accuracy.

\minisection{Projection of 3D bounding box center.}
To estimate the 3D layout from single images more accurately, we extend the regression process to predict a projected 2D point of the 3D bounding box center from an RoI aligned feature $F$ using Smooth L1 loss~\cite{huber1964loss}.
Estimating a projection of the 3D center is crucial since a small gap in the image coordinate can cause a tremendous shift in 3D.
With the extended multi-head detector, the model simultaneously regresses a bounding box $B$ and an estimated center $\hat{C}$ from an object proposal.
We discuss how an estimated center $\hat{C}$ is used in~\secref{sec:3d_estimation} for lifting predicted 3D bounding boxes.

\begin{figure*}[htp]
	\begin{center}
		\includegraphics[width=1.0\linewidth]{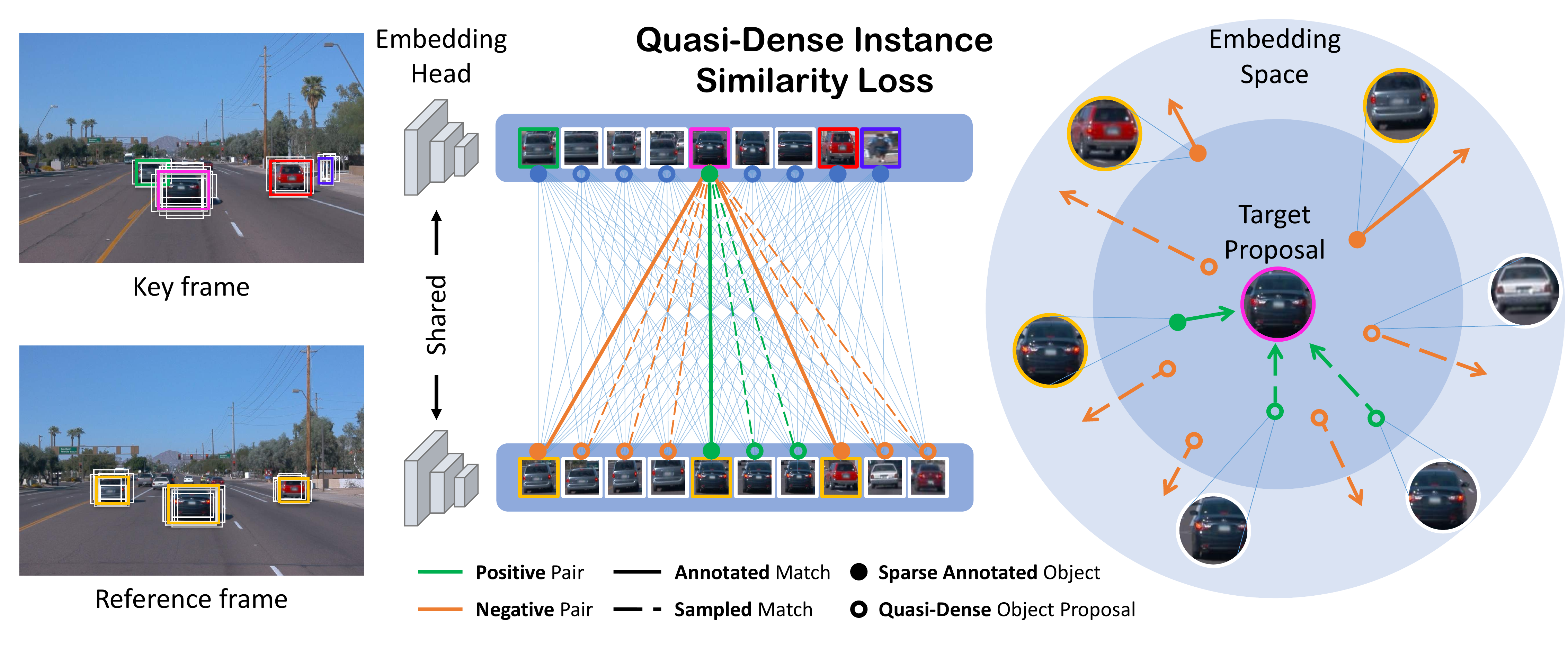}
	\end{center}
	\figcaption{The illustration of the quasi-dense similarity learning.}{We leverage quasi-dense object proposals (all the circles), instead of traditional sparse ground truth (solid circles), to train a discriminative feature space by comparing the region proposal pairs between the key frame and the reference frame. The quasi-dense instance similarity loss pulls the feature embedding of different object away from its paired target proposal and draws the embedding of same object pairs together in a high dimensional space.}
	\label{fig:quasi_dense}
\end{figure*}

\subsection{Quasi-dense Similarity Learning}
\label{sec:quasi_dense_similarity_feature}
Data association, one challenging problem in the multiple object tracking field, requires distinguishable feature embeddings to match detections and tracklets~\cite{hermans2017defense, valmadre2017siamesefc}.
We leverage quasi-dense proposals to train a discriminative feature space using contrastive loss.
\figref{fig:quasi_dense} depicts the difference of a quasi-dense approach comparing to conventional sparse ground-truth based approaches.
Unlike \textit{sparse} feature learning, which is commonly used in multiple object tracking that learns only from ground-truth bounding boxes, quasi-dense similarity learning utilizes all object proposals to discriminate positive proposals from negative ones.
Unlike \textit{dense} approaches, which is commonly used in single object tracking that compare tracklets with the whole image, our approach focuses on potential regions of interest to reduce the computational redundancy.

To obtain the proposal-to-proposal quasi-dense instance embedding correspondences, we establish pair-wise instance similarity loss between two neighboring frames, namely, a key frame and a reference frame.
The multiple-scale RPN generates RoIs from the two images and obtains their regional feature maps at different levels.
We assign RoIs with IoU scores to an object over $0.7$ as \textit{positive} proposals, and those lower than $0.3$ as \textit{negative} proposals.
Using the regional feature maps, we propose a quasi-dense embedding head to extract feature embeddings from positive and negative proposals.
Adapting contrastive characteristics to the tracking problem, we defined region proposal pairs from two frames matched to the same object identity as positive pairs, and those matched to different identities are negative ones.

Given a key frame at time $a$, we sample a reference frame within a temporal interval $n$, where $n\in [-3, 3]$ throughout all the experiments.
For each target proposal $\mathbf{s}_{a}$, we optimize its feature embedding $F_{\mathbf{s}_{a}}$ using cross entropy loss combining a non-parametric softmax activation~\cite{wu2018nonparamsoftmax}.
\begin{align}
	\mathcal{L}_\subfix{embed} & = -\log
	\frac{\text{exp}(F_{\mathbf{s}_{a}} \cdot F_{\mathbf{s}_{a+n}}^{+})}
	{\text{exp}(F_{\mathbf{s}_{a}} \cdot F_{\mathbf{s}_{a+n}}^{+}) + \sum_{F_{\mathbf{s}_{a+n}}^{-}}\text{exp}(F_{\mathbf{s}_{a}} \cdot F_{\mathbf{s}_{a+n}}^{-})},
	\label{eqn:quasi_dense_loss}
\end{align}
where feature embedding $F_{\mathbf{s}_{a}}$ of the target proposal is learned to associate its positive referenced embedding $F_{\mathbf{s}_{a+n}}^{+}$, and distinguish all its negative referenced ones $F_{\mathbf{s}_{a+n}}^{-}$.
However, the formula above considers only one positive object proposal.
To fully leverage the quasi-dense characteristics, we can balance the number of positive and negative examples by comparing the target proposal to all positive proposals.

Since we accumulate the whole positive targets, Equation~\ref{eqn:quasi_dense_loss} can be reformulated as follows
\begin{align}
	\mathcal{L}_\subfix{embed} & = \log [1 + \sum_{F_{\mathbf{s}_{a+n}}^{+}} \sum_{F_{\mathbf{s}_{a+n}}^{-}} \text{exp}(F_{\mathbf{s}_{a}} \cdot F_{\mathbf{s}_{a+n}}^{-} - F_{\mathbf{s}_{a}} \cdot F_{\mathbf{s}_{a+n}}^{+})]
	\label{eqn:quasi_dense_loss_update}
\end{align}
The loss term $\mathcal{L}_\text{embed}$ of the above formulation minimizes the cosine distance of the target proposal to all positive referenced examples while maximizing the cosine distance to all negative examples. 
By balancing positive and negative examples, we encourage the network to learn an embedding space that discriminates between instances most effectively while being invariant to input perturbations like change in viewpoint or lighting.
In addition, we apply an auxiliary loss encouraging the learning of cosine similarity of target proposal $F_{\mathbf{s}_{a}}$ and references $F_{\mathbf{s}_{a+n}}$
\begin{align}
	\mathcal{L}_\subfix{aux} & = (\frac{F_{\mathbf{s}_{a}} \cdot F_{\mathbf{s}_{a+n}}}{||F_{\mathbf{s}_{a}}|| \cdot ||F_{\mathbf{s}_{a+n}}||} - \mathds{1}({\mathbf{s}_{a}, \mathbf{s}_{a+n}}))^2
	\label{eqn:quasi_dense_loss_aux}
\end{align} 
where $\mathds{1}({\mathbf{s}_{a}, \mathbf{s}_{a+n}})$ is the matching pair indicator function producing $1$ if $\mathbf{s}_{a}$ and $\mathbf{s}_{a+n}$ match to the same object and $0$ otherwise.
We define the full quasi-dense instance similarity loss as 
\begin{align}
	\mathcal{L}_\subfix{similarity} & = \lambda_\subfix{embed}\,\mathcal{L}_\subfix{embed}+\mathcal{L}_\subfix{aux}
	\label{eqn:quasi_dense_loss_all}
\end{align}

In short, the embedding head optimizes the quasi-dense feature embedding pairs to discriminate different instances using instance similarity loss $\mathcal{L}_\subfix{similarity}$. Next, we discuss how we estimate the 3D extent of an object.

\subsection{3D Bounding Box Estimation}
\label{sec:3d_estimation}
We estimate complete 3D bounding box information (\figref{fig:overview}(b)) from an RoI in the image via a feature representation of the pixels in the 2D bounding box.
The RoI feature vector $F$ is extracted from a deep convolutional backbone network.
Each of the 3D information is estimated by passing the RoI features through a $2$-layer convolutional sub-network with shared parameters and subsequently a $4$-layer 3x3 convolution sub-network, which extends the stacked linear layers design of Mousavian~\etal~\cite{mousavian2017deep3dbox}.
We estimate all necessary parameters for inferring 3D bounding boxes from images, whereas Mousavian~\etal~\cite{mousavian2017deep3dbox} focus on object orientation and dimensions from 2D boxes.
Besides, our approach integrates 2D detection with 3D estimation, 3D tracking and motion refinement, while Mousavian~\etal~\cite{ mousavian2017deep3dbox} crops the input image with pre-computed boxes.

This network is trained using ground truth depth $d$, 3D bounding box center projection $C$, dimensions $D$, and orientation $O$.
We explain how we estimate each component in detail in the following paragraphs.

\minisection{3D World Location.}
Unlike previous approaches~\cite{mono3d, mousavian2017deep3dbox, Hu2019Mono3DT} that lift objects in camera coordinates, we infer 3D location $P$ in the world coordinates from monocular images.
The network learns to regress a logarithmic depth value $\log(\hat{d})$ scaled with a constant $r$ with Smooth L1 loss~\cite{huber1964loss}.
Compared with the characteristics of inverse depth value encoding $1/d$ in~\cite{Hu2019Mono3DT} which compresses the target range ($[5, 100]$) drastically to ($[0.80, 0.99]$), our scaled logarithmic target provides a more realistic scaling inside our target range of the encoded depth value ($[1.39, 4]$).
The scaled logarithmic approach remedies the curse of regression that small encoded depth value change results in a large depth estimation variation.
A projected 3D location $P$ is calculated using an estimated 2D projection of the 3D object center $C$ as well as the depth $d$ and camera transformation $\mathbf{M}$.

\begin{figure}[htp]
	\begin{center}
		\includegraphics[width=1.0\linewidth]{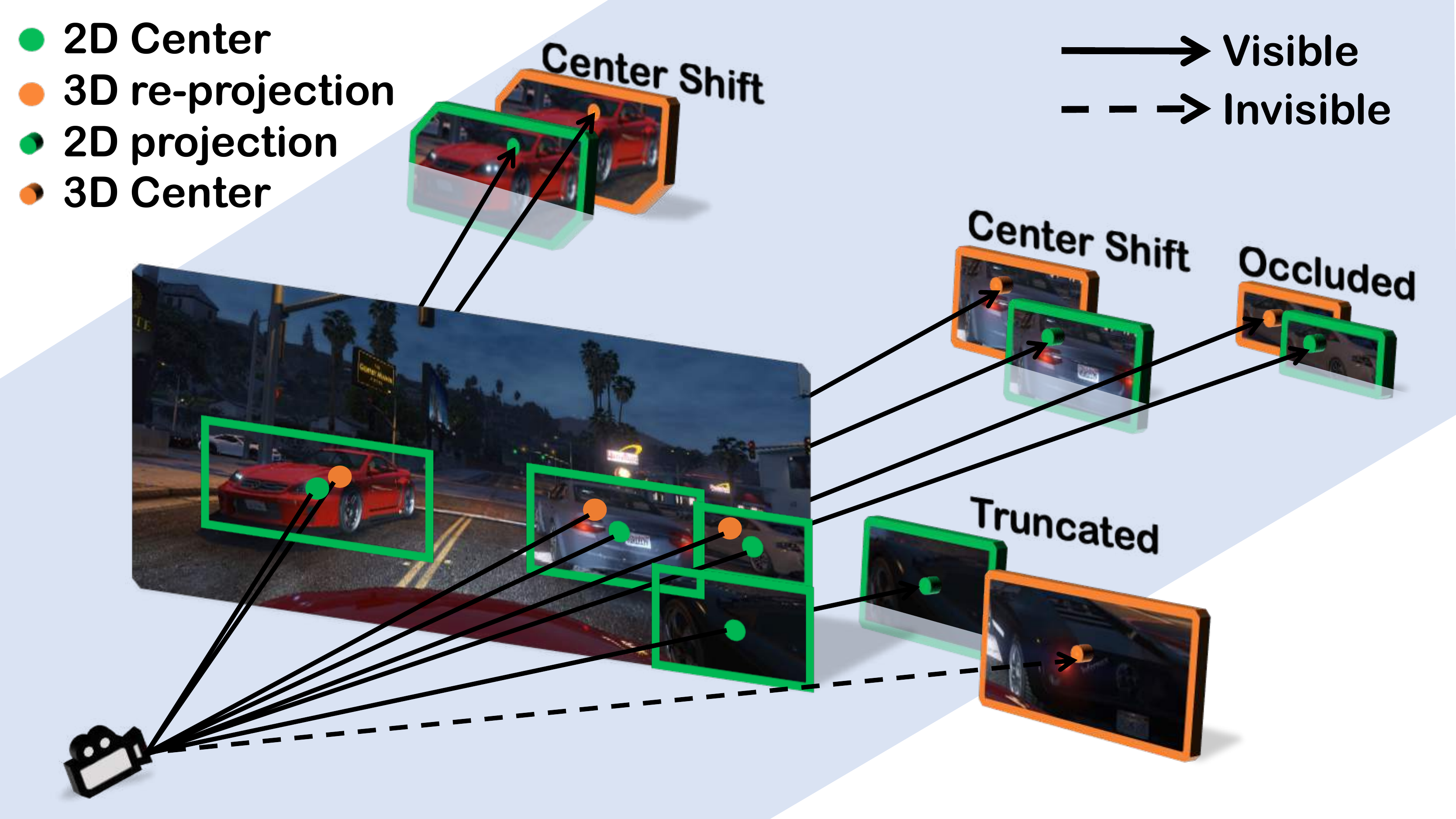}
	\end{center}
	\figcaption{3D projected centers are perferred over 2D bounding boxes.}{
		A visible 2D center projection point may wrongly locate the object away from the ground plane in the 3D coordinates and would inevitably suffer from more severe scenarios, occluded or truncated.
		The center shift causes object misalignment for GT and predicted tracks and harms 3D IoU AP performance.}
	\label{fig:3d_center}
\end{figure}

\minisection{Initial Projection of 3D Bounding Box Center.}
The 3D object location in the current camera coordinate system can be obtained by estimating the depth of the detected 2D bounding box $B$ as well as the 3D object center projection $C$ and the camera calibration.
The 3D world position $P$ can be inferred by transforming the 3D object location based on the observer's pose.
We find that accurately estimating the 3D object center and its projection is critical for accurate 3D bounding box localization.
However, the projection of the 3D center can be far away from the 2D bounding box center.
Naturally, a center shift from the 2D box center since the 2D bounding box encloses the 3D object from the observer's perspective.
After occlusion and truncation, the estimated 2D bounding box can hardly cover the correct dimension with only the visible area of the objects. 
In contrast, the 3D bounding box is defined by the full physical dimensions of an object.
The projected 3D center can lie even outside the detected 2D boxes.
For instance, the 3D bounding box of a truncated object can be out of the camera view. These situations are illustrated in \figref{fig:3d_center}.

\minisection{Single-frame 3D Confidence.}
Given the high variance in estimating the object-observer relationship in the real world, we find a single-frame depth estimation is prone to predict an erroneous guess if the target object is only partially observable or in unforeseen situations.
While the uncertainty of the model estimation has been well-studied~\cite{kendall2017uncertainty, brazil2020kinematic}, we aim to solve the uncertainty problem in a spatial-temporal approach, focusing on the confidence of the 3D estimation.
During training, the network is supervised by the target confidence score $c_\subfix{depth} = \exp({-|\hat{d} - d|/r})$, which we formulated as an exponential of negative depth distance of prediction and ground truth scaled by a constant $r$.
The learned confidence $\hat{c}_\subfix{depth}$ is used during inference as a 3D objectness score in the detection phase, and as a weighting score in the tracking phase.
The confidence helps to handle the ambiguity of the single-frame depth estimation by balancing single-frame depth estimation and multiple-frame motion model location prediction.

\minisection{Object Orientation.}
Given the image coordinate distance $\hat{u} = \hat{u}_{c} - w_\subfix{img} / 2$ from an object center $\hat{u}_{c}$ to the horizontal center of an image $w_\subfix{img} / 2$ and the focal length $f$, the object rotation $\theta_\subfix{cam}$ in the camera coordinates transformed from an observed vehicle heading $\theta_\subfix{obs}$ with simple geometry, $\theta_\subfix{cam} = [\theta_\subfix{obs} + \arctan (\hat{u} / f)] \mod 2\pi$.
Later we can restore the global rotation $\theta$ in the world coordinates using ego-vehicle heading obtained in $\mathbf{R}$.
Following Mousavian~\etal~\cite{mousavian2017deep3dbox} for $\theta_\subfix{obs}$ estimation, we first classify the angle into two bins and then regress the residual relative to the bin center using Smooth L1 loss~\cite{huber1964loss}.

\minisection{Object Dimension.}
In driving scenarios, the high variance of the distribution of the dimensions of different categories of vehicles (\eg, car, bus) results in difficulty classifying various vehicles using unimodal object proposals.
Therefore, we regress the dimensions $(\hat{l}, \hat{w}, \hat{h})$ to the ground truth dimensions $(l, w, h)$ using Smooth L1 loss~\cite{huber1964loss}.

To sum up, estimating an object's 3D properties provides us with an observation of its location $P$ with orientation $O$, dimensions $D$ and 2D projection of its 3D center $C$. Next, we discuss how we associate the estimation across frames.

\subsection{Data Association and Tracking}
\label{sec:3d_tracking}
Given a set of candidate detections $\mathbb{S}_a = \{\mathbf{s}^1_a,\ldots,\mathbf{s}^M_a\}$ and a set of tracks $\mathbb{T}_a = \{\mathbf{\tau}^J,\ldots,\mathbf{\tau}^K\}$ at frame $a$ where $1 \leq J \leq K \leq N$ from $N$ trajectories, our goal is to associate each track $\mathbf{\tau}$ with a candidate detection $\mathbf{s}$, spawn new tracks, or end a track (\figref{fig:overview}(c)) in an online fashion.

We solve the data association problem by using a weighted bipartite matching algorithm.
Affinities between tracks and new detections are calculated from three criteria,
(1) the deep representation similarity $\mathbf{A}_\subfix{deep}(\mathbf{\tau}_a, \mathbf{s}_a)$ of the appearance embeddings $f_\subfix{app}$ between an accumulated feature $F_{\mathbf{\tau}_a}$ from existing tracks and a learned feature $F_{\mathbf{s}_a}$ from object detections;
(2) the overlap of bounding boxes $\mathbf{A}_\subfix{iou}(\mathbf{\tau}_a, \mathbf{s}_a)$ between the projection of a 3D bounding box in current trajectories forward in time $B_{\mathbf{\tau}_a}$ and one of the bounding box candidates $B_{\mathbf{s}_a}$;
and (3) the motion similarity $\mathbf{A}_\subfix{motion}(\mathbf{\tau}_a, \mathbf{s}_a)$ of a pseudo object motion vector $V_{\mathbf{s}_a} = \overrightarrow{P_{\mathbf{\tau}_a} P_{\mathbf{s}_a}}$ and the accumulated motion vector $V_{\mathbf{\tau}_a}$ from current trajectories.
Each trajectory is projected forward in time using the estimated velocity of an object and camera ego-motion.
Here, we assume that ego-motion is given by a sensor, like GPS, an accelerometer, gyro, or IMU.

We define an affinity matrix $\mathbf{A}(\mathbb{T}_a, \mathbb{S}_a) \in \mathbb{R}^{||\mathbb{T}_a|| \cdot ||\mathbb{S}_a||}$ between the information of existing tracks $\mathbb{T}_a$ and candidate detections $\mathbb{S}_a$ as a joint probability of appearance, location, and velocity correlation.
\begin{align}
	\mathbf{A}_\subfix{deep}(\mathbf{\tau}_a, \mathbf{s}_a) &= \frac{\sigma_x(F_{\mathbf{\tau}_a} F_{\mathbf{s}_a}^T) + \sigma_y(F_{\mathbf{\tau}_a} F_{\mathbf{s}_a}^T)}{2}
	\label{eqn:affinity_deep} \\
	\mathbf{A}_\subfix{iou}(\mathbf{\tau}_a, \mathbf{s}_a) &= \exp(\frac{-|B_{\mathbf{\tau}_a} -  B_{\mathbf{s}_a}|}{c}),
	\label{eqn:affinity_iou} \\
	\mathbf{A}_\subfix{motion}(\mathbf{\tau}_a, \mathbf{s}_a) &= \exp(\frac{-|V_{\mathbf{\tau}_a} - V_{\mathbf{s}_a}|}{c}))
	\label{eqn:affinity_motion}
\end{align}
where $\sigma_x$, $\sigma_y$ are sigmoid functions applied along $x$, $y$ direction.

The affinity matrix, $\mathbf{A}(\mathbb{T}_a, \mathbb{S}_a)$, aggregates the semantic, spatial and temporal similarities.
\begin{equation}
	\begin{split}
		\mathbf{A}(\mathbf{\tau}_a, \mathbf{s}_a) 
		&= w_\subfix{deep} \mathbf{A}_\subfix{deep}(\mathbf{\tau}_a, \mathbf{s}_a) \\ 
		&+ (1 - w_\subfix{deep}) \mathbf{A}_\subfix{motion}(\mathbf{\tau}_a, \mathbf{s}_a) \cdot \mathbf{A}_\subfix{iou}(\mathbf{\tau}_a, \mathbf{s}_a)
		\label{eqn:affinity_sum}
	\end{split}
\end{equation}
The weight $w_{deep}$ balances appearance similarity and a hybrid of motion and 3D location overlap.
We utilize a mixture of discriminative quasi-dense feature embeddings, bounding box overlaps, and location reasoning as similarity measures across frames, similar to the design of POI~\cite{yu2016poi}.
We solve data association greedily instead of using Kuhn-Munkres algorithm~\cite{Kuhn1955hungarian} in Mono3DT~\cite{Hu2019Mono3DT}.

Comparing to 2D tracking, 3D-oriented tracking is more robust to ego-motion, visual occlusion, overlaps, and re-appearances.
When a target is temporarily occluded, the corresponding 3D motion estimator can roll-out for a period of time and relocate the 2D location at each new point in time via the camera coordinate transformation.

\minisection{Data Association Scheme.}
\label{sec:data_association}
Similar to previous methods~\cite{Wojke2017deepsort,xiang2015mdptrack,Sadeghian2017untrackable}, we model the lifespan of a tracker using four major subspaces in MDP state space: $\{\mathtt{birth}, \mathtt{tracked}, \mathtt{lost}, \mathtt{death}\}$.
For each new set of detections, the tracker is updated ($\mathtt{tracked}$) using pairs with the highest affinity score (\eqnref{eqn:affinity_sum}).
Each unmatched detection spawns a new tracklet ($\mathtt{birth}$); however, an unmatched tracklet is not immediately terminated ($\mathtt{death}$), as tracklets can naturally disappear ($\mathtt{lost}$) into an occluded region and reappear later.
We address the dynamic object inter-occlusion problem by introducing a motion vector propagation.
A $\mathtt{lost}$ tracklet will not update its feature representation until it reappears, but we still predict its 3D location using the estimated motion.
We continue to predict the 3D location of unmatched tracklets until they disappear from our tracking range (\eg $0.15m$ to $100m$) or die out after their lifespan (\eg $10$ time-steps).
With the sweeping scheme above, we keep disappeared object tracks inside the tracker memory to be able to recover from object occlusions while still keeping the computational requirements for storing all object tracks at a feasible level.

\begin{figure}[t]
	\begin{center}
		\includegraphics[width=1.0\linewidth]{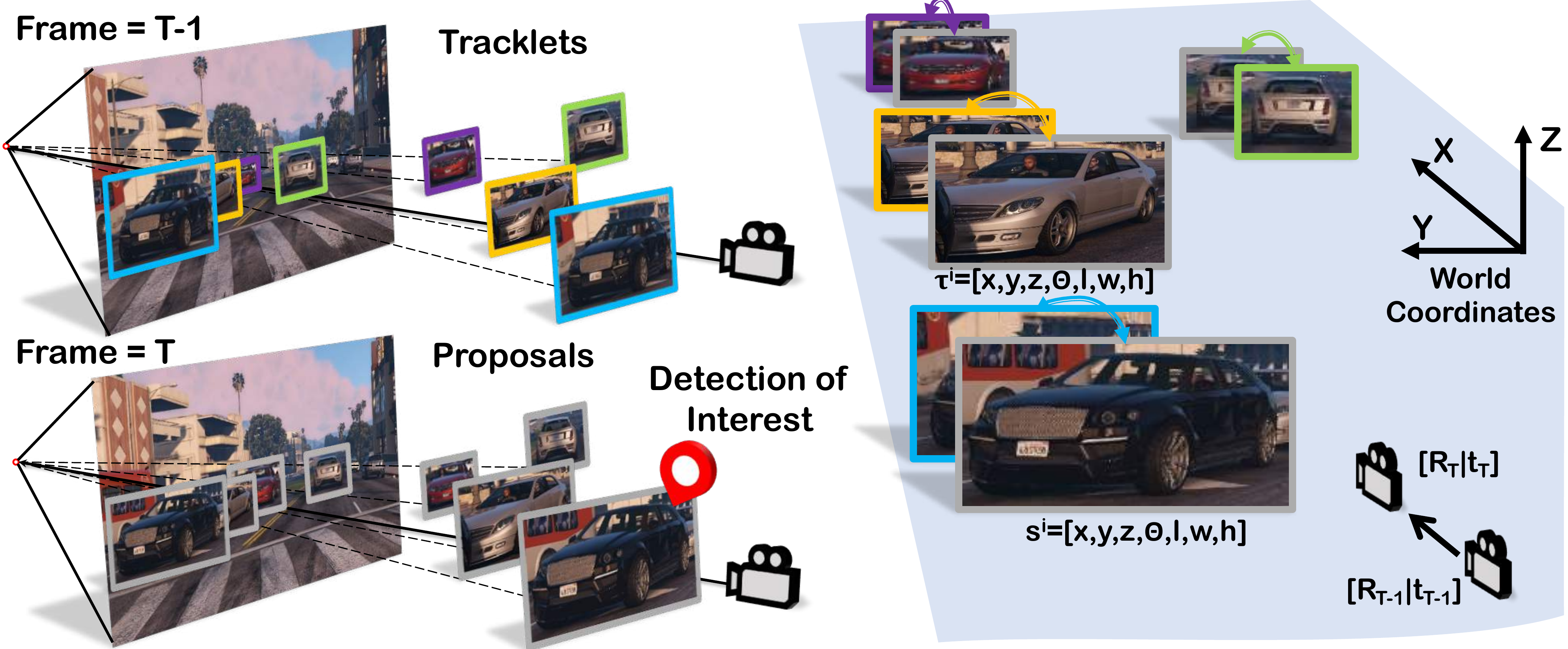}
	\end{center}
	\figcaption{Illustration of depth-ordering matching.}{
		Given the tracklets and detections, we sort them into a list by depth order.
		For each detection of interest (DoI), we calculate similarities of the object state between DoI and each tracklet.
		The location of the object in the world coordinates naturally provides higher probabilities to tracklets near the DoI.
	}
	\label{fig:ordering}
\end{figure}

\minisection{Depth-Ordering Matching.}
\label{sec:object_3d_matching}
We introduce instance depth ordering for assigning a detection to neighbor tracklets, which models the strong prior of relative depth ordering found in human perception.
Extending the formulation in Hu~\etal~\cite{Hu2019Mono3DT}, we consider probabilities of matching each detection of interest (DoI) to potential tracklets using centroid distance of their object state $\mathbf{s}_a$.
Additionally, we introduce object orientation and dimensions to discover different neighbor objects with similar centroid distances.

From the view of each DoI, we compare the pairwise distance of object state $\mathbf{s}_a$ with all the tracklets $\mathbf{\tau}_a$ in the world coordinates.
We directly enforce a smooth distance transition of an exponential function to replace the discrete depth order relationship.
To cancel out the ordering ambiguity of a distant tracklet, we leverage the nature of negative exponential to cut off the matching probability of a distant tracklet smoothly.
So \eqnref{eqn:affinity_iou} becomes
\begin{equation}
	\begin{split}
		\mathbf{A}_\subfix{iou}(\mathbf{\tau}_a, \mathbf{s}_a) &= \exp(\frac{-|\mathbf{\tau}_a - \mathbf{s}_a|}{r}),
		\label{eqn:affinity_iou_update}
	\end{split}
\end{equation}
where the exponential function with a constant $r$ captures the concept that distant objects contribute less to the affinities.
It naturally provides higher probabilities of linking similar neighbor tracklets than those far away.
In this way, we solve the data association problem of moving objects with the help of 3D trajectories in world coordinates.
\figref{fig:ordering} depicts the pipeline of depth ordering.

\begin{figure}[t]
	\begin{center}
		\includegraphics[width=1.0\linewidth]{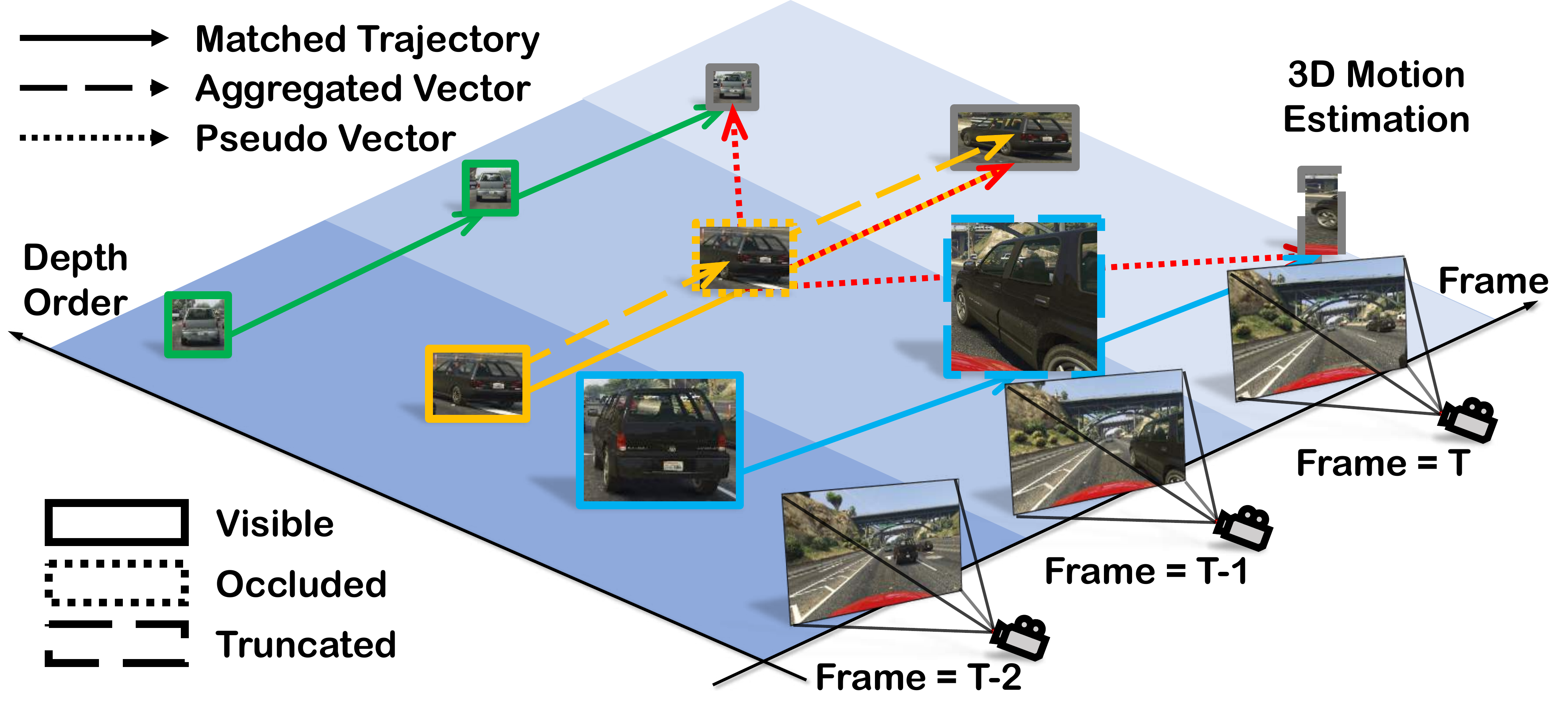}
	\end{center}
	\figcaption{Demonstration of motion-aware data association.}{
		The yellow tracklet is visible all the time, while the red tracklet is occluded by the blue tracklet at frame $T-1$.
		Given pseudo motion vectors (dot arrows from a tracklet to all detection candidates) and the temporal-aggregated motion vector (dashed arrows) by the tracklet, our motion-based data association scheme is encouraged match reappeared candidates in trajectories. 
	}
	\label{fig:data_association_scheme}
\end{figure}

\minisection{Motion-aware Data Association.}
\label{sec:motion_aware}
Assuming a linear movement of an object, we expect a $\mathtt{lost}$ object to reappear at a predictable location based on its velocity and the previously seen locations.
\eqnref{eqn:affinity_motion} encodes both the motion magnitude $||V||$ and heading angle $\angle V$ into the similarity.
Specifically, $\mathbf{A}_\subfix{motion}$ considers accumulated motion vectors $V_{\mathbf{\tau}_a}$ from tracklet trajectories and pseudo motion vectors $V_{\mathbf{s}_a}$ from single-frame depth estimation.
Since the network can not estimate a velocity directly from a single frame, we leverage the bipartite affinity that aggregates the pseudo motion vectors $\overrightarrow{P_{\mathbf{\tau}_a} P_{\mathbf{s}_a}}$ from all $P_{\mathbf{\tau}_a}$ to $P_{\mathbf{s}_a}$.
However, a motion-only similarity in~\eqnref{eqn:affinity_motion} does not distinguish similar motion vectors in different locations.
We propose a new $\mathbf{A}_\subfix{motion}(\mathbf{\tau}_a, \mathbf{s}_a)$ that counts in the motion vectors as well as the centroid location of the objects.
\begin{equation}
	\begin{split}
		\mathbf{A}_\subfix{motion}(\mathbf{\tau}_a, \mathbf{s}_a) &= \mathbf{w_\subfix{cos}}\,\mathbf{A}_\subfix{centroid} + (1-\mathbf{w_\subfix{cos}})\,\mathbf{A}_{pseudo}
		\label{eqn:affinity_motion_update}
	\end{split}
\end{equation}
which composed of the following terms
\begin{equation}
	\begin{split}
		\mathbf{A}_\subfix{centroid} &= \exp(\frac{-|P_{\mathbf{\tau}_a} - P_{\mathbf{s}_a}|}{r}) \\
		\mathbf{A}_\subfix{pseudo} &= \exp(\frac{-|V_{\mathbf{\tau}_a} - V_{\mathbf{s}_a}|}{r}) \\
		\mathbf{w}_\subfix{cos} &= \angle(V_{\mathbf{\tau}_a}, V_{\mathbf{s}_a})
		\label{eqn:affinity_motion_update_decompose}
	\end{split}
\end{equation}
where $\mathbf{w_\subfix{cos}}$ is the normalized cosine similarity of two motion vectors $\angle(V_{\mathbf{\tau}_a}, V_{\mathbf{s}_a}) \in [0, 1]$.
The $\mathbf{w_\subfix{cos}}$ favors centroid distance if two motion vectors within a hemisphere of the same direction and weights more on pseudo motion otherwise.
Therefore, we can find the most aligned motion vector that assigns the detected object $\mathbf{s}_a$ to the target tracklet $\mathbf{\tau}_a$.
\figref{fig:data_association_scheme} illustrates how our motion-aware data association scheme works.
In the next subsection, we show how a motion model helps refine the estimated 3D object state.

\subsection{Motion Model Refinement}
\label{sec:lstm_motion}

For any new tracklet, the detection network is trained to predict the object state $s$ by leveraging RoI features.
For any previously tracked object, the motion model network learns a mixture of multi-view monocular 3D estimates by merging the object state from previous visible frames and the current frame.
Given the learned motion and updated object state, the motion model network predicts the next possible object state to assist the data association procedure.
The update and predict cycles refine the final object state in tandem.

\minisection{VeloLSTM: Deep Motion Estimation and Update.}
To exploit the temporal consistency of objects, we associate the information across frames using two LSTMs for object state prediction and updating (\figref{fig:overview}(c)).
We use two-layer LSTMs with $128$-dim hidden state size to keep track of a 3D object state in the world coordinates $\mathbf{s}^{(i)}_a = \{P, O, D\} = \{x, y, z, \theta, l, w, h\}$ from the $64$-dim output feature.

Prediction LSTM (P-LSTM) models the dynamic object state $\mathbf{s}_{a}$ in 3D world coordinates based on the previous refined object state $\overline{\mathbf{s}}_{a-1}$ with predicted object velocity $\dot{\mathbf{s}}_{a} = \{\Delta P, \Delta O, \Delta D\}$ from previously updated velocities $\dot{\mathbf{s}}_{a-n:a-1}$.
We encode previous $n=5$ frames of object velocity into a $64$-dim velocity feature to model object motion and acceleration from the trajectory.
P-LSTM predicts the next possible object state $\tilde{\mathbf{s}}_{a} = \overline{\mathbf{s}}_{a-1} + \dot{\mathbf{s}}_{a}$. 

Updating LSTM (U-LSTM) balances current observed object state $\hat{\mathbf{s}}_{a}$ from 3D estimation module and previously predicted state $\tilde{\mathbf{s}}_{a-1}$ considering current 3D confidence of depth estimation $\hat{\mathbf{c}}_\subfix{depth}$.
We encode observed, predicted object states into $64$-dim state features and 3D confidence scores into a $64$-dim confidence feature.
By concatenating observation, prediction state features and confidence feature together, we obtained a $192$-dim feature embedding as the input of U-LSTM.
Given the fusion information, U-LSTM refines the object state $\bar{\mathbf{s}}_{a}$ and updates velocities $\dot{\mathbf{s}}_{a-n:a-1}$.

Modeling motion in 3D world coordinates naturally cancels out adverse effects of ego-motion, allowing our model to handle missed and occluded objects.
The LSTMs keep updating the predicted object state $\tilde{\mathbf{s}}_{a-1}$ while assuming a linear velocity model if there is no matched bounding box.
Therefore, we model 3D motion (\figref{fig:overview}(d)) in world coordinates allowing occluded tracklets to move along motion-plausible paths while managing the birth and death of moving objects.

Both LSTM modules are trained with predicted object states $\tilde{\mathbf{s}}_{a}$, refined object states $\overline{\mathbf{s}}_{a}$ from detector prediction, and ground truth object state $\mathbf{s}_{a}$ trajectories.
The refinement loss $\mathtt{L}_\subfix{refine}(\overline{\mathbf{s}}_{a}, \mathbf{s}_{a})$ and prediction loss $\mathtt{L}_\subfix{predict}(\tilde{\mathbf{s}}_{a}, \mathbf{s}_{a})$ aim to reduce the distance of estimated and ground-truth object state using Smooth L1 loss~\cite{huber1964loss}.
The linear motion loss $\mathtt{L}_\subfix{linear}(y_{a-n:a}) = \sum_{a-n:a}|(y_{a} - y_{a-1}) - (y_{a-1} - y_{a-2})| / \Delta t$ focuses on the smooth transition over time $\Delta t$ of object state refinement $\mathtt{L}_\subfix{linear}(\overline{\mathbf{s}}_{a-n:a})$ and prediction $\mathtt{L}_\subfix{linear}(\tilde{\mathbf{s}}_{a-n:a})$.

Conclusively, our pipeline consists of a single-frame monocular 3D object detection model for object-level pose inference and recurrent neural networks for inter-frame object association and matching.
We extend the region processing to include 3D estimation by employing multi-head modules for each object instance.
We introduce a motion-aware association to solve the inter-object occlusion problem.
For tracklet matching, depth ordering lowers the mismatch rate by filtering out distant candidates from a target.
The LSTM motion estimator updates the velocity and states of each object independent of camera movement or interactions with other objects.
The final pipeline produces accurate and smooth object trajectories in the 3D world coordinate system.

%% file: 04Dataset.tex
\section{3D Vehicle Tracking Simulation Dataset}
\label{sec:gta_vehicle_dataset}

It is laborious and expensive to annotate a large-scale 3D bounding box image dataset even in the presence of LiDAR data, although it is much easier to label 2D bounding boxes on tens of thousands of videos~\cite{yu2018bdd100k}.
Therefore, no such dataset collected from real sensors is available to the research community.
To resolve the data problem, we turn to driving simulation to obtain accurate 3D bounding box annotations at no cost of human efforts.
Our data collection and annotation pipeline extend the previous works like VIPER~\cite{richter2017playing} and FSV~\cite{pk-fsvg-2018}, especially in terms of linking identities across frames.

Our simulation is based on \textit{Grand Theft Auto V}, a modern game that simulates a functioning city and its surroundings in a photo-realistic three-dimensional world.
To associate object instances across frames, we utilize in-game API to capture global instance id and corresponding 3D annotations directly.
In contrast, VIPER leverages a weighted matching algorithm based on a heuristic distance function, which can lead to inconsistencies.
It should be noted that our pipeline is real-time, providing the potential of large-scale data collection, while VIPER requires expensive off-line processings.

\begin{figure}[htpb]
    \begin{subfigure}{0.45\linewidth}
        \includegraphics[width=\linewidth]{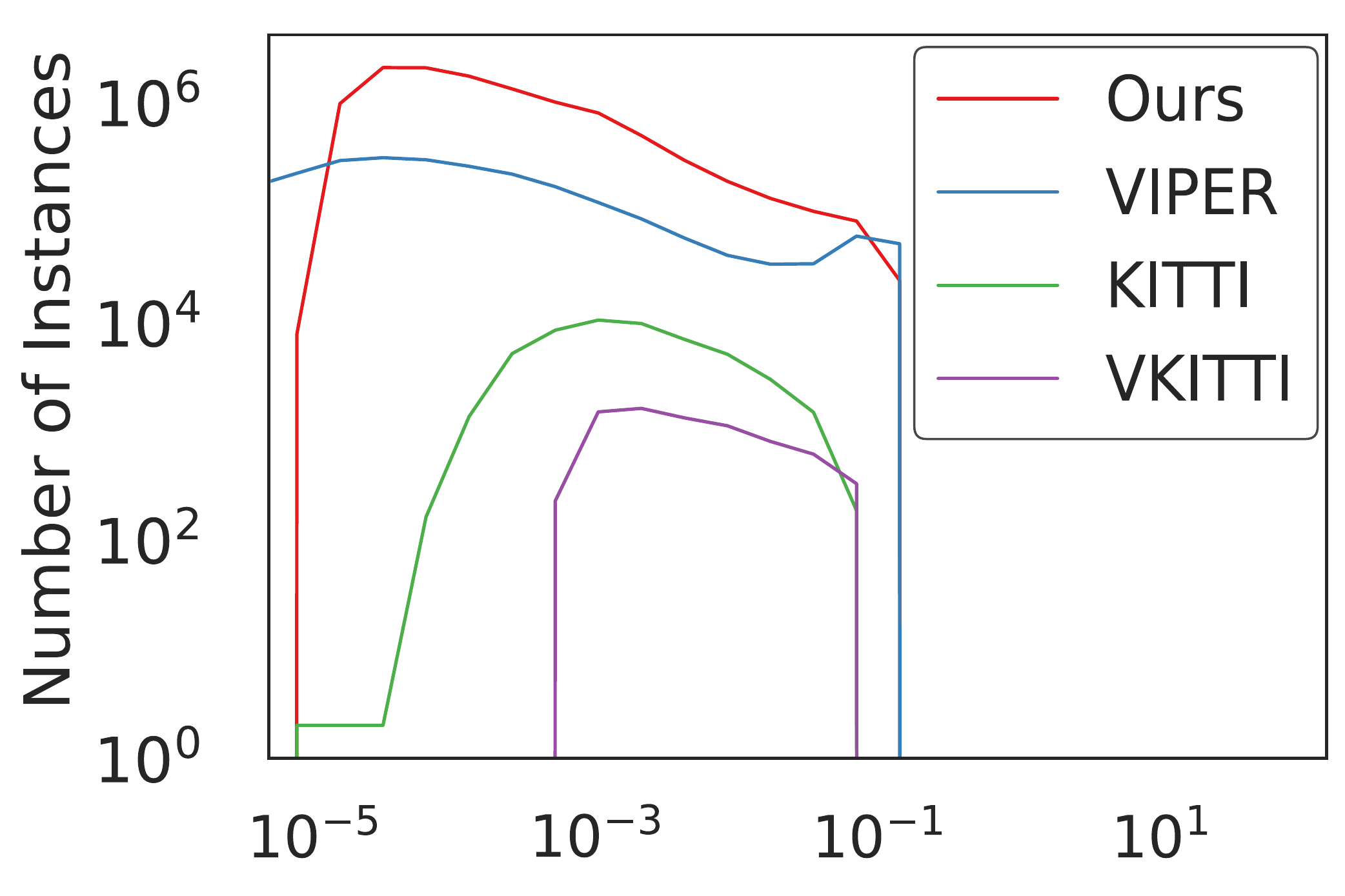}
        \caption{Instance scale}\label{fig:statistics_scale}
    \end{subfigure}
    \begin{subfigure}{0.45\linewidth}
        \includegraphics[width=\linewidth]{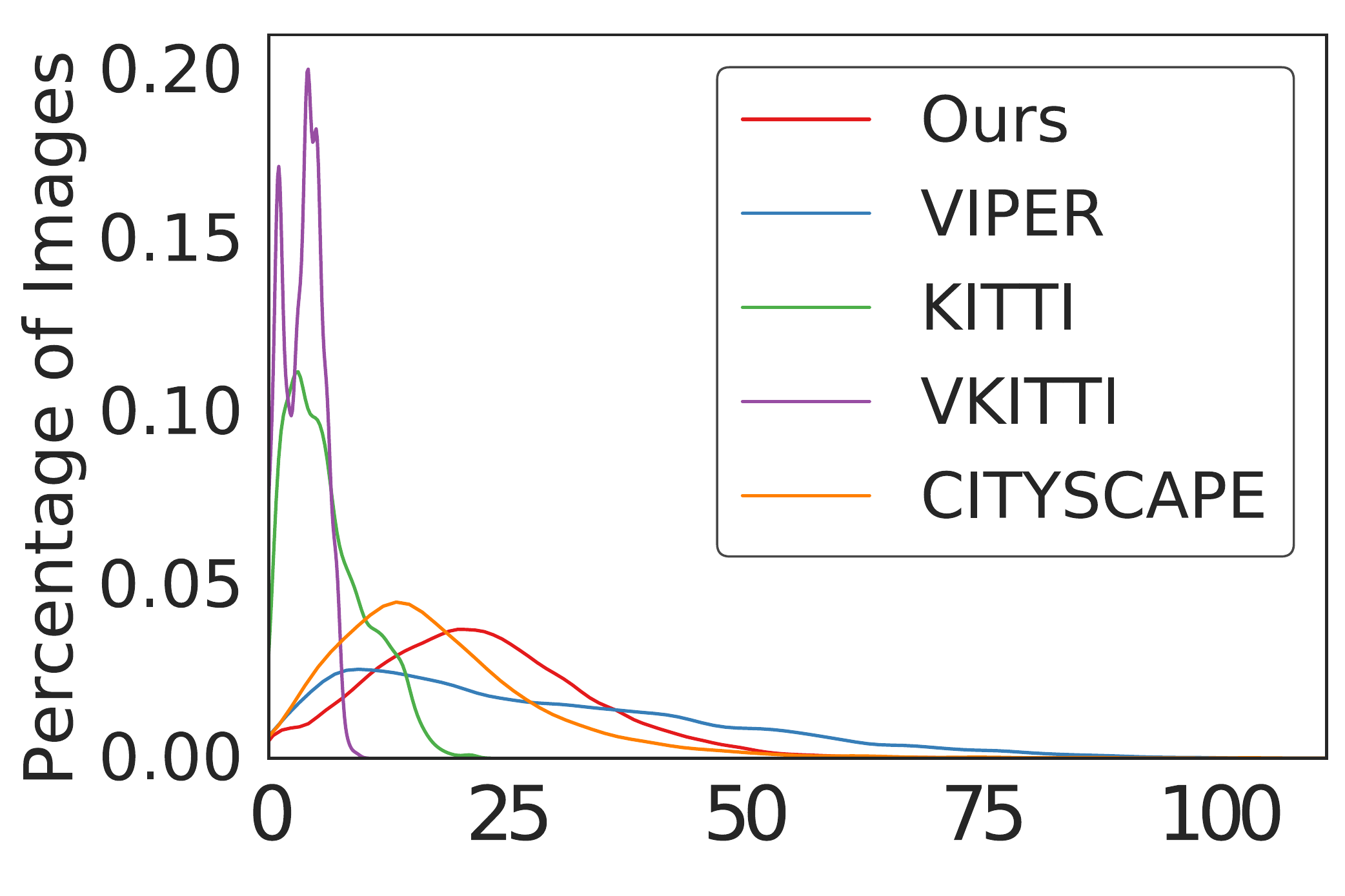}
        \caption{Instances per image}\label{fig:statistics_ins}
    \end{subfigure}
    \begin{subfigure}{0.45\linewidth}
        \includegraphics[width=\linewidth]{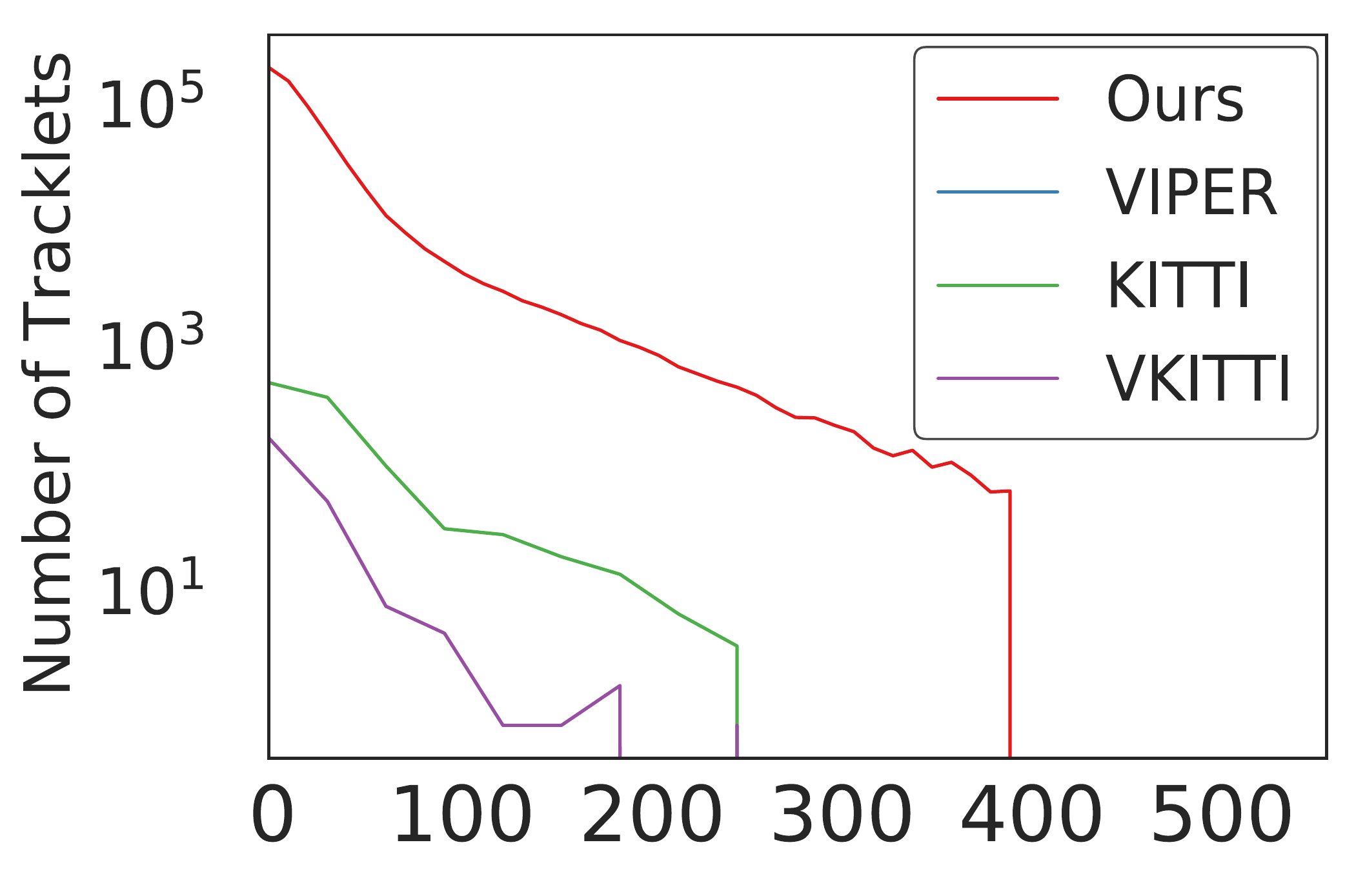}
        \caption{Frames of tracking}\label{fig:statistics_duration}
    \end{subfigure}
    \begin{subfigure}{0.45\linewidth}
        \includegraphics[width=\linewidth]{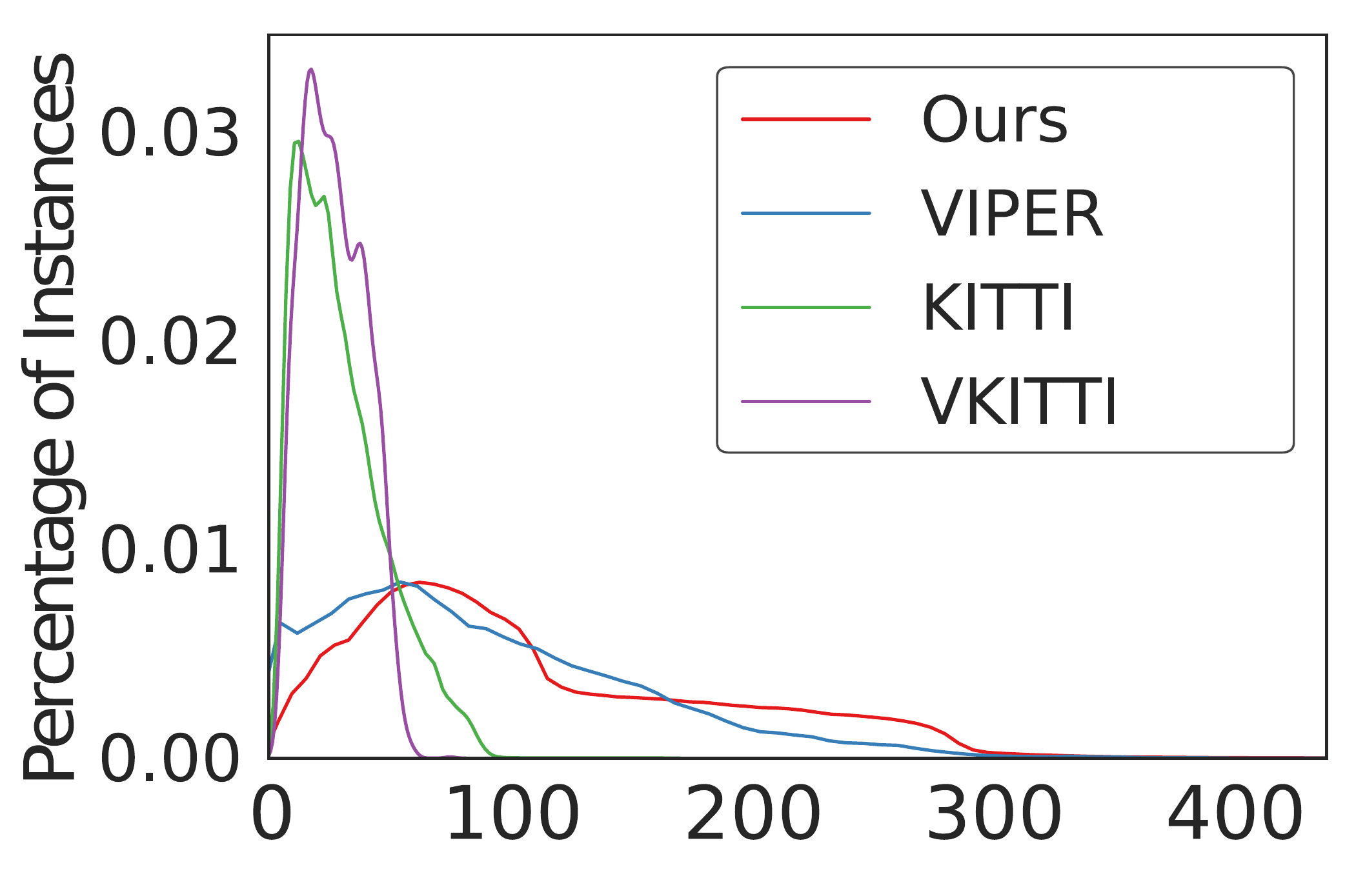}
        \caption{Vehicle distance}\label{fig:statistics_distance}
    \end{subfigure}
    \figcaption{Statistical summary of our dataset}{in comparison of KITTI~\cite{geiger2012kitti}, VKITTI~\cite{gaidon2016virtual}, VIPER~\cite{richter2017playing}, and Cityscapes\cite{cordts2016cityscapes}}
    \label{fig:statistics}
\end{figure}

\minisection{Dataset Statistics.}
Compared to the others, our dataset has more diversity regarding instance scales (\figref{fig:statistics_scale}) and closer instance distribution to real scenes (\figref{fig:statistics_ins}).
To help understand our dataset and its difference, we show more statistics in the appendix.

%% file: 05Experiments.tex
\section{Experiments}
\label{sec:experiments}

We evaluate our 3D detection and tracking pipeline on the real-world driving scenes, \ie, Waymo Open dataset, nuScenes Tracking, KITTI MOT benchmarks, and our large-scale simulation dataset featuring a wide variety of road conditions in a diverse virtual environment.
To follow the submission policy on the above public testing benchmarks, we only evaluate the final performance of our best result on the public testing benchmark.

\subsection{Dataset Details}
\label{sec:dataset_details}

\minisection{GTA 3D Vehicle Tracking dataset} recorded raw data at $12$ FPS, which is helpful for temporal aggregation.
With the goal of autonomous driving in mind, we focus on vehicles closer than $150m$, and also filtered out the bounding boxes whose areas are smaller than $256$ pixels.
The dataset is then split into train, validation and test set with a ratio $10:1:4$.

\minisection{KITTI MOT benchmark~\cite{geiger2012kitti}} provides real-world driving scenarios.
Since the KITTI benchmark is relatively small-scale, we additionally train on the whole KITTI detection training set to enhance our 3D estimation module in the connection of lifting 3D bounding boxes from a single image.
We train a full model with the whole detection and tracking training set for the public benchmark submission.
For bounding box center comparison, we train on the whole detection training set and half of the tracking training set, and then we evaluate the performance difference on the other half of the tracking training set.

\minisection{nuScenes dataset~\cite{caesar2019nuscenes}} is another real-world benchmark containing street scenes in Boston and Singapore captured from a moving vehicle equipped with different sensors, \ie, $6$ cameras, $1$ LiDAR, and $4$ radars.
It provides 3D annotations for LiDAR data with $10$ object classes for detection task, and $7$ object classes for tracking task.
There are $700$ training sequences, $150$ validation sequences and $150$ test sequences in the nuScenes dataset.
Every sequence collects images at $12$ FPS, denoted as \textit{full frames}, and only those sampled keyframes, annotated at $2$ FPS, are used for evaluation.
Each sequence consists of about $40$ keyframes per camera.
We apply our motion-based pipeline on the frames in a higher frame rate (\textit{full frames}) and produce refined tracking results on a keyframe basis.

\minisection{Waymo Open Dataset~\cite{waymo}} collects images at $10Hz$ from $5$ different directions with partially overlapped regions: front, front left, front right, side left, and side right.
There are $1,150$ segments of $20s$ each, $798$ segments for training, $202$ segments for validation, and $150$ segments for testing.
Every segment contains about $198$ frames per camera.
Waymo dataset provides $3$ types of annotations: 2D annotations, 3D LiDAR annotations, and projected 3D LiDAR annotations.
The annotated object classes are vehicles, pedestrians, cyclists, and signs. However, signs are not used in the 3D detection and 3D tracking benchmark.


\minisection{Cross Camera Aggregation.}
For both nuScenes and Waymo, we align all camera results together in an online fashion because their official evaluations only support the LiDAR-based perspective.
The cross-camera aggregation criteria focus on removing duplicated objects across different cameras whose 3D euclidean distance is within $2$ meters. For an object with smaller width or height, such as a pedestrian, the threshold is set to $1$ meter.
Once duplicates are found, we keep the bounding box with a higher 3D score and discard the lower ones to handle the overlapping problem.

On the nuScenes dataset, we found using only camera data achieving competitive results compared to LiDAR baselines since all $6$ cameras can be stitched together to reconstruct the 360-degree scene.
On the other hand, the five camera setting on Waymo Open Dataset lacks information at the rear camera and covers only $5/8$ azimuth angle of the horizon.
But objects are annotated in LiDAR sweeps. 
The limitation above refrains our method from reaching meaningful evaluation scores on Waymo Open 3D detection or tracking benchmark.
Therefore, we conduct a view angle ablation study in~\tabref{tab:waymo_validation_evaluation} to validate the power of our data-driven method.

\subsection{Training and Evaluation}
\label{sec:training_and_evaluation}

\minisection{Network Specification.}
Our 2D detection and 3D center estimation are based (\secref{sec:object_detection}) on Faster RCNN~\cite{ren2015faster} with image classification pre-trained weights on ImageNet~\cite{russakovsky2015imagenet} from TorchVision~\cite{adam2019pytorch}.
For the KITTI dataset with a smaller amount of data, we choose a light-weight yet powerful DLA-34-up~\cite{yu2018dla} architecture, while for the large driving datasets, \eg, nuScenes, GTA, Waymo, we resort to a $101$-layer ResNet~\cite{resnet}.
The anchors of the RPN in Faster R-CNN span $5$ scales and $5$ ratios, in order to detect small objects at distance as well as larger, close objects.
We use a maximum of $2000$ candidate objects before NMS.
The 3D estimation network (\secref{sec:3d_estimation}) estimate 3D bounding box parameters from features extracted from RoI align~\cite{he2017mask}.
For the quasi-dense similarity learning (\secref{sec:quasi_dense_similarity_feature}), we use $\lambda_{embed} = 0.25$ throughout the experiments.
The LSTM motion module (\secref{sec:lstm_motion}) is trained for $100$ epochs on $10$ sample frames per object trajectory with $128$ sequences per batch for large-scale datasets and $8$ for KITTI.

\minisection{Training Procedure.}
The general training procedure is conducted for $24$ epochs using SGD~\cite{robbins1951sgd} with a momentum of $0.9$ and a weight decay of $10^{-4}$.
The learning rate linearly increases from $5\times10^{-4}$ to $5\times10^{-3}$ over the initial $1000$ warm-up training steps.
We use a batch size of $8$ images per GPU and train on $4$ GPUs, resulting in an effective mini-batch size of $32$.
We apply image horizontal flipping as data augmentation in the training phase, and there is no augmentation performed for test/validation phase.

We optimize our model on each dataset with different training procedures according to its amount of data and the GPU memory limit.
The image resolution is resized and padded to fit the minimum down sample scale $32$.
We use $1600\times900$ resolution for nuScenes, $1280\times854$ for Waymo, $1485\times448$ for KITTI, and $1920\times1080$ for GTA.
Following CenterTrack~\cite{zhou2020centertrack}, we fine-tune our KITTI model from a GTA 3D detection model.
Given the multiple cameras setup in nuScenes and Waymo, we treat each camera video as a different sequence during both training and inference.
Since nuScenes doesn't provide 2D annotations, we project the annotated 3D bounding boxes to the image plane to train the 2D branch of our model. 
On the Waymo Open dataset, we use only the 3D LiDAR annotations and project them to the image coordinates as our 2D annotations since the 2D and 3D bounding boxes are annotated independently.
For more details on training strategies, please refer to the appendix.

\minisection{3D Detection Evaluation.}
We evaluate the performance of 3D detection with refined object state estimation of different tracking methods.
For KITTI and GTA, we use the formal evaluation metrics of detection 3D mAP from KITTI~\cite{geiger2012kitti} in the 3D Object Detection Evaluation.
To inspect the precision-recall curve deeper, we also use detection 3D mAP from COCO~\cite{lin2014mscoco} with a range of IoU thresholds from $50$ to $95$ at a step of $5$.
As stated in~\cite{monodis}, we use $40$ recall positions in 3D object detection evaluation instead of the $11$ recall positions proposed in the original Pascal VOC benchmark~\cite{everinham2010voc}.
For nuScenes, we use the benchmark evaluation metrics, called nuScenes detection score (NDS), which is the weighted sum of mean Average Precision (mAP) and several True Positive (TP) metrics.
For Waymo, we also use the 3D detection benchmark evaluation metrics where their primary metric is mean Average Precision weighted by Heading (mAPH) with different 3D IoU thresholds for each class.

\minisection{Multiple Object Tracking Evaluation.}
For KITTI and GTA, we use the official KITTI 2D tracking benchmark metrics following CLEAR~\cite{bernardin2008evaluating} and Li et al.~\cite{li2009learning}, including Multiple Object Tracking Accuracy (MOTA), Multiple Object Tracking Precision (MOTP), Miss-Match (MM), False Positive (FP), False Negative (FN), Mostly-Tracked (MT), Partly-Tracked (PT), and Mostly-Lost (ML), etc.
MOTA indicates the tracking performance of a tracker in dealing with three common errors, \ie, false positives, missed targets and identity switches.
\begin{equation}
        MOTA = 1 - \frac{IDS  + FP + FN}{GT} \label{eqn:MOTA}
\end{equation}
where $GT$ stands for number of ground-truth positives.

For nuScenes, we follow the 3D tracking benchmark evaluation metrics using Average Multi-Object Tracking Accuracy (AMOTA) as the main metric. They average the MOTA metric over different recall thresholds
\begin{align}
    \begin{split}
        AMOTA &= \frac{1}{n-1} \sum_{r \in \{ \frac{1}{n-1}, \frac{2}{n-1}, ..., 1 \} } {MOTA}_{\subfix{r}}, \label{eqn:AMOTA} \\
        {MOTA}_{\subfix{r}} &= \text{max} (0, \\
         & 1 - \mathbf{\alpha} \frac{IDS_{\subfix{r}} + FP_{\subfix{r}} + FN_{\subfix{r}} - (1 - r){GT}}{r {GT}})
    \end{split}
\end{align}
with recall threshold $\mathbf{r}$ is calculated at each of the interpolation $\mathbf{n}$-points for $n = 40$, tolerance ratio $\mathbf{\alpha}$ aims to retain the non-zero values of MOTA$_{\subfix{r}}$.
We report $\mathbf{\alpha} = 1.0$, denoted as AMOTA@1, for the official benchmark and $\mathbf{\alpha} = 0.2$, denoted as AMOTA@0.2, for backward comparison with previous arts. 
Instead of using 3D IoU threshold, the nuScenes tracking benchmark choose a centroid distance threshold of $2$ meters on the ground plane as the matching criterion.

For Waymo, we report the official 3D tracking evaluation metrics, which follow the metrics of CLEAR~\cite{bernardin2008evaluating} with different 3D IoU thresholds for each class.
However, the limitations above (\secref{sec:dataset_details}) refrain our method from reaching meaningful evaluation scores on benchmarks using 3D IoU based metrics.

\minisection{Discussion of Evaluation Metrics}
The analysis from~\cite{caesar2019nuscenes} has shown that objects, \ie, pedestrians, bicycles, are hard for camera-based methods to have meaningful IoU scores, while those objects are difficult to be detected in LiDAR.
Similarly, we observe from quantitative results on Waymo that the IoU threshold at $0.7$ prohibits the camera-based method from evaluating meaningful tracking performance. 
However, the score does not cohere to our qualitative results.
We argue that high IoU thresholds become impractical with uncertain object state $\mathbf{s}_{a}$ measurements as the metric does not reward a tracker for associating all matches right if it makes a moderate error on the object state.

Therefore, we report extended 3D tracking evaluation~\cite{Chang2019argoverse} for a better understanding of the association measure in the 3D extent. The MOTA$_\subfix{3D}$ is based on the object centroid distance, denoted as MOTP$_\subfix{C}$, from tracked objects to the ground truth within a detection association range, \ie, $2$ meters. MOTP$_\subfix{O}$ calculates orientation angle difference to the vertical axis, and MOTP$_\subfix{I}$ stands for the amodal shape error, using $1 - \text{IoU}$ with aligned orientation and centroid.

We conduct ablation studies in~\tabref{tab:cross_dataset_centroid_based_evaluation} with detailed metrics on translation, scale, orientation as accompanying metrics to validate the power of our data-driven method.


\subsection{Ablation Experiments}
\label{sec:ablation_experiments}
In the following, we discuss the results of our ablation experiments on the design choices of each component (3D box estimation, data association, motion modeling) in our quasi-dense 3D tracking pipeline.
We report the 3D tracking performance comparison on nuScenes validation set for dropping sub-affinity scores in \tabref{tab:nusc_ablation_drop_affinity}, using full images in \tabref{tab:nusc_ablation_full_frame}, and swapping different modules in \tabref{tab:nusc_ablation_swap_module}.
We also investigate the importance of using a 3D object center using the KITTI sub-val set in~\tabref{tab:kitti_projection_3d_center}.

\minisection{Importance of each sub-affinity matrix.}
From~\tabref{tab:nusc_ablation_drop_affinity}, we observe that adding a deep feature affinity ($\mathbf{A}_{\subfix{deep}}$) distinguishes two near-overlapping objects and thus increasing AMOTA@1 with an observable margin. Adding 3D object state affinity ($\mathbf{A}_{\subfix{iou}}$) also contributes to AMOTA@1 effectively. Lastly, we found in~\tabref{tab:nusc_ablation_full_frame} that applying our model on \textit{full frames} during inference helps the motion model estimate an objects' velocity more accurately. Motion-based methods benefit from using a higher frame rate setting, \textit{full frames}, with a $(0.2421-0.2329)/0.2421 = 3.8\%$ improvement on LSTM and $0.6\%$ on KF3D.

\begin{table}[thpb]
    \figcaption{Ablation study of tracking performance with different methods on nuScenes validation set.}{Results in (a) suggest that the deep feature affinity matrix contributes to data association. From (b), we observe a robust trend that a motion model with a higher frame rate image sequence encourages tracking performance.}
    \begin{subtable}[t]{\linewidth}
    \centering
    \small
    \caption{Ablation study of dropping affinity matrix.}
    \adjustbox{width=\linewidth}{
        \begin{tabular}{c|cccc|rr}
            \toprule
            {Method} & $\mathbf{A}_{\subfix{deep}}$ & $\mathbf{A}_{\subfix{iou}}$ & $\mathbf{A}_{\subfix{depth}}$ & $\hat{\mathbf{c}}_{depth}$ & {AMOTA@1} $\uparrow$ & {AMOTP} $\downarrow$ \\
            \midrule
            \multirow{5}{*}{KF3D}
            & - & \checkmark & \checkmark & \checkmark & 0.1779 & 1.596 \\
            & \checkmark & - & \checkmark & \checkmark & 0.2013 & 1.540 \\
            & \checkmark & \checkmark & - & \checkmark & 0.2213 & \textbf{1.535} \\
            & \checkmark & \checkmark & \checkmark & - & 0.2301 & \textbf{1.535} \\
            & \checkmark & \checkmark & \checkmark & \checkmark & \textbf{0.2306} & \textbf{1.535} \\
            \bottomrule
        \end{tabular}
    }
    \label{tab:nusc_ablation_drop_affinity}
    \end{subtable}

    \begin{subtable}[t]{\linewidth}
    \centering
    \small
    \medskip 
    \caption{Ablation study of \textit{full frames}.}
    \adjustbox{width=0.9\linewidth}{
        \begin{tabular}{c|c|rr}
            \toprule
            {Method} & {full frames} & {AMOTA@1} $\uparrow$ & {AMOTP} $\downarrow$ \\
            \midrule
            \multirow{2}{*}{KF3D}
            & - & 0.2306 & 1.535 \\
            & \checkmark & \textbf{0.2321} & \textbf{1.530} \\
            \midrule
            \multirow{2}{*}{VeloLSTM}
            & - & 0.2329 & 1.528 \\
            & \checkmark & \textbf{0.2421} & \textbf{1.518} \\
            \bottomrule
        \end{tabular}
    }
    \label{tab:nusc_ablation_full_frame}
    \end{subtable}
\end{table}

\minisection{Importance of sub-affinity module design.}
Comparing boxes IoU sub-affinity ($\mathbf{A}_{\subfix{iou}}$), adding depth-order matching, denoted \textit{3D} in~\tabref{tab:nusc_ablation_swap_module}, as overlapping threshold helps identify falsely rejected objects and increases overall AMOTA@1 scores by a relative $(0.2306-0.1381)/0.2306 = 40.1\%$ AMOTA@1.
With motion-aware association, denoted \textit{motion} in~\tabref{tab:nusc_ablation_swap_module}, our full model filtering out objects with wrong motion velocity reaches $0.2306$ AMOTA@1 comparing to $0.2234$ using only \textit{cosine} angle difference module or $0.2300$ using \textit{centroid} distance module.

\begin{table}[thpb]
    \centering
    \small
    \figcaption{Tracking performance comparison with different designs of affinity matrix on nuScenes validation set.}{Results suggest that using 3D IoU and motion-based sub-affinity matrix yielding the best validation performance of our final pipeline.}
    \adjustbox{width=0.80\linewidth}{
        \begin{tabular}{ll|rr}
            \toprule
            $\mathbf{A}_{\subfix{iou}}$ & $\mathbf{A}_{\subfix{depth}}$ & {AMOTA@1} $\uparrow$ & {AMOTP} $\downarrow$ \\
            \midrule
            2D & \multirow{3}{*}{motion} & 0.1381 & 1.674 \\
            BEV & & 0.1929 & 1.599 \\
            3D & & \textbf{0.2306} & \textbf{1.535} \\
            \midrule
            \multirow{3}{*}{3D} & cosine & 0.2234 & \textbf{1.534} \\
            & centroid & 0.2300 & 1.536 \\
            & motion & \textbf{0.2306} & 1.535 \\
            \bottomrule
        \end{tabular}
    }
    \label{tab:nusc_ablation_swap_module}
\end{table}


\minisection{Importance of 3D center projection estimation.}
We estimate the 3D location of a bounding box by predicting the projection of its center and depth, while Mousavian~\etal~\cite{mousavian2017deep3dbox} uses the center of detected 2D boxes directly.
\tabref{tab:kitti_projection_3d_center} shows the comparison of these two methods on the KITTI dataset.
The result indicates that the correct 3D projections provide a higher tracking capacity for a motion module to associate candidates and significantly reduce the centroid misalignment (MOTP$_\subfix{C}$).

\begin{table}[htpb]
    \figcaption{Importance of using projection of 3D bounding box center estimation on KITTI sub-validation set.}{
        We evaluate our KF3D model using different center inputs $C$ on the Car category to reveal the importance of estimating the projection of a 3D center.
        The increase of MOTA and higher 3D IoU AP with COCO $(50:5:95)$ suggest that the projection of a 3D center benefits our tracking pipeline over the 2D center.
    }
    \centering
    \adjustbox{width=\linewidth}{
        \begin{tabular}{l|lrr||lrr}
            \toprule
            Method         & Range  & MOTA $\uparrow$ & MOTP$_\subfix{C}$ $\downarrow$ & Difficulty & $AP_\subfix{bev}^\subfix{COCO}$ $\uparrow$ & $AP_\subfix{3d}^\subfix{COCO}$ $\uparrow$ \\
            \midrule
            \multirow{3}{*}{2D Cen}
                           & 0-30m  & 77.28 & 0.56  & Easy   & 24.91 & 15.11  \\
                           & 0-50m  & 70.78 & 0.60  & Medium & 23.53 & 15.87  \\
                           & 0-100m & 65.15 & 0.62  & Hard   & 21.99 & 14.99  \\
            \midrule
            \multirow{3}{*}{3D Cen}
                           & 0-30m  & \textbf{79.50} & \textbf{0.44}  & Easy   & \textbf{41.71} & \textbf{36.74} \\
                           & 0-50m  & \textbf{72.53} & \textbf{0.51}  & Medium & \textbf{33.73} & \textbf{29.30} \\
                           & 0-100m & \textbf{66.77} & \textbf{0.54}  & Hard   & \textbf{31.05} & \textbf{26.67} \\
            \bottomrule
        \end{tabular}
    }
    \label{tab:kitti_projection_3d_center}
\end{table}


\subsection{Motion Modeling Comparison.}
\label{sec:motion_modeling_comparison}
For motion-based models, update-and-predict cycles help single-frame 3D estimates.
To examine the effectiveness of our VeloLSTM, we compare variants of motion models to pure detection in refining the object locations along their trajectories.

\minisection{Pure Detection (Detection).}
A single-frame monocular 3D estimate provides a noisy measurement of an object learned from the data distribution.

\minisection{Dummy Motion Model (Momentum).}
Comparing with the non-linear motion model, we experimented with a dummy motion model that directly updated the object state with a fixed momentum on observed measurements and predicted the next possible state with zero motion.
By applying motion momentum, we expect the dummy model to output linearly smoothed trajectories.

\minisection{Kalman Filter 3D baseline (KF3D).}
We take a Kalman filter-based motion model as our baseline which models $$\{x, y, z, \theta, l, w, h, \Delta x, \Delta y, \Delta z\}$$ in 3D world coordinates.
The well-tuned Kalman filter~\cite{kalman1960new} models a sequence of observed measurements with Gaussian noise assumptions to find a prediction location that minimize a mean of square error.

\minisection{Deep Motion Estimation and Update (VeloLSTM).}
As mentioned in~\secref{sec:lstm_motion}, we propose a VeloLSTM network to model the vehicle motion.
To analyze the effectiveness of our deep motion model, we compare our LSTM model with traditional 3D Kalman filter (KF3D) and single-frame 3D estimation (Detect) using 3D detection and tracking evaluation metrics from nuScenes.
\tabref{tab:nusc_motion_helps_3dap} shows that KF3D provides a small improvement in NDS via trajectory smoothing within prediction and observation.
On the other hand, our VeloLSTM module provides a learned estimation based on past $n=5$ velocity predictions and current frame observation, which may compensate for the observation error.
Our VeloLSTM module achieves the highest AMOTA@1, AMOTP, and NDS, among the other methods.

\begin{table}[htpb]
    \centering
    \figcaption{Comparison of location refinement with different motion model on nuScenes validation set.}{Result suggests the VeloLSTM achieves the highest performance among the other methods. Besides, employing a motion model benefits 3D detection and tracking. We use \textit{full frames} and final affinity matrices.}
    \small
    \adjustbox{width=0.9\linewidth}{
        \begin{tabular}{l|rrr}
            \toprule
            {Method} & {NDS} & {AMOTA@1} $\uparrow$ & {AMOTP} $\downarrow$ \\
            \midrule
            Detection & 0.3622 & 0.222 & 1.538 \\
            Momentum & 0.3620 & 0.227 & 1.532 \\
            KF3D & 0.3634 & 0.232 & 1.530 \\
            VeloLSTM & \textbf{0.3666} & \textbf{0.242} & \textbf{1.518} \\
            \bottomrule
        \end{tabular}
    }
    \label{tab:nusc_motion_helps_3dap}
\end{table}


\subsection{Real-world Evaluation.}
\label{sec:real_world_evaluation}
\begin{figure*}[htpb]
    \adjustbox{width=1.0\linewidth}{
        \setlength{\tabcolsep}{2pt}
        \begin{tabular}{ccc}
            \includegraphics[width=0.33\textwidth]{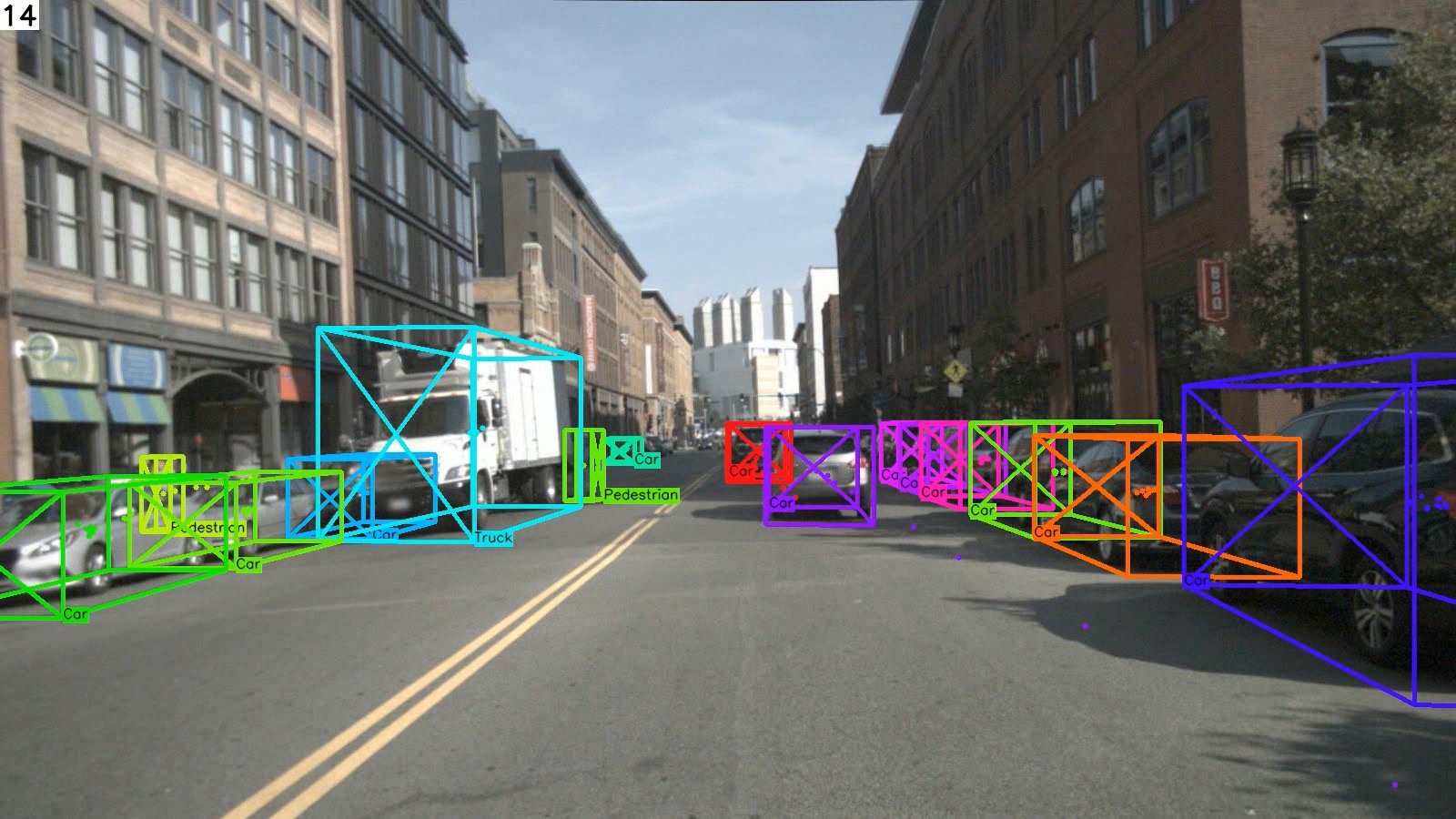} &
            \includegraphics[width=0.33\textwidth]{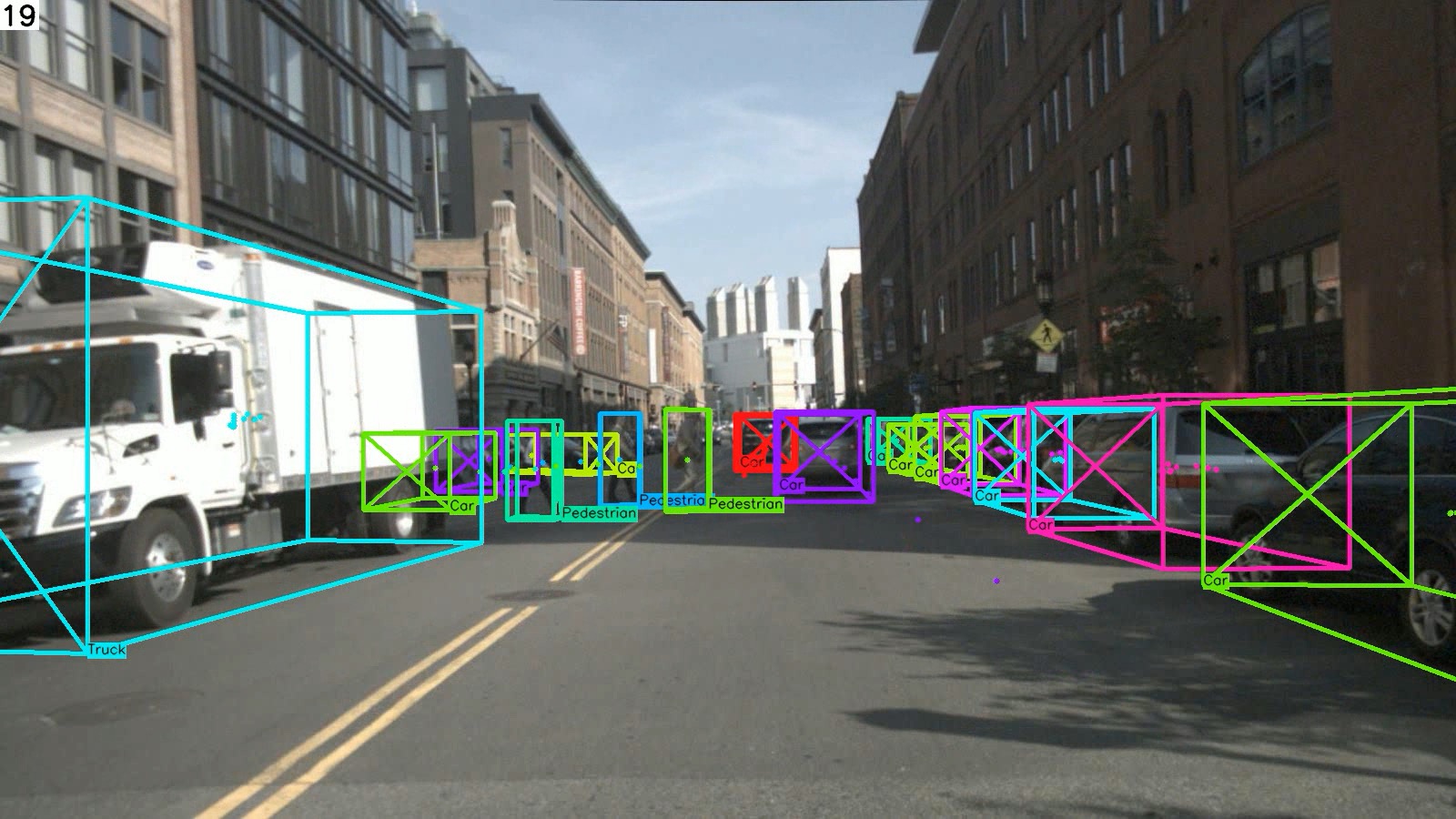} &
            \includegraphics[width=0.33\textwidth]{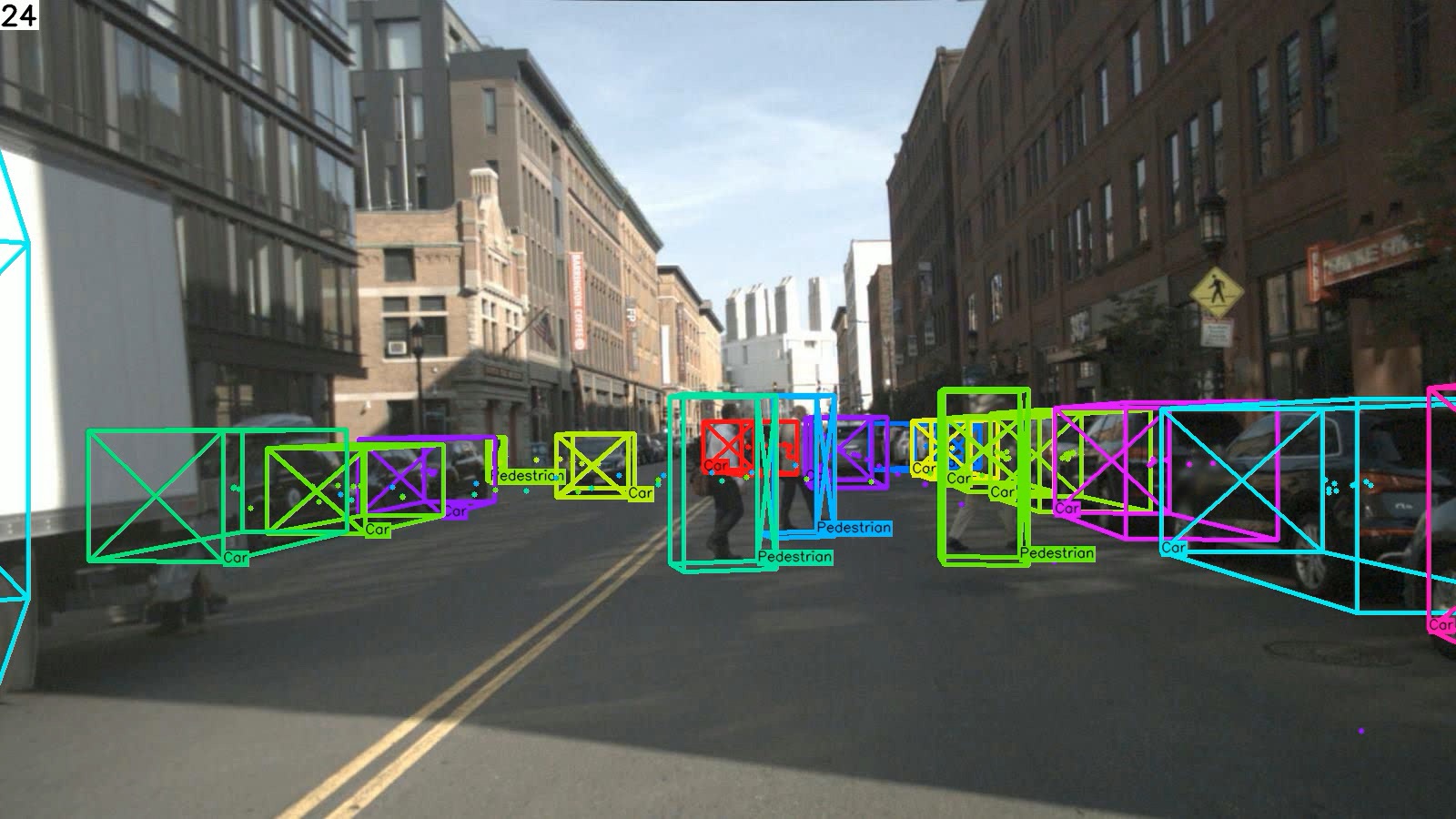} \\

            \includegraphics[width=0.33\textwidth]{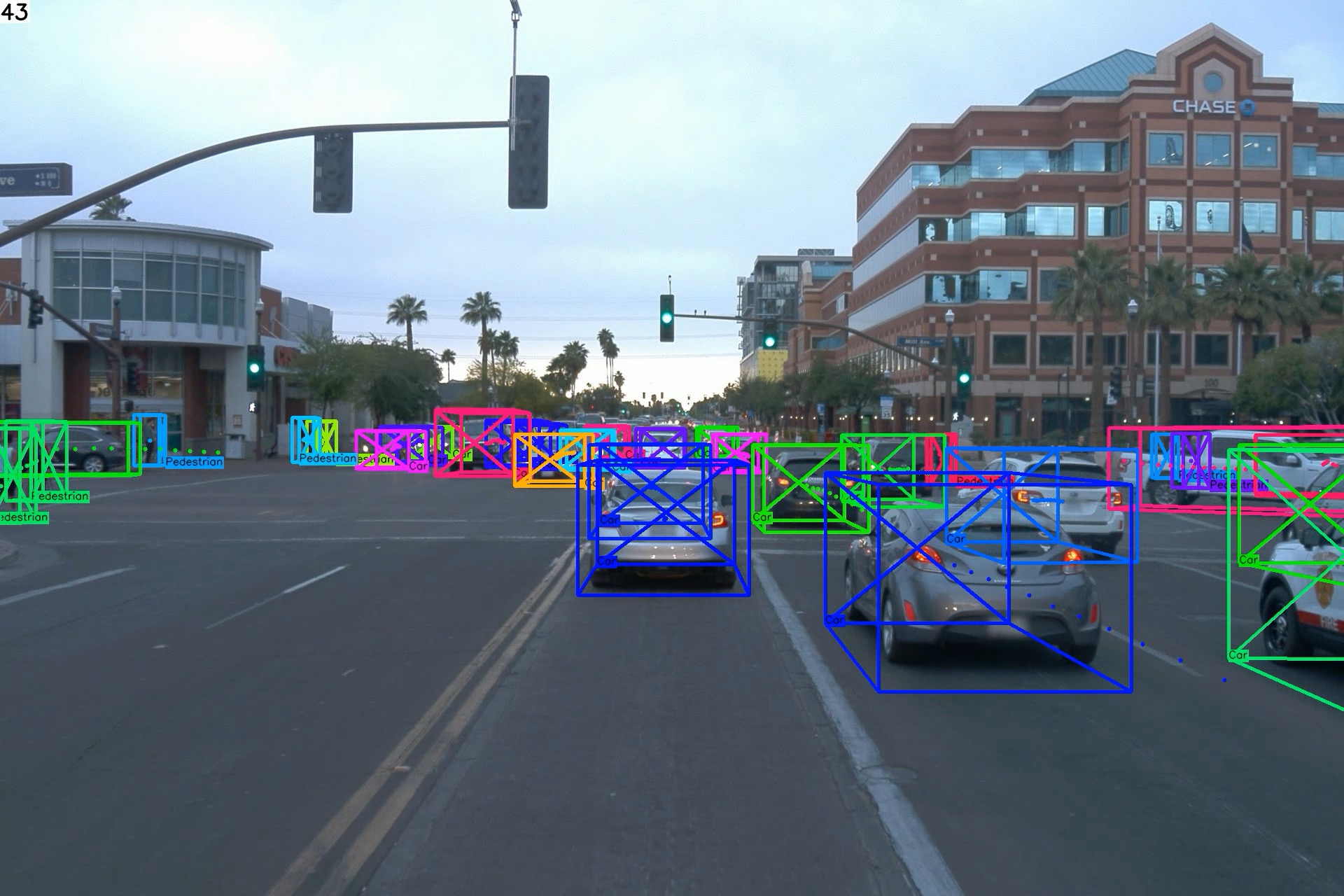} &
            \includegraphics[width=0.33\textwidth]{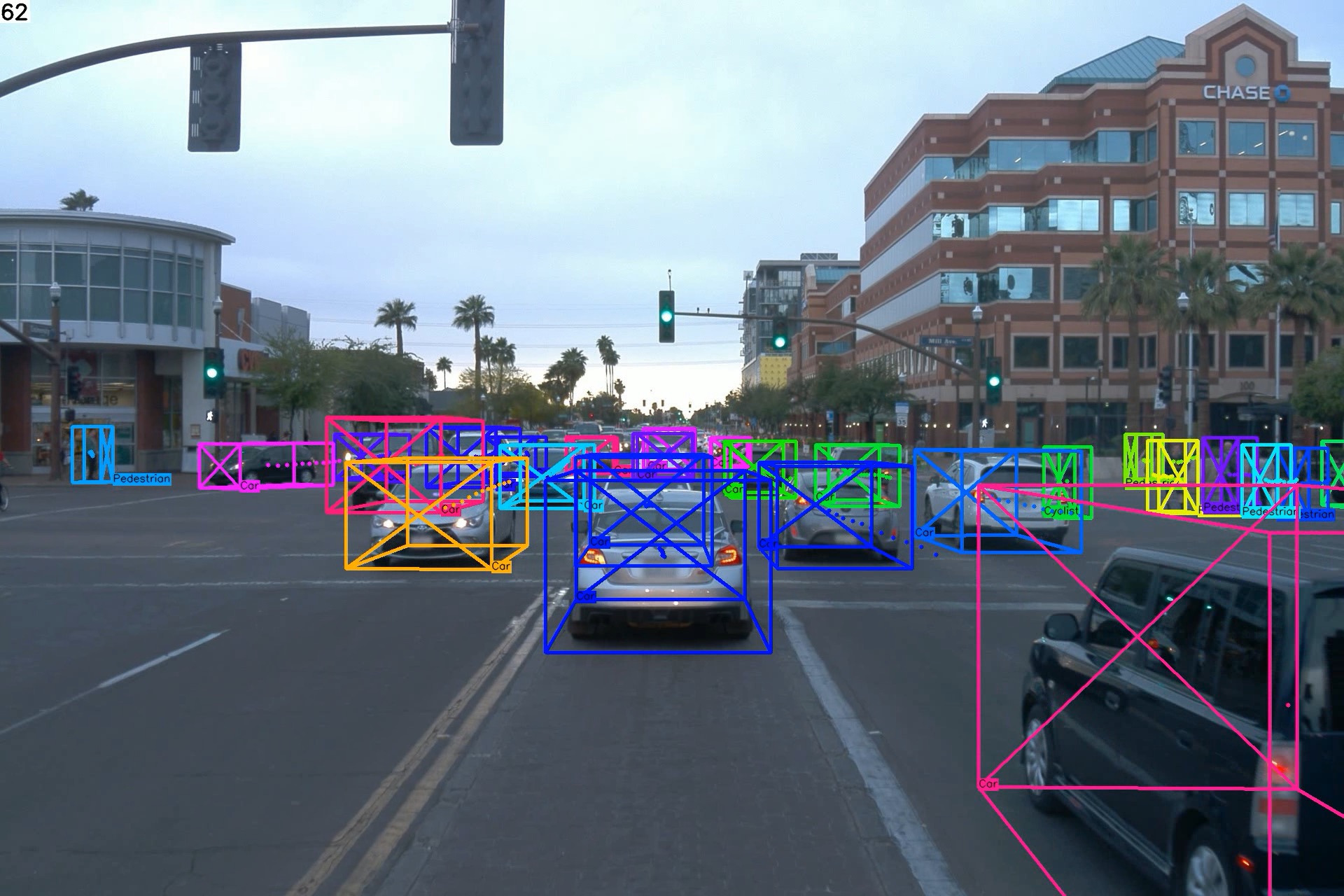} &
            \includegraphics[width=0.33\textwidth]{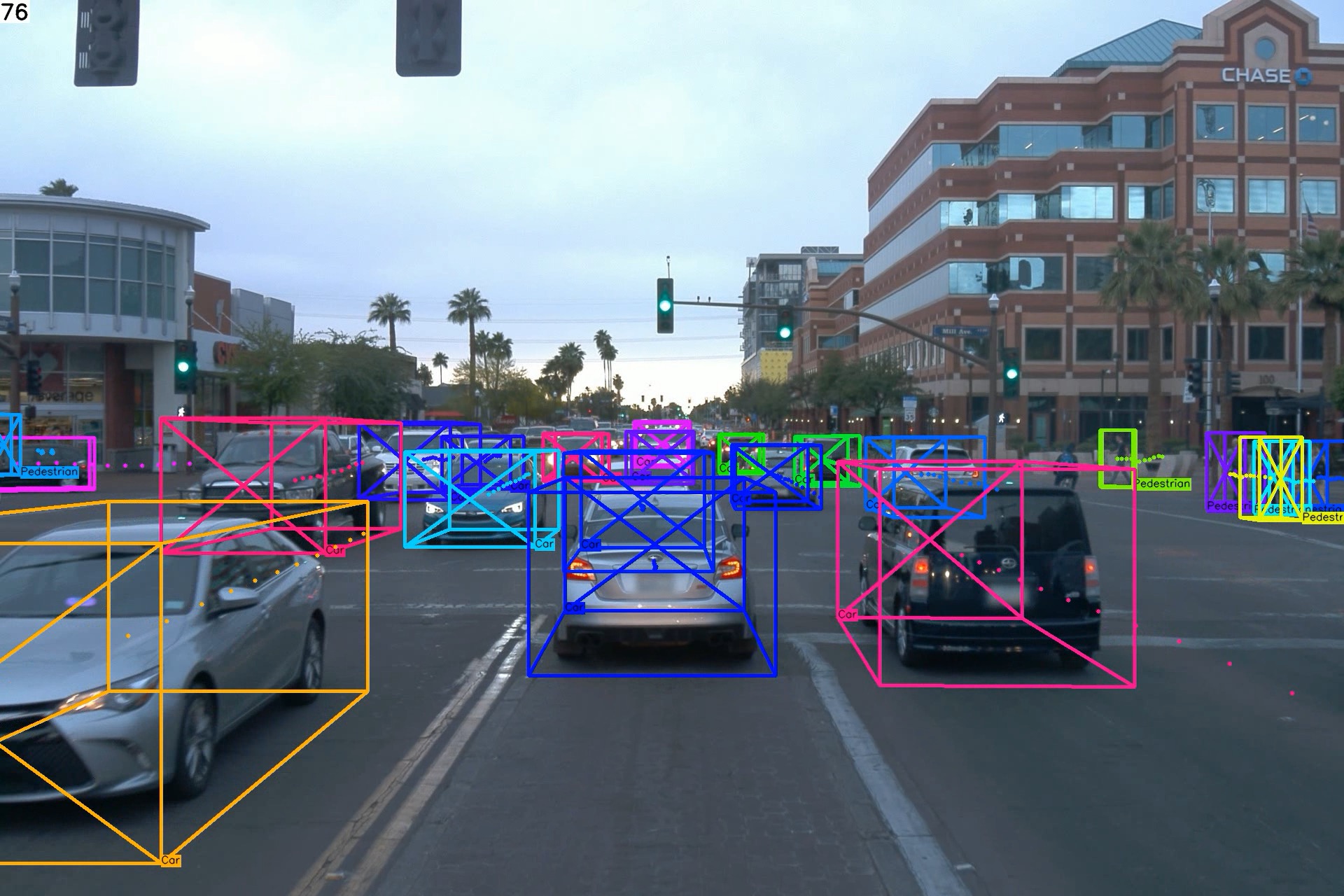}
        \end{tabular}
    }
    \figcaption{Qualitative results on testing set of nuScenes (First row), Waymo Open (Second row) datasets.}{Our proposed quasi-dense 3D tracking pipeline estimates accurate 3D extent and robustly associates tracking trajectories from a monocular image. We show predicted 3D bounding boxes and trajectories colored with tracking IDs.
    }
    \label{fig:qualitative_result}
\end{figure*}

Besides evaluating on large-scale synthetic data, we utilize nuScenes~\cite{caesar2019nuscenes}, Waymo Open~\cite{waymo} and KITTI~\cite{geiger2012kitti} benchmarks to compare our model abilities with other state of the arts.
Major results are listed in \tabref{tab:nusc_tracking_test} for nuScenes dataset and \tabref{tab:waymo_tracking_test} and \tabref{tab:waymo_detection_test} for Waymo Open dataset.
Qualitative results on all three datasets are shown in~\figref{fig:qualitative_result}.

\begin{figure}[htpb]
    \adjustbox{width=1.0\linewidth}{
        \setlength{\tabcolsep}{2pt}
        \begin{tabular}{c}
        \includegraphics[width=\textwidth]{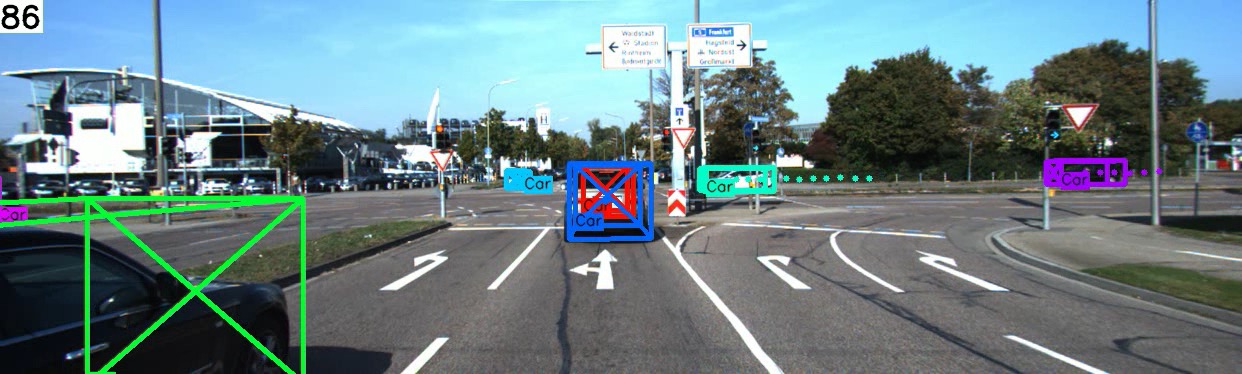} \\
        \includegraphics[width=\textwidth]{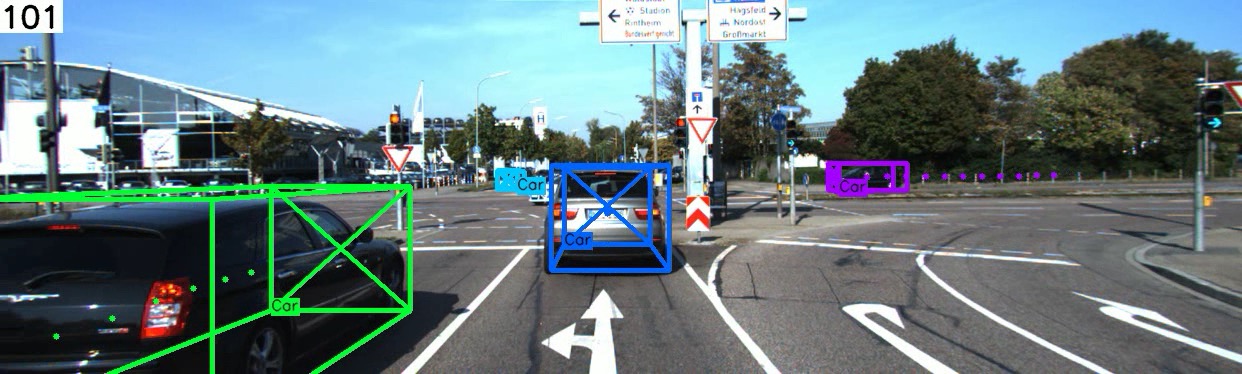} \\
        \includegraphics[width=\textwidth]{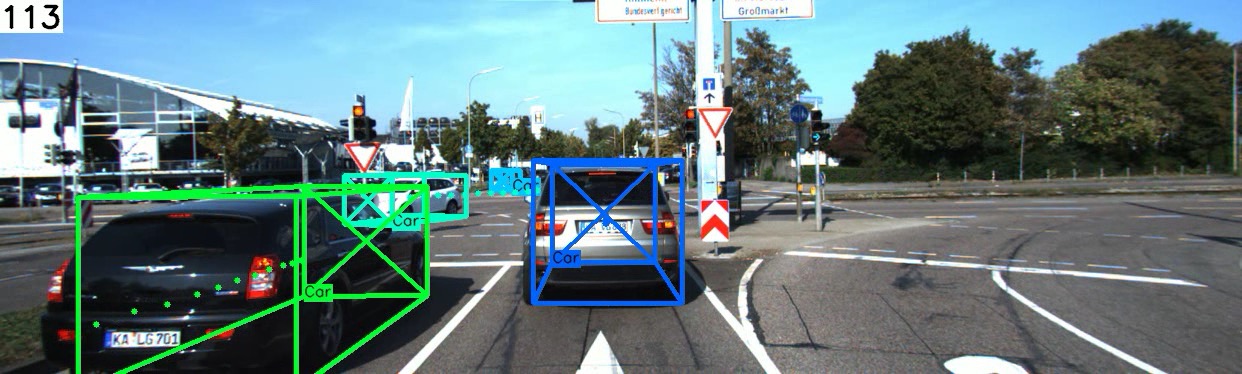} \\
        \includegraphics[width=\textwidth]{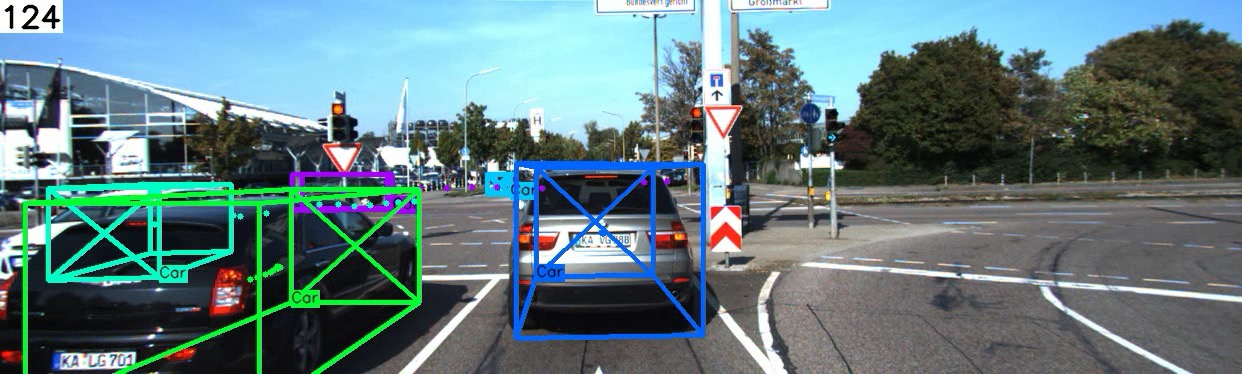}
        \end{tabular}
    }
    \figcaption{Qualitative results on testing set of KITTI datasets.}{Our VeloLSTM continues to predict the 3D object state of unmatched tracklets after they disappear. With motion-based data association, our tracker successfully recovers tracked vehicle (cyan-colored box) from object occlusions after the vehicle re-appeared. We show predicted 3D bounding boxes and trajectories colored with tracking IDs.
    }
    \label{fig:qualitative_result_kitti}
\end{figure}

\minisection{nuScenes tracking challenge.}
Our monocular 3D tracking method outperforms all the published methods with a large margin.
Center-Track-Vision~\cite{zhou2020centertrack} uses two consecutive frames to generate inter-frame motion for object detection and 3D tracking.
Center-Track-Open~\cite{zhou2020centertrack} fuses LiDAR information in the Center-Track-Vision pipeline with Megvii-detector~\cite{zhu2019cbgs} to generate 3D detection.
The LiDAR-based baselines~\cite{caesar2019nuscenes} uses state-of-the-art LiDAR-based detectors~\cite{shi2018pointrcnn, lang2019pointpillars, zhu2019cbgs} to estimate accurate bounding boxes and feed into a Kalman-Filter-based 3D tracker, AB3DMOT~\cite{weng2020ab3dmot}.
Our method takes only one image for 3D object detection and tracking and leverages our deep motion model to refine object trajectories compared to the prior arts.
The nuScenes tracking dataset linearly interpolates GT tracks to avoid track fragments from LiDAR point filtering and removes GT objects without LiDAR points. Both invisible objects with annotation and visible objects without annotation prohibit the camera-based methods from optimizing bounding box estimation.
Nevertheless, our quasi-dense 3D tracking approach reaches $0.217$ AMOTA with near five times tracking accuracy of the best vision-only submission among all published methods.

\begin{table}[htpb]
    \centering
    \figcaption{Tracking performance on the testing set of nuScenes tracking benchmark~\cite{caesar2019nuscenes}.}{We report the average AMOTA@1 and AMOTP over $7$ categories on the benchmark with only published methods shown. Our quasi-dense 3D tracking pipeline outperforms the best camera-based submission (underlined) by near $500\%$ while bridging the gap to LiDAR-based methods on the nuScenes 3D tracking benchmark.
    }
    \adjustbox{width=\linewidth}{
        \begin{tabular}{ll|rr}
            \toprule
            Method               & Modality       & AMOTA@1 $\uparrow$ & AMOTP $\downarrow$ \\
            \midrule
            \textbf{QD3DT (Ours)}        & Camera         & \textbf{0.217}       & 1.550 \\
            Megvii-AB3DMOT~\cite{caesar2019nuscenes}  & LiDAR          & 0.151                & 1.501 \\
            CenterTrack-Open~\cite{zhou2020centertrack} & LiDAR + Camera & 0.108                & \textbf{0.989} \\
            CenterTrack-Vision~\cite{zhou2020centertrack} & Camera         & \underline{0.046}                & 1.543 \\
            PointPillars-AB3DMOT~\cite{caesar2019nuscenes}  & LiDAR          & 0.029                & 1.703 \\
            Mapillary-AB3DMOT~\cite{caesar2019nuscenes}  & LiDAR          & 0.018                & 1.790 \\
            \bottomrule
        \end{tabular}
    }
    \label{tab:nusc_tracking_test}
\end{table}

\minisection{Waymo Open benchmark.}
We provide a strong baseline being the first entry in camera-only modality for both 3D tracking and detection challenges shown in~\tabref{tab:waymo_tracking_test} and~\ref{tab:waymo_detection_test}.


\begin{table}[htpb]
    \begin{subtable}[t]{\textwidth}
        \centering
        \caption{3D Tracking Performance}
        \adjustbox{width=0.9\linewidth}{
            \begin{tabular}{l|l|rr}
                \toprule
                {Method} & {Modality} & {MOTA} $\uparrow$ & {MOTP} $\downarrow$ \\
                \midrule
                HorizonMOT3D~\cite{wang20201st} & Lidar+Camera & 0.6407 & 0.1577 \\
                CenterPoint~\cite{yin2020center} & Lidar & 0.5938 & 0.1637 \\
                pillars\_kf\_baseline~\cite{waymo} & Lidar & 0.4008 & 0.1856 \\
                \midrule
                QD3DT (Ours) & Camera & 0.0001 & 0.0658 \\
                \bottomrule
            \end{tabular}
        }
        \label{tab:waymo_tracking_test}
    \end{subtable}

    \begin{subtable}[t]{\textwidth}
        \centering
        \medskip 
        \caption{3D Detection Performance.}
        \adjustbox{width=0.9\linewidth}{
            \begin{tabular}{l|l|rr}
                \toprule
                {Method} & {Modality} & {mAPH} $\uparrow$ & {mAP} $\uparrow$ \\
                \midrule
                HorizonMOT3D~\cite{wang20201st} & Lidar+Camera & 0.7783 & 0.7823 \\
                PV-RCNN~\cite{openpcdet2020} & Lidar & 0.7323 & 0.7369 \\
                CenterPoint~\cite{yin2020center} & Lidar & 0.7299 & 0.7342 \\
                \midrule
                QD3DT (Ours) & Camera & 0.0233 & 0.0242 \\
                \bottomrule
            \end{tabular}
        }
        \label{tab:waymo_detection_test}
    \end{subtable}
    \figcaption{3D detection and tracking performance in Vehicle Level 2 difficulty on the testing set of Waymo dataset.}{Our pipeline serves as a strong baseline with non-zero scores using only camera modality in the 3D detection and tracking challenge on LiDAR-dominated benchmark.}
\end{table}

\begin{table}[htpb]
    \centering
    \figcaption{3D detection and tracking performance in Vehicle Level 2 difficulty on the validation set of Waymo dataset.}{We compare 3D detection and tracking results on different extent of the observable area. LiDAR-based GT covers a 360-degree area, while camera-based GT filters out annotations outside of camera view. MOTA and mAPH increase greatly while lowering IoU thresholds slightly.}
    \small
    \adjustbox{width=0.9\linewidth}{
        \begin{tabular}{l|l|c|rr}
            \toprule
            {Modality} & {Range} & {IoU} & {MOTA} $\uparrow$ & {mAPH} $\uparrow$ \\
            \midrule
            \multirow{3}{*}{LiDAR-based GT} &
            \multirow{3}{*}{Overall} & 0.3 & 0.1183 & 0.2242 \\
            & & 0.5 & 0.0195 & 0.1161 \\
            & & 0.7 & 1.8e-06 & 0.0286 \\
            \midrule
            \multirow{3}{*}{Camera-based GT} &
            \multirow{3}{*}{Overall} & 0.3 & 0.1867 & 0.3401 \\
            & & 0.5 & 0.0308 & 0.1743 \\
            & & 0.7 & 2.8e-06 & 0.0408 \\
            \bottomrule
        \end{tabular}
    }
    \label{tab:waymo_validation_evaluation}
\end{table}


\subsection{Evaluation Metrics.}
\label{sec:evaluation_metrics}
We observe from~\tabref{tab:waymo_tracking_test} that the IoU threshold at $0.7$ prohibits the camera-based method from evaluating meaningful tracking performance. 
However, the score does not cohere to our qualitative results in~\figref{fig:qualitative_result}.
Therefore, we conduct ablation studies to validate the power of our data-driven method.
We first compare 3D detection and tracking results on different extent of observable area,~\ie, for the annotations from the whole field of view, denoted \textit{LiDAR-based GT}, and ones from full camera field of view, denoted \textit{Camera-based GT}.
The \textit{LiDAR-based GT} setting constantly yields inferior performance with approximate a $5/8$ to the \textit{Camera-based GT} ones in different IoU thresholds in~\tabref{tab:waymo_validation_evaluation}.
The difference matches the ratio of an observable area by cameras and area by a LiDAR.

We turn to centroid-based metrics to validate our pipeline's performance since the IoU-based metrics set up an unreachable bar for the camera-based method.
With centroid-based metrics, the tracking scores of our final model on nuScenes and Waymo Open datasets list in~\tabref{tab:cross_dataset_centroid_based_evaluation}.
The similarity in MOTA and MOTP performance suggests that our pipeline has enough capacity to handle different datasets with minimal change of training procedure, underlining the robustness of our tracking pipeline.

From~\tabref{tab:cross_dataset_centroid_based_evaluation} and~\tabref{tab:waymo_validation_evaluation}, our method acts comparable in IoU-based metrics with a threshold at $0.3$ to centroid-based metrics with a threshold at $2$ meters on the camera-based GT setting.
We argue that the IoU threshold at $0.7$ is overly biased toward a precise 3D object attributes estimation, like depth estimation, 3D center, orientation, 3D center or dimensions.
A trajectory with perfect matching but a moderate error in one of the object attributes, \eg, $1$m off in depth, is not counted as a true positive. However, it is treated equally as those with a completely wrong estimation \eg, $10$m away.
Besides, the experiment in~\tabref{tab:cross_dataset_centroid_based_evaluation} and~\tabref{tab:waymo_validation_evaluation} show an interesting discrepancy in reported performance by different types of evaluation metrics, which supports analysis in~\cite{caesar2019nuscenes} to choose centroid-based evaluation metrics as a fair comparison to both camera-based and LiDAR-based methods.

\begin{table}[htpb]
    \figcaption{Centroid-based evaluation across datasets.}{
        This experiment uses our final model and the \textit{full frames} setting for nuScenes.
        We evaluate the results in different datasets to reveal the importance of choosing evaluation metrics. The similarity of MOTA and MOTP performance suggests that our pipeline has been well trained and performs equally well on the Waymo Open dataset.
    }
    \centering
    \adjustbox{width=\linewidth}{
        \begin{tabular}{l|lrrrr}
            \toprule
            Method         & Range  & MOTA $\uparrow$ & MOTP$_\subfix{C}$ $\downarrow$ & MOTP$_\subfix{O}$ $\downarrow$ & MOTP$_\subfix{I}$ $\downarrow$ \\
            \midrule
            \multirow{3}{*}{Waymo}
                           & 0-30m  & 57.33 & 0.61 & 14.66 & 0.18 \\
                           & 0-50m  & 36.87 & 0.74 & 16.10 & 0.18 \\
                           & 0-100m & 18.66 & 0.79 & 17.63 & 0.19 \\
            \midrule
            \multirow{3}{*}{nuScenes}
                           & 0-30m  & 44.47 & 0.64 & 14.18 & 0.19 \\
                           & 0-50m  & 25.93 & 0.73 & 14.84 & 0.19 \\
                           & 0-100m & 12.14 & 0.75 & 15.05 & 0.19 \\
            \bottomrule
        \end{tabular}
    }
    \label{tab:cross_dataset_centroid_based_evaluation}
\end{table}


\subsection{Amount of Data Matters.}
\label{sec:amount_of_data_matters}
We train the depth estimation module with $1\%$, $10\%$, and $100\%$ training data on the GTA dataset.
The results show how we can benefit from more data in \tabref{tab:data_amount}, where a consistent trend of performance improvement emerges as the number of data increases.
The trend of our results with a different amount of training data indicates that large-scale 3D annotation is helpful, especially with the ground truth of distant and small objects.

\begin{table}[htpb]
    \figcaption{Performance of 3D estimation on object detection IoU mAP.}{
        The evaluation demonstrates the effectiveness of our model from each separate metric.
        With different amounts of training data in our GTA dataset, the results suggest that large data capacity benefits the performance of a data-hungry network.}
    \centering
    \adjustbox{width=0.9\linewidth}{
        \begin{tabular}{lr|cccc}
            \toprule
             \multirow{2}{*}{Dataset} & \multirow{2}{*}{Amount} & \multicolumn{4}{c}{Medium}  \\
                                     &                         & $AP_\subfix{bev}^{70}$    & $AP_\subfix{3d}^{70}$       & $AP_\subfix{bbox}^{70}$  & $AP_\subfix{aos}^{70}$  \\
            \midrule
            \multirow{3}{*}{GTA}
                                     & {1\%}                   & 4.10              & 1.31            & 78.18           & 75.59           \\
                                     & {10\%}                 & 14.61            & 8.45            & 91.66           & 90.83           \\
                                     & {100\%}               & \textbf{21.61}         & \textbf{15.63}        & \textbf{94.62}        & \textbf{94.23}         \\
            \bottomrule
        \end{tabular}
    }
    \label{tab:data_amount}
\end{table}


\subsection{Comparison of matching algorithms.}
\label{sec:comparison_of_matching_algorithm}
A tracker solves the assignment problem often using combinatorial optimization algorithms, \eg, Hungarian algorithm.
However, from the comparison in~\tabref{tab:nusc_ablation_matching_algo}, we found the well-trained instance embeddings and 3D object states are robust in matching possible pairs without needing an optimizing algorithm. 
Besides, our method benefits from the lower computational complexity compared to using polynomial-time optimization algorithms.
Therefore, we choose to greedily match bipartite quasi-dense pairs of detections and tracklets throughout the experiments.

\begin{table}[htpb]
    \centering
    \figcaption{Comparison of the different matching algorithms on nuScenes validation set.}{We use only keyframes and all affinity matrices as our setting. We found that using greedy matching yields similar results to the Hungarian matching but with less computation complexity on well-trained quasi-dense embedding pairs.}
    \adjustbox{width=0.9\linewidth}{
        \begin{tabular}{l|rrrrr}
            \toprule
            Matching Algorithm & {AMOTA@1} $\uparrow$ & {AMOTA@0.2} $\uparrow$ \\
            \midrule
            Hungarian~\cite{Kuhn1955hungarian} & 0.230 & 0.3479 \\
            Greedy & 0.230 & 0.3479 \\
            \bottomrule
        \end{tabular}
    }
    \label{tab:nusc_ablation_matching_algo}
\end{table}

%% file: 06Conclusion.tex
\section{Conclusion}
We propose an online 3D object detection and tracking framework, combining quasi-dense similarity learning and 3D instance dynamics, to track moving objects in a 3D world.
Our updated pipeline consists of four parts: a single-frame monocular 3D object inference model, cross-frames contrastive feature learning network, multimodal affinity matching scheme for inter-frame object association, and an LSTM-based motion model for refining the 3D extents from location trajectories.
Besides, we introduce a 3D detection confidence to provide a balancing cue of single-frame depth estimation and multi-frame motion model refinement.
We present 3D bounding boxes depth-ordering matching for robust instance association and utilize motion-based 3D trajectory prediction for re-identification of occluded vehicles.
We design an object movement learning module, termed VeloLSTM, to update each object's location independent of camera movement.
Moreover, our ablation study and experiments show that our quasi-dense 3D tracking pipeline takes advantage of dynamic 3D trajectories and offers robust data association on urban-driving situations.
On the Waymo Open benchmark, we established the first strong baseline in camera modality with positive scores in both the 3D detection and tracking challenge.
Our quasi-dense 3D tracking pipeline outperforms the camera-based state of the art by near $500\%$ while bridging the gap to LiDAR-based methods on nuScenes 3D tracking benchmark.

%% file: 08Acknowledgements.tex
\ifCLASSOPTIONcompsoc
  \section*{Acknowledgments}
\else
  \section*{Acknowledgment}
\fi

Hu would like to thank the National Center for High-performance Computing for computer time and facilities, and Google PhD Fellowship Program for their support.
The authors gratefully acknowledge the support of Berkeley AI Research, Berkeley DeepDrive and Ministry of Science and Technology of Taiwan (MOST-110-2634-F-007-016), MOST Joint Research Center for AI Technology and All Vista Healthcare.

%% file: 07Appendix.tex

%
\appendices

\section*{Foreword}
This appendix provides technical details about our monocular quasi-dense 3D object tracking network, our training procedures for different datasets, and more qualitative and quantitative results.
\secref{apn:data_stats} offers frame- and object-based statistical summaries of our proposed simulation dataset.
\secref{apn:training_detail} describes our training procedure, and network setting of each dataset.
\secref{apn:experiments} illustrates various comparisons, including inference time, network settings and performance results, of our method on nuScenes, Waymo, GTA, and KITTI dataset.

\section{Dataset Statistics}
\label{apn:data_stats}
To help understand our dataset and its difference, we show more statistics.

\minisection{Dataset Comparison.}
\tabref{tab:dataset} demonstrates a detailed comparison with related datasets, including detection, tracking, and driving benchmarks. 
KITTI-D~\cite{geiger2012kitti} and KAIST~\cite{hwang2013multispectral} are mainly detection datasets, while KITTI-T~\cite{geiger2012kitti}, MOT15~\cite{leal2015motchallenge}, MOT16~\cite{milan2016mot16}, and UA-DETRAC~\cite{wen2015ua} are primarily 2D tracking benchmarks. 
The common drawback could be the limited scale, which cannot meet the growing demand for training data. 
Compared to the related synthetic dataset, Virtual KITTI~\cite{gaidon2016virtual} and VIPER~\cite{richter2017playing}, we additionally provide fine-grained attributes of object instances, such as color, model, maker attributes of a vehicle, motion and control signals, which leaves the space for the imitation learning system.

\begin{table*}[htpb]
    \figcaption{Comparison to related dataset for detection and tracking (Upper half: real-world, Lower half: synthetic).}{We only count the size and types of annotations for training and validation (D=detection, T=tracking, C=car, P=pedestrian). To our knowledge, our dataset is the largest 3D tracking benchmark for dynamic scene understanding, with control signals of driving, sub-categories of object.} 
	\label{tab:dataset}
	\centering
    \adjustbox{width=0.9\linewidth}{
    	\begin{tabular}{cccccccccc}
    	\toprule
		Dataset & Task & Object & Frame & Track & Boxes & 3D & Weather & Occlusion & Ego-Motion\\ \midrule
		KITTI-D~\cite{geiger2012kitti} & D & C,P & 7k & - & 41k & \checkmark & - & \checkmark & - \\
		KAIST~\cite{hwang2013multispectral} & D & P & 50k & - & 42k & - & \checkmark & \checkmark & - \\
		KITTI-T~\cite{geiger2012kitti} & T & C & 8k & 938 & 65k & \checkmark & - & \checkmark & \checkmark \\
		MOT15-3D~\cite{leal2015motchallenge} & T & P & 974 & 29 & 5632 & \checkmark & \checkmark & - & - \\
		MOT15-2D~\cite{leal2015motchallenge} & T & P & 6k & 500 & 40k & - & \checkmark & - & - \\
		MOT16~\cite{milan2016mot16} & T & C,P & 5k & 467 & 110k & - & \checkmark & \checkmark & - \\
		UA-DETRAC~\cite{wen2015ua} & D,T & C & 84k & 6k & 578k & - & \checkmark & \checkmark & - \\
		\midrule
		Virtual KITTI~\cite{gaidon2016virtual} & D,T & C & 21k & 2610 & 181k & \checkmark & \checkmark & \checkmark & \checkmark \\
		VIPER~\cite{richter2017playing} & D,T & C,P & 184k & 8272 & 5203k & \checkmark & \checkmark & \checkmark & \checkmark  \\
		Ours & D,T & C,P & 688k & 325k & 10068k & \checkmark & \checkmark & \checkmark & \checkmark \\ 
		\bottomrule
	    \end{tabular} 
	}
\end{table*}

\begin{figure}[htpb]
    \minipage{1.0\linewidth}
    \includegraphics[width=1.0\linewidth, keepaspectratio]{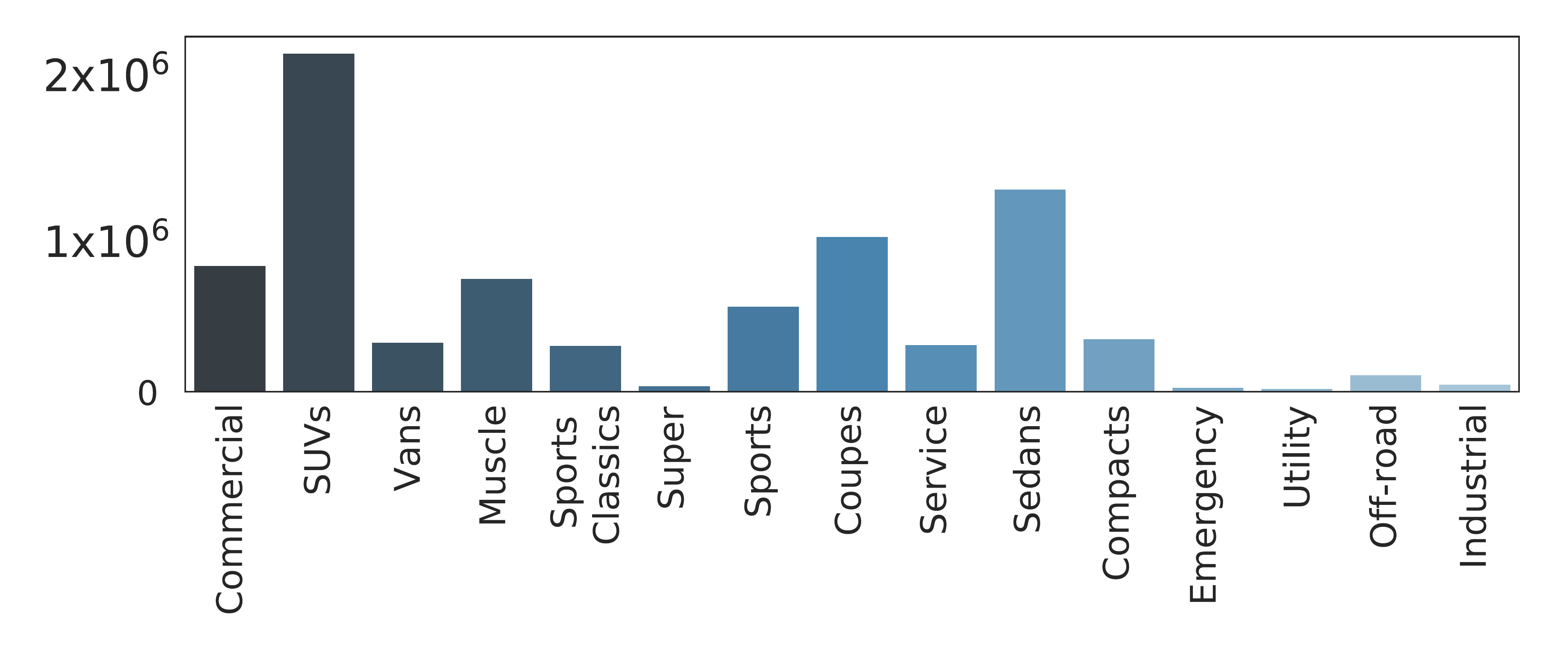}
    \subcaption{Category}\label{fig:statistics_category}
    \endminipage
    \hfill
    \minipage{0.49\linewidth}
    \includegraphics[width=1.0\linewidth, keepaspectratio]{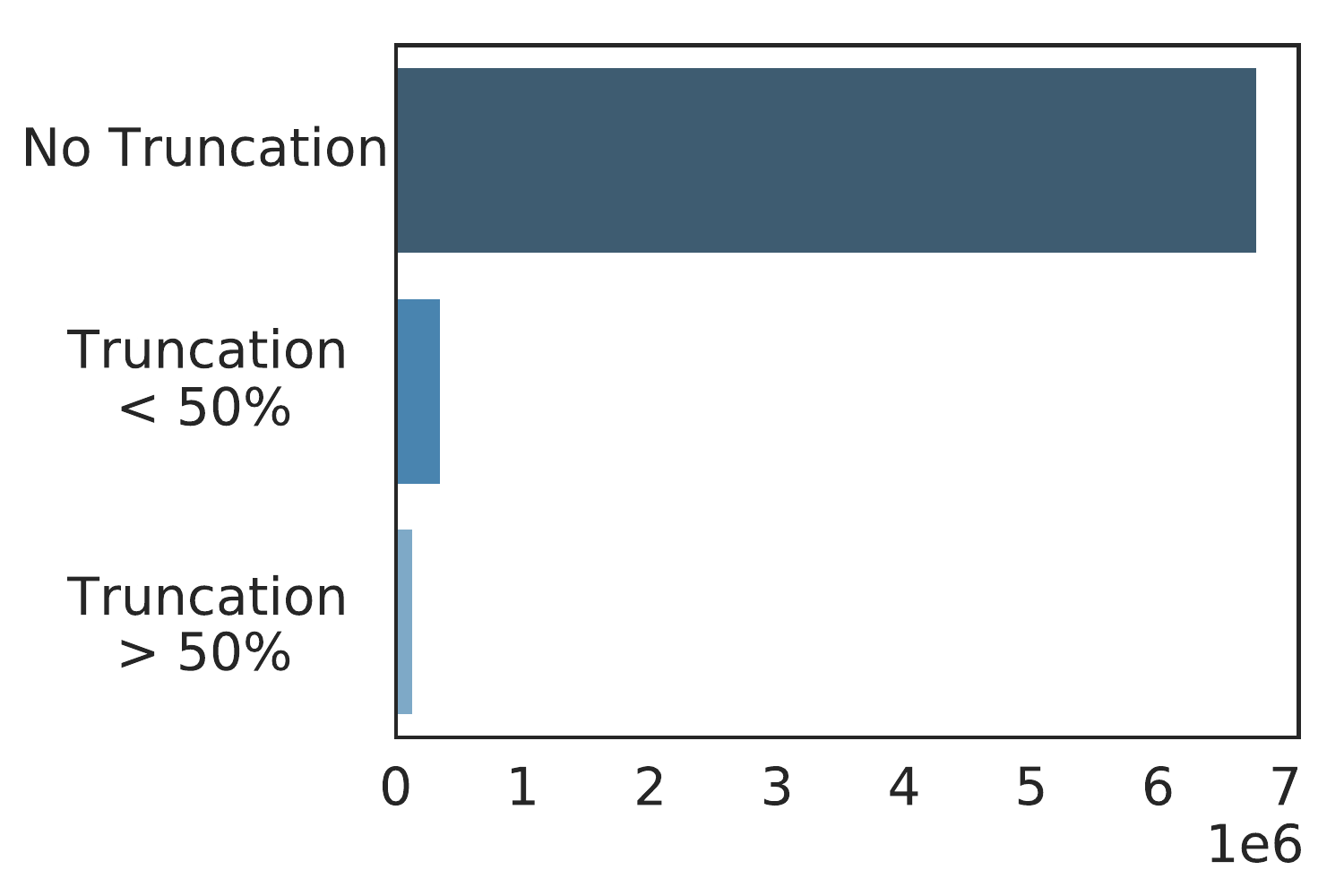}
    \subcaption{Truncation}\label{fig:statistics_truncated}
    \endminipage
    \minipage{0.49\linewidth}
    \includegraphics[width=1.0\linewidth, keepaspectratio]{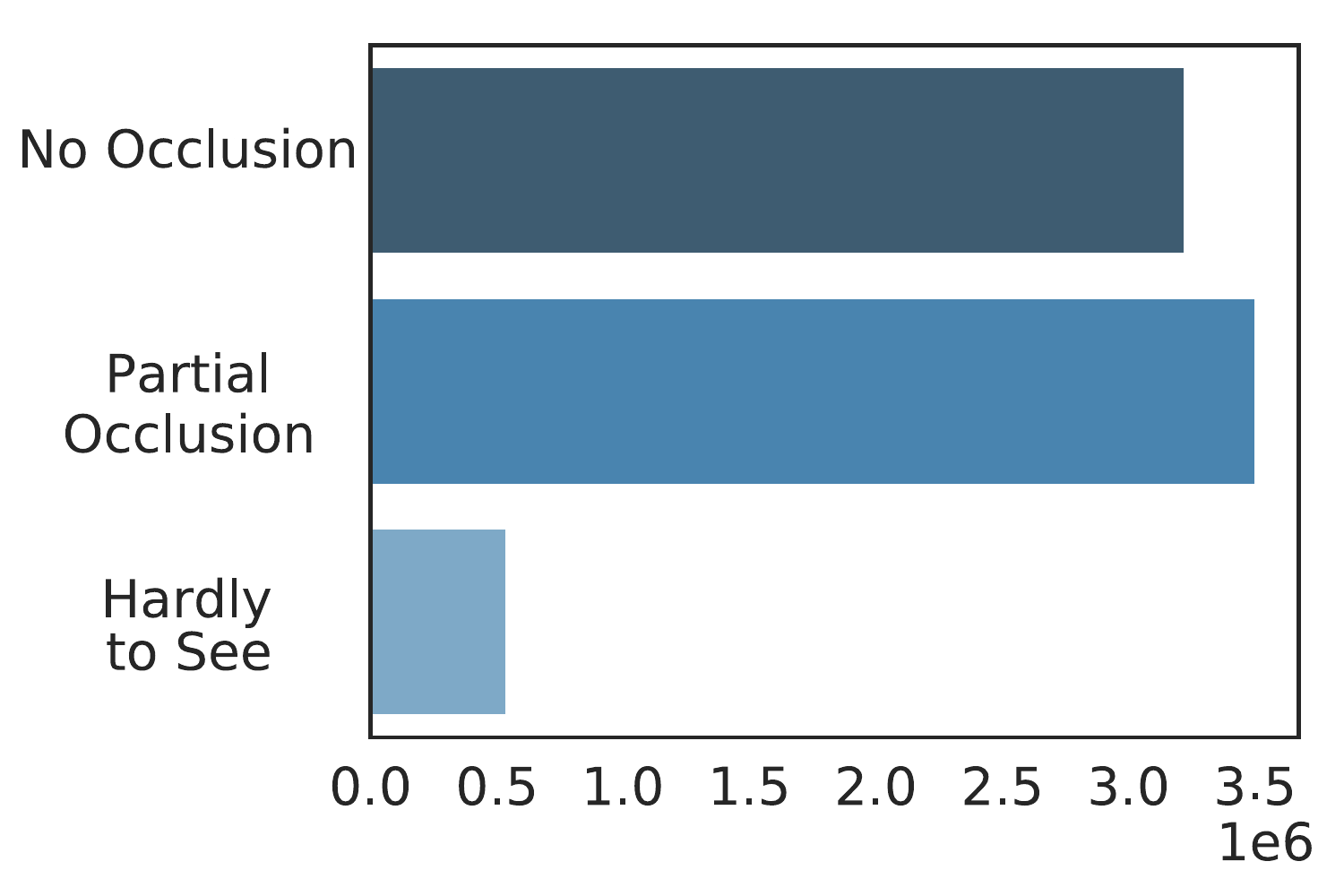}
    \subcaption{Occlusion}\label{fig:statistics_occlusion}
    \endminipage
    \figcaption{The statistics of object in our dataset.}{}
    \label{fig:statistics_object}
\end{figure}

\minisection{Number of Instances in Each Category.}
The vehicle diversity is also very large in the GTA world, featuring $15$ fine-grained subcategories. We analyzed the distribution of the $15$ subcategories in \figref{fig:statistics_category}.
Besides instance categories, we also show the distribution of occluded (\figref{fig:statistics_occlusion}) and truncated (\figref{fig:statistics_truncated}) instances to support the problem of partially visible in the 3D coordinates.

\begin{figure}[htpb]
    \minipage{0.49\linewidth}
    \includegraphics[width=1.0\linewidth, keepaspectratio]{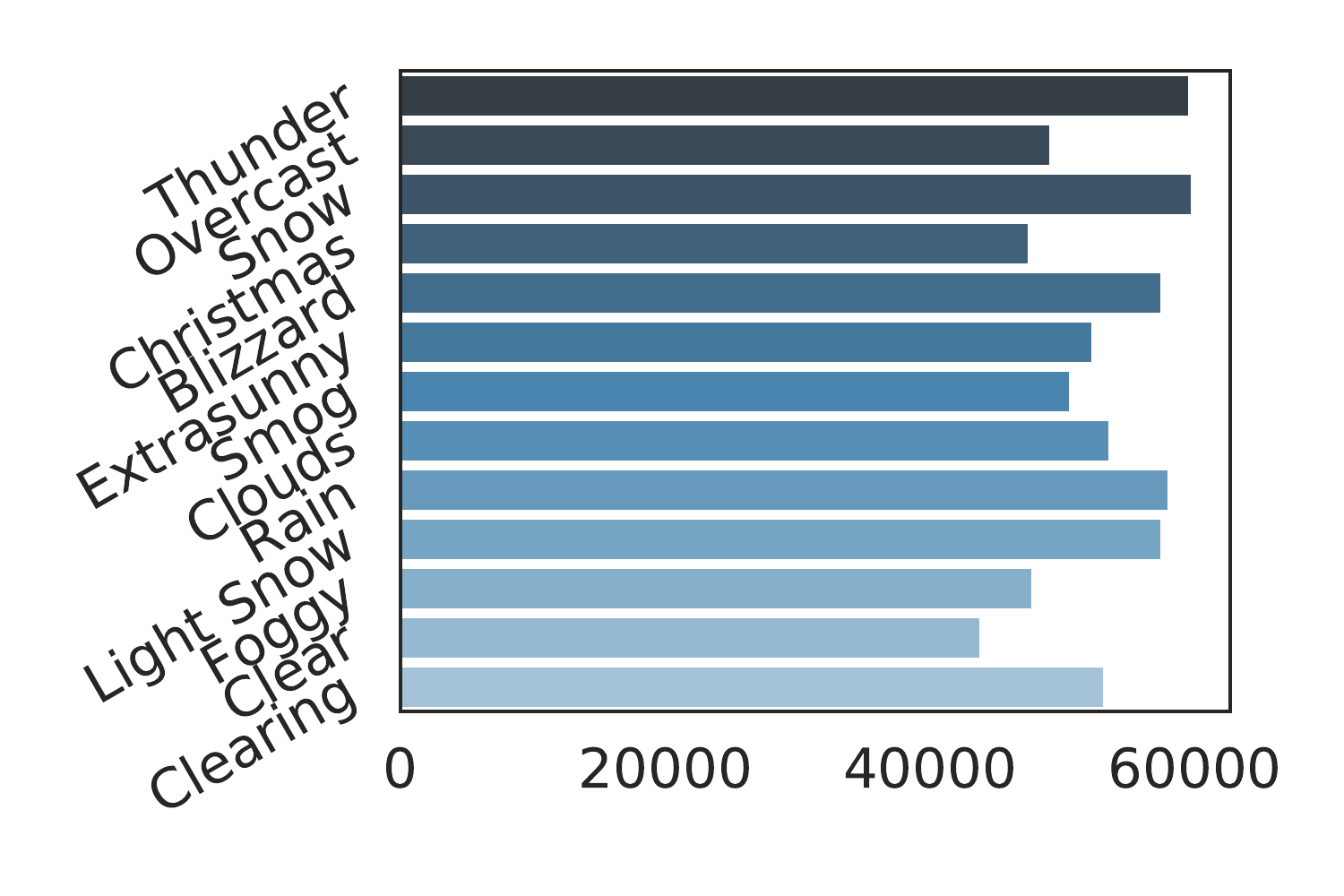}
    \subcaption{Weather}\label{fig:statistics_weather}
    \endminipage
    \minipage{0.49\linewidth}
    \includegraphics[width=1.0\linewidth, keepaspectratio]{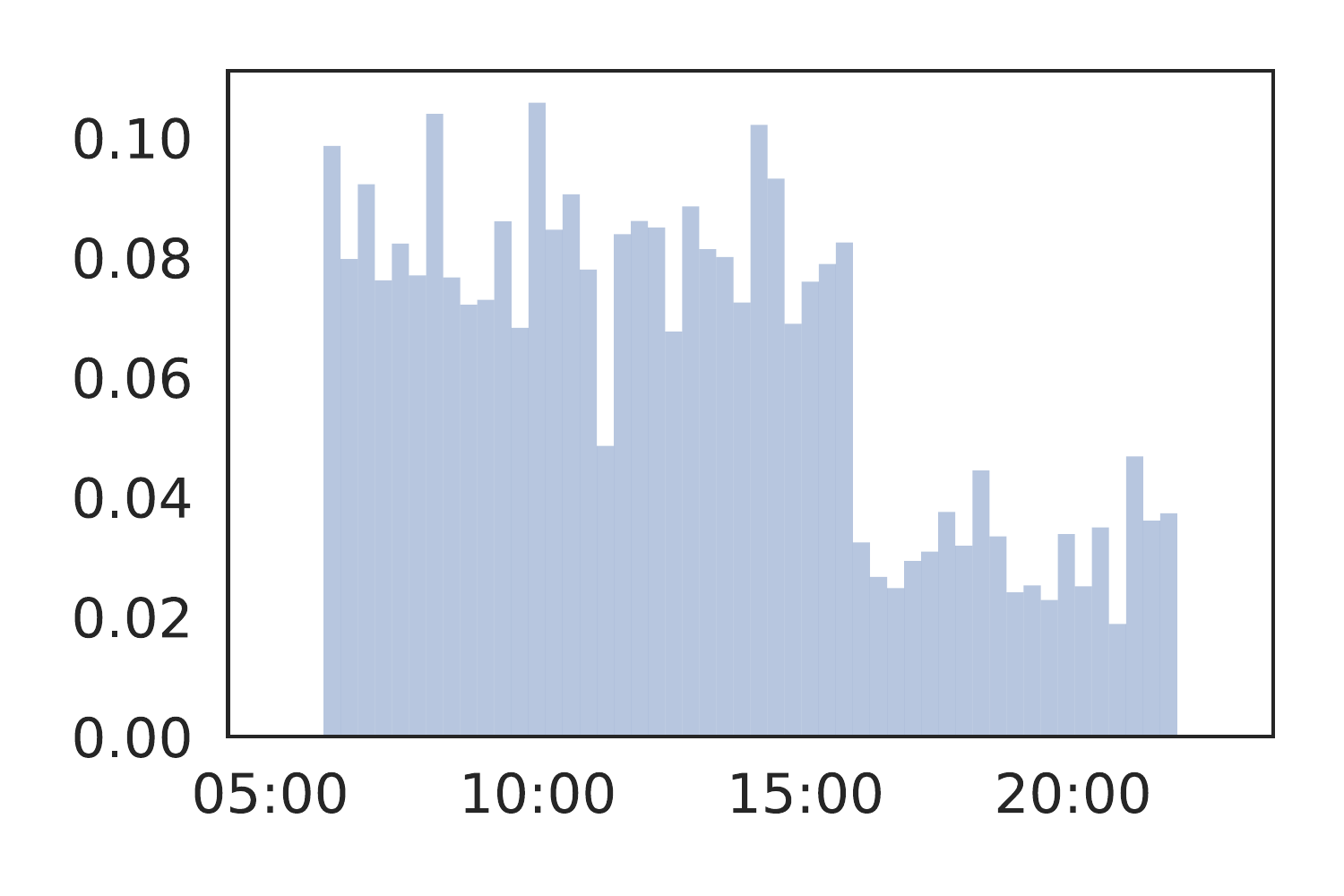}
    \subcaption{Hours}\label{fig:statistics_hours}
    \endminipage
    \figcaption{The statistics of scene in our dataset.}{}
    \label{fig:statistics_scene}
\end{figure}

\minisection{Weather and Time of Day.}
\figref{fig:statistics_scene} shows the distribution of weather, hours of our dataset.
It features a full weather cycle and time of a day in a diverse virtual environment.
By collecting various weather cycles (\figref{fig:statistics_weather}), our model learns to track with a higher understanding of environments.
With different times of a day (\figref{fig:statistics_hours}), the network handles changeable perceptual variation.

\begin{figure*}[htpb]
    \adjustbox{width=1.0\linewidth}{
        \setlength{\tabcolsep}{2pt}
        \begin{tabular}{rl}
            \includegraphics[width=0.49\linewidth, keepaspectratio]{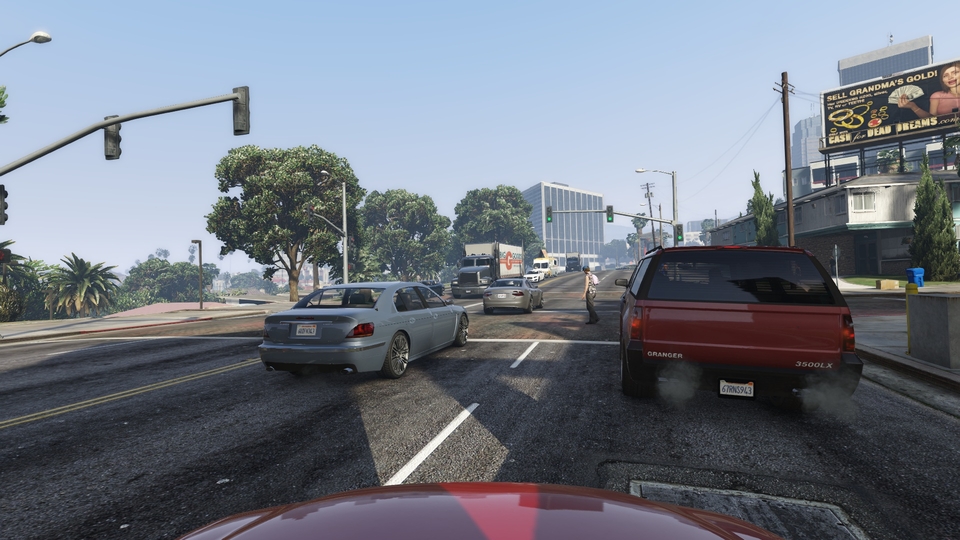} &
            \includegraphics[width=0.49\linewidth, keepaspectratio]{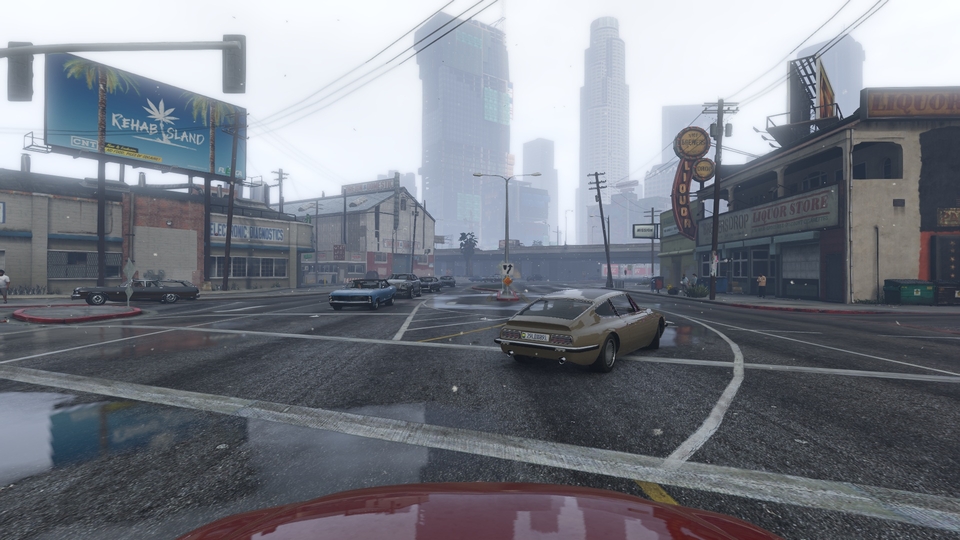}   \\
            \includegraphics[width=0.49\linewidth, keepaspectratio]{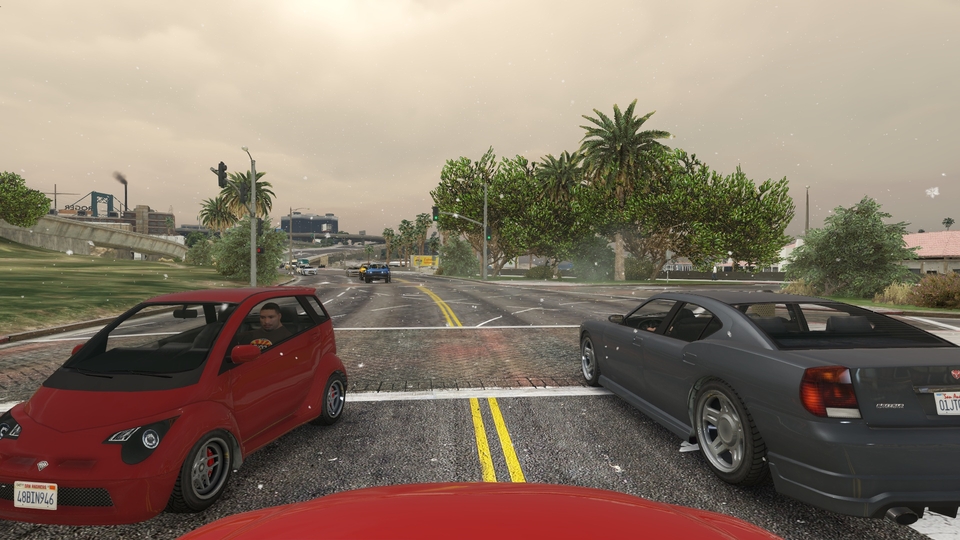} &
            \includegraphics[width=0.49\linewidth, keepaspectratio]{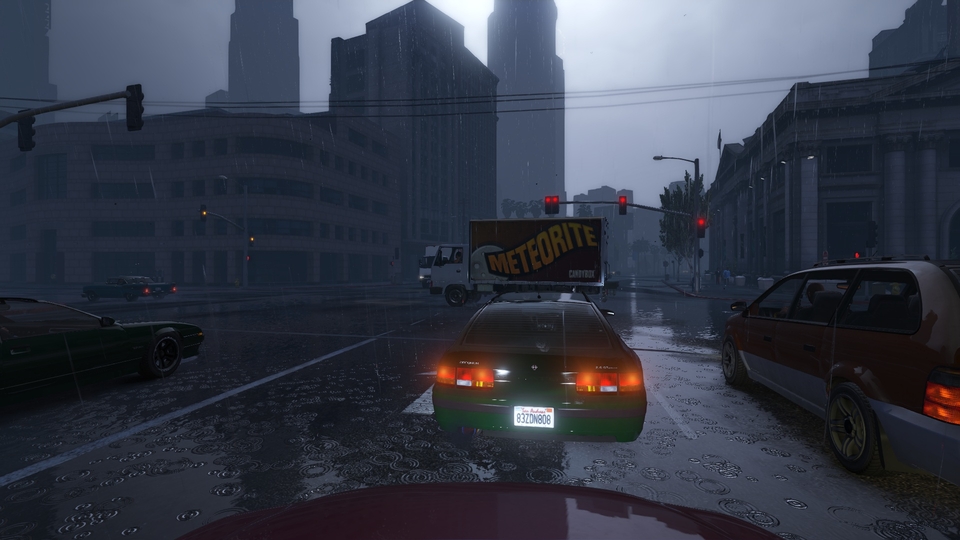}   \\
            \includegraphics[width=0.49\linewidth, keepaspectratio]{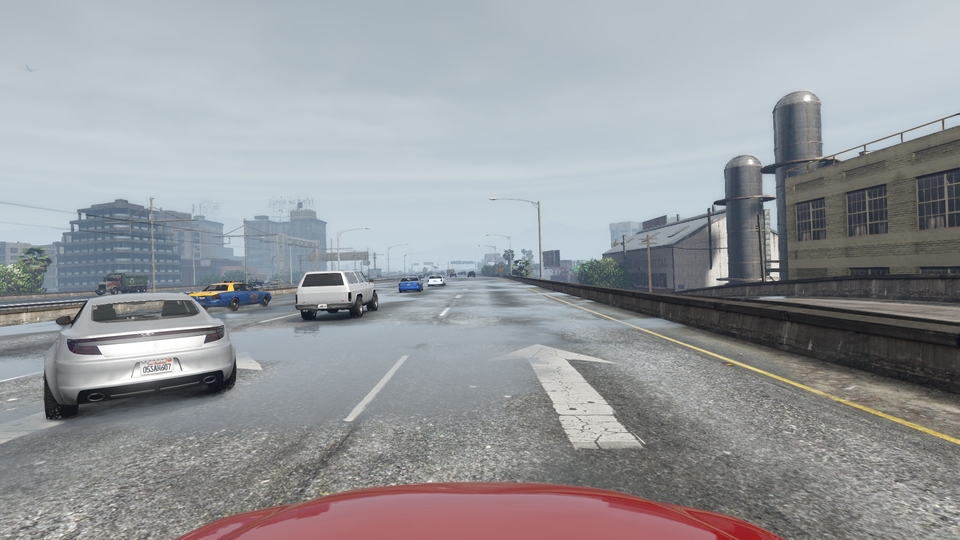} &
            \includegraphics[width=0.49\linewidth, keepaspectratio]{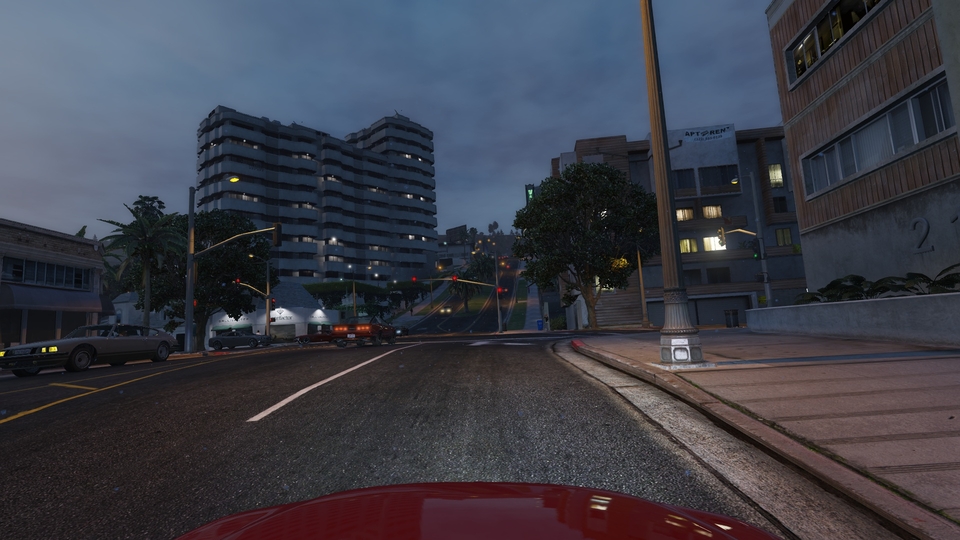}   \\
            \includegraphics[width=0.49\linewidth, keepaspectratio]{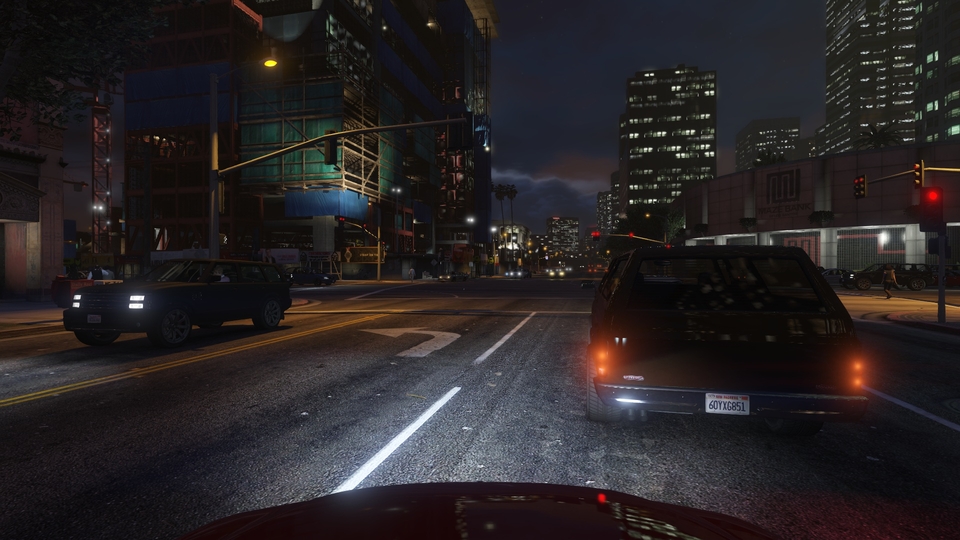} &
            \includegraphics[width=0.49\linewidth, keepaspectratio]{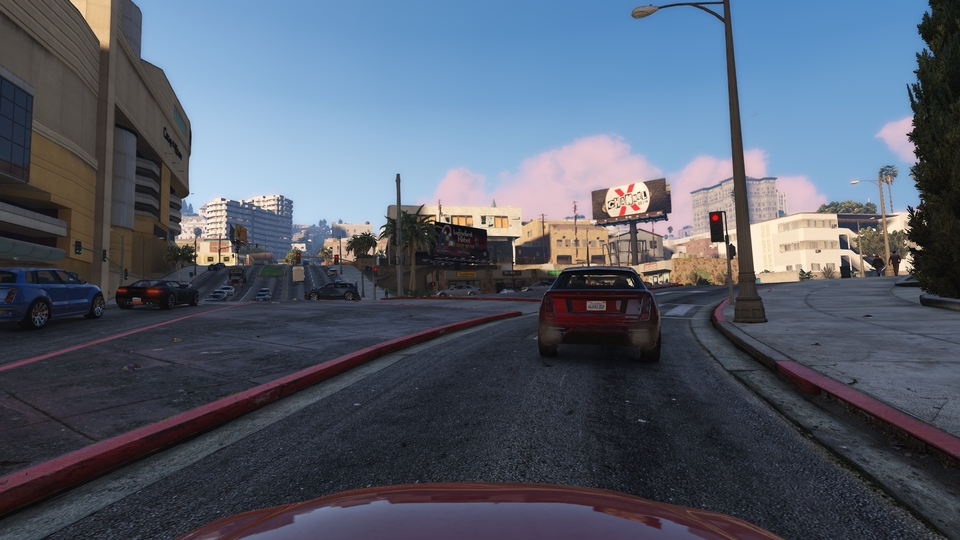}
        \end{tabular}
    }
    \figcaption{Examples of our GTA dataset.}{We collected a diverse set of driving scenes in different weather, location and time of a day.}
    \label{fig:dataset_example}
\end{figure*}

\minisection{Examples of Our Dataset.}
\figref{fig:dataset_example} shows some visual examples in different time, weather and location.

\section{Training Details}
\label{apn:training_detail}

We optimize our model on each dataset with different training procedures according to its amount of data and the limit of the GPU memory.

\minisection{Training Procedure.}
For nuScenes, we train our model on the training set with all $10$ object classes and evaluate the performance on the validation set with $10$ object classes for detection task and $7$ object classes for tracking task.
We trained the model on $8$ GPUs with a total batch size of $32$ for $24$ epochs. The learning rate was linearly increased from $1\times10^{-3}$, to $1\times10^{-2}$ over the initial $1000$ warm-up training steps decreased by $0.5$ after $16$ and $22$ epochs.

For Waymo, we trained the model on $8$ GPUs with a total batch size of $16$ for $24$ epochs. The learning rate was linearly increased from $5\times10^{-4}$, to $5\times10^{-3}$ over the initial $1000$ warm-up training steps. Also, due to the large amount of the training data, we decrease the learning rate by $0.5$ after $8$, $12$, $16$, $20$, and $22$ epochs.

For GTA, we trained the model on $8$ GPUs with a total batch size of $24$ for $24$ epochs. The learning rate was linearly increased from $1\times10^{-3}$, to $1\times10^{-2}$ over the initial $1000$ warm-up training steps. Due to the large amount of the training data, we decreased the learning rate by $0.5$ after $8$, $12$, $16$, $20$, and $22$ epochs.

For KITTI, we fine-tuned the model from a GTA 3D detection model on $4$ GPUs with a total batch size of $8$ for $24$ epochs. The learning rate was linearly increased from $5\times10^{-4}$, to $5\times10^{-3}$ over the initial $1000$ warm-up training steps and decreased by $0.5$ after $16$ and $22$ epochs.

\section{Experiments}
\label{apn:experiments}



\minisection{Inference Time.}
The average inference time is $123.3+2.3+12.3+1.9+22.2=162.0$ ms with $11095$ frames on a single RTX 2080Ti GPU (see Table~\ref{tab:inference_time} for details).
Note that the KITTI benchmark focuses only on the non-detection part inference time ($22.2$ ms).

\begin{table}[htpb]
    \adjustbox{width=0.7\linewidth}{
        \begin{tabular}{l|c}
            \toprule
            Phase & Second \\
            \midrule
            \textbf{Detection} & \textbf{0.1256} \\
            \midrule
            2D Box, 3D Estimation & 0.1233 \\
            Contrastive Feature & 0.0023 \\
            \midrule
            \textbf{3D transform} & \textbf{0.0142} \\
            \midrule
            3D Lifting (img to world) & 0.0123 \\
            Reprojection (world to img) & 0.0019 \\
            \midrule
            \textbf{Tracking} & \textbf{0.0222} \\
            \midrule
            LSTM Predict & 0.0076 \\
            Greedy Matching & 0.0019 \\
            LSTM Update & 0.0009 \\
            Misc. & 0.0118 \\
            \midrule
            \textbf{Total} & \textbf{0.1620} \\
            \bottomrule
        \end{tabular}
    }
    \figcaption{Inference Time (second) of our proposed framework on KITTI tracking benchmark.}{We recorded the wall clock of the execution time consumption of each module, which might have slight overhead, \eg, functions with result dumping. Bold fonts are the summation of each sub-module and misc. time. Note that elapsed time for object detection is not included in the specified runtime of the KITTI benchmark.}
    \label{tab:inference_time}
\end{table}

\minisection{Data Association Weights.}
We use different weights of appearance during our experiment, 3D IoU overlap, and motion overlap for corresponding methods.
We select weight ratios based on the results of our validation set.
For deep appearance only methods, we give $100\%$ weighting to $w_{\subfix{deep}}$.
For 3D related methods, we set $w_{\subfix{deep}}$ with $0.5$ to balance appearance and 3D extents.

\minisection{Tracking Performance on nuScenes dataset.}
We report the full table of nuScenes tracking benchmark in~\tabref{tab:nusc_tracking_test_full}.

\begin{table*}[tpb]
    \centering
    \figcaption{Tracking performance on the testing set of nuScenes tracking benchmark~\cite{caesar2019nuscenes}.}{We report the average AMOTA@1 and AMOTP over $7$ categories on the benchmark. Camera-based and published methods are shown. Our quasi-dense 3D tracking achieves the top ranking with about $5$ times the 3D tracking performance of the best camera-based submission, CenterTrack Vision~\cite{zhou2020centertrack} (score underlined), at the time of submission. We mark the best result in \textbf{bold} and runner-up in \textit{italic}.
    }
    \adjustbox{width=0.8\linewidth}{
        \begin{tabular}{lll|rr}
            \toprule
            Method               & Published                 & Modality       & AMOTA@1.0 $\uparrow$ & AMOTP $\downarrow$ \\
            \midrule
            \textbf{Ours}        &                            & Camera         & \textbf{0.217}       & 1.550 \\
            DEFT                 &                            & Camera         & \textit{0.177}       & 1.564 \\
            Megvii-AB3DMOT       & \cite{caesar2019nuscenes}  & LiDAR          & 0.151                & 1.501 \\
            CenterTrack-Open     & \cite{zhou2020centertrack} & LiDAR + Camera & 0.108                & \textbf{0.989} \\
            DBNet                &                            & Camera         & 0.072                & \textit{1.489} \\
            ProTracker           &                            & Camera         & 0.072                & 1.628 \\
            Buffalo\_Vision      &                            & Camera         & 0.059                & 1.490 \\
            CaTracker            &                            & Camera         & 0.053                & 1.611 \\
            CenterTrack-Vision   & \cite{zhou2020centertrack} & Camera         & \underline{0.046}                & 1.543 \\
            GDLG                 &                            & Camera         & 0.045                & 1.819 \\
            PointPillars-AB3DMOT & \cite{caesar2019nuscenes}  & LiDAR          & 0.029                & 1.703 \\
            Mapillary-AB3DMOT    & \cite{caesar2019nuscenes}  & LiDAR          & 0.018                & 1.790 \\
            \bottomrule
        \end{tabular}
    }
    \label{tab:nusc_tracking_test_full}
\end{table*}

\minisection{Tracking Performance on KITTI dataset.}
As mentioned in the main paper, we resort to KITTI~\cite{geiger2012kitti} tracking benchmarks to compare our model abilities in the real-world scenario.
We have improved our previous results~\cite{Hu2019Mono3DT} to third place with five ranks improvement and have increased from $84.52$ to $86.41$ in MOTA.
Our quasi-dense 3D tracking method ranked in the top tier among all the published methods upon the time of submission.
Please note that we focus on tracking module improvements and not over-design the detector-specific architecture to fit on a small dataset.
Results are listed in \tabref{tab:kitti_full_tracking}.

\begin{table*}[htpb]
    \figcaption{Tracking performance on the testing set of KITTI tracking benchmark.}{Improving from mono3DT~\cite{Hu2019Mono3DT}, we have greatly increased the ranking position by leveraging the quasi-dense contrastive feature in the data association part. However, the goal objective of our training aims for 3D object tracking, and we are not optimizing the network for the 2D tracking task. Only published methods are reported.
    We mark the best result in \textbf{bold} and runner-up in \textit{italic}.}
    \centering
    \adjustbox{width=0.8\linewidth}{
        \begin{tabular}{lcc|rrrrrr}
            \toprule
            {Benchmark}                             & LiDAR     & Stereo   & {MOTA} $\uparrow$ & {MOTP} $\uparrow$ & {MT} $\uparrow$ & {ML} $\downarrow$ & {FP} $\downarrow$ & {FN} $\downarrow$ \\
            \midrule
            SRK\_ODESA~\cite{mykheievskyi2020odesa} &            &            & \textbf{90.03}    & 84.32             & \textbf{82.62}  & \textbf{2.31}     & \textit{451}      & \textit{2887}     \\
            CenterTrack~\cite{zhou2020centertrack}  &            &            & \textit{89.44}    & 85.05             & \textit{82.31}  & \textbf{2.31}     & 849               & \textbf{2666}     \\
            Ours                                    &            &            & 86.41             & \textbf{85.82}    & 75.38           & \textit{2.46}     & 804               & \textit{3761}     \\
            Quasi-Dense~\cite{pang2020quasidense}   &            &            & 85.76             & 85.01             & 69.08           & 3.08              & 517               & 4288              \\
            JRMOT~\cite{shenoi2020jrmot}            & \checkmark &            & 85.70             & 85.48             & 71.85           & 4.00              & 772               & 4049              \\
            MASS~\cite{karunasekera2019mass}        &            &            & 85.04             & 85.53             & 74.31           & 2.77              & 742               & 4101              \\
            MOSTFusion~\cite{voigtlaender2019mots}  &            & \checkmark & 84.83             & 85.21             & 73.08           & 2.77              & 681               & 4260              \\
            mmMOT~\cite{zhang2019mmMOT}             & \checkmark &            & 84.77             & 85.21             & 73.23           & 2.77              & 711               & 4243              \\
            Mono3DT~\cite{Hu2019Mono3DT}            &            &            & 84.52             & 85.64             & 73.38           & 2.77              & 705               & 4242              \\
            BeyondPixels~\cite{MOTBeyondPixels}     &            &            & 84.24             & \textit{85.73}    & 73.23           & 2.77              & 705               & 4247              \\
            AB3DMOT~\cite{weng2020ab3dmot}          & \checkmark &            & 83.84             & 85.24             & 66.92           & 11.38             & 1059              & 4491              \\
            PMBM~\cite{Scheidegger2018pmbm}         & \checkmark &            & 80.39             & 81.26             & 62.77           & 6.15              & 1007              & 5616              \\
            extraCK~\cite{gunduz2018lightweight}    &            &            & 79.99             & 82.46             & 62.15           & 5.54              & 642               & 5896              \\
            MCMOT-CPD~\cite{Lee2016MCMOTCPD}        &            &            & 78.90             & 82.13             & 52.31           & 11.69             & \textbf{316}      & 6713              \\
            NOMT~\cite{choi2015nomt}                &            & \checkmark & 78.15             & 79.46             & 57.23           & 13.23             & 1061              & 6421              \\
            MDP~\cite{xiang2015mdptrack}            &            &            & 76.59             & 82.10             & 52.15           & 13.38             & 606               & 7315              \\
            DSM~\cite{frossard2018dsm}              &            &            & 76.15             & 83.42             & 60.00           & 8.31              & 578               & 7328              \\
            SCEA~\cite{hong2016scea}                &            &            & 75.58             & 79.39             & 53.08           & 11.54             & 1306              & 6989              \\
            CIWT~\cite{Osep17ICRAciwt}              &            &            & 75.39             & 79.25             & 49.85           & 10.31             & 954               & 7345              \\
            NOMT-HM~\cite{choi2015nomt}             &            & \checkmark & 75.20             & 80.02             & 50.00           & 13.54             & 1143              & 7280              \\
            mbodSSP~\cite{lenz2015followme}         &            &            & 72.69             & 78.75             & 48.77           & 8.77              & 1918              & 7360              \\
            \bottomrule
        \end{tabular}
    }
    \label{tab:kitti_full_tracking}
\end{table*}

\begin{figure*}[htpb]
    \minipage{0.50\textwidth}
    \includegraphics[width=1.0\linewidth, keepaspectratio]{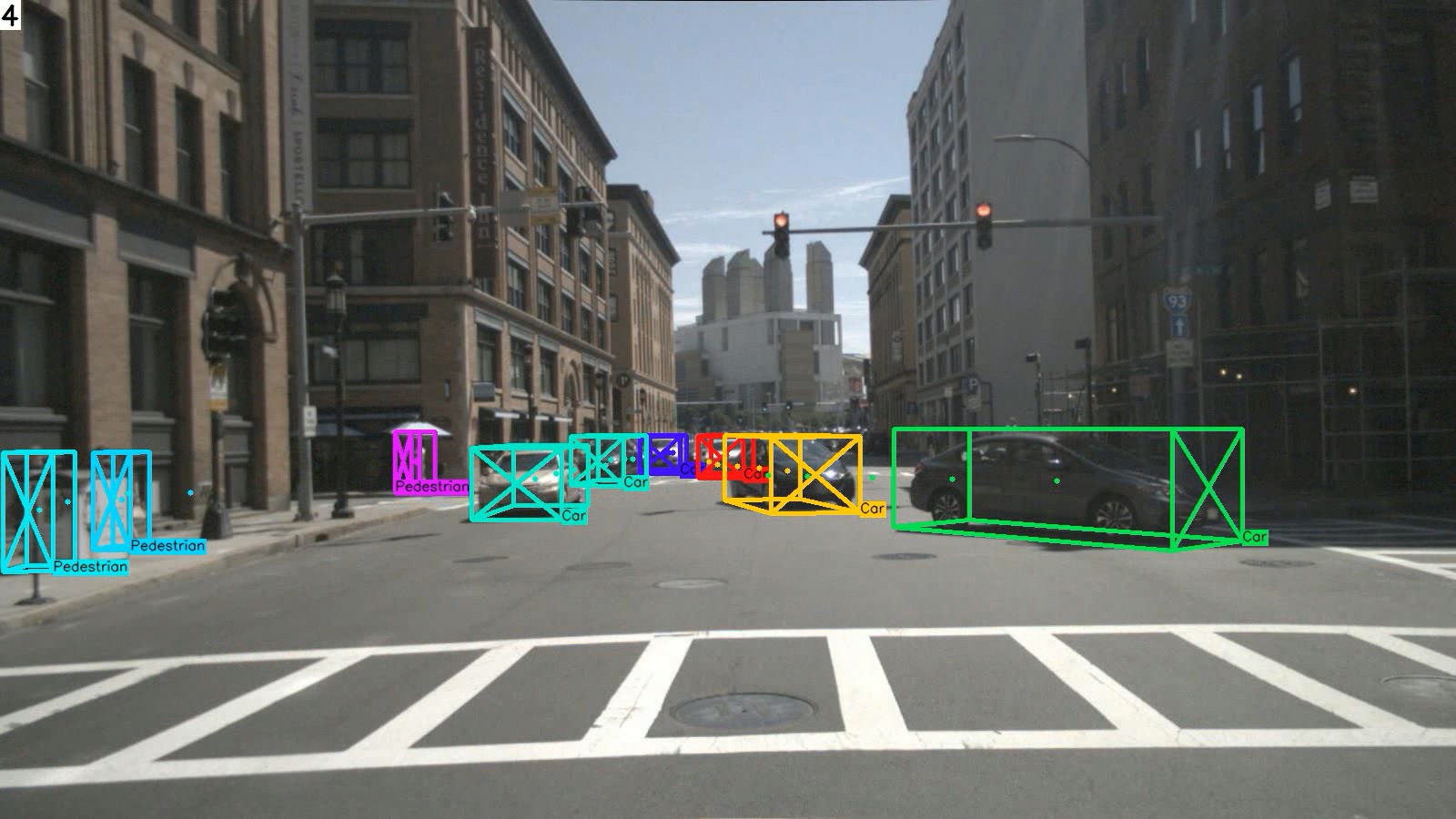}
    \includegraphics[width=1.0\linewidth, keepaspectratio]{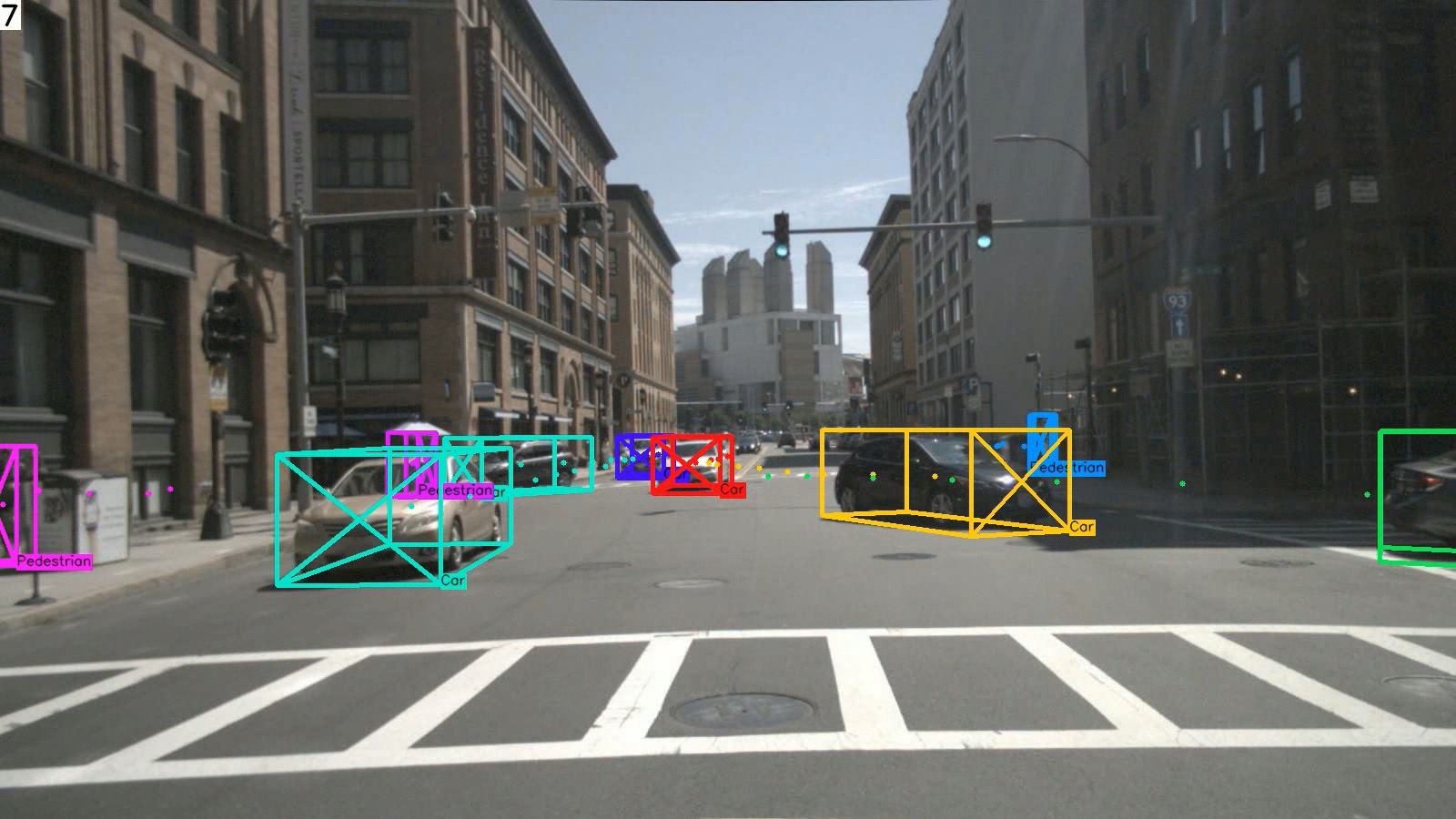}
    \includegraphics[width=1.0\linewidth, keepaspectratio]{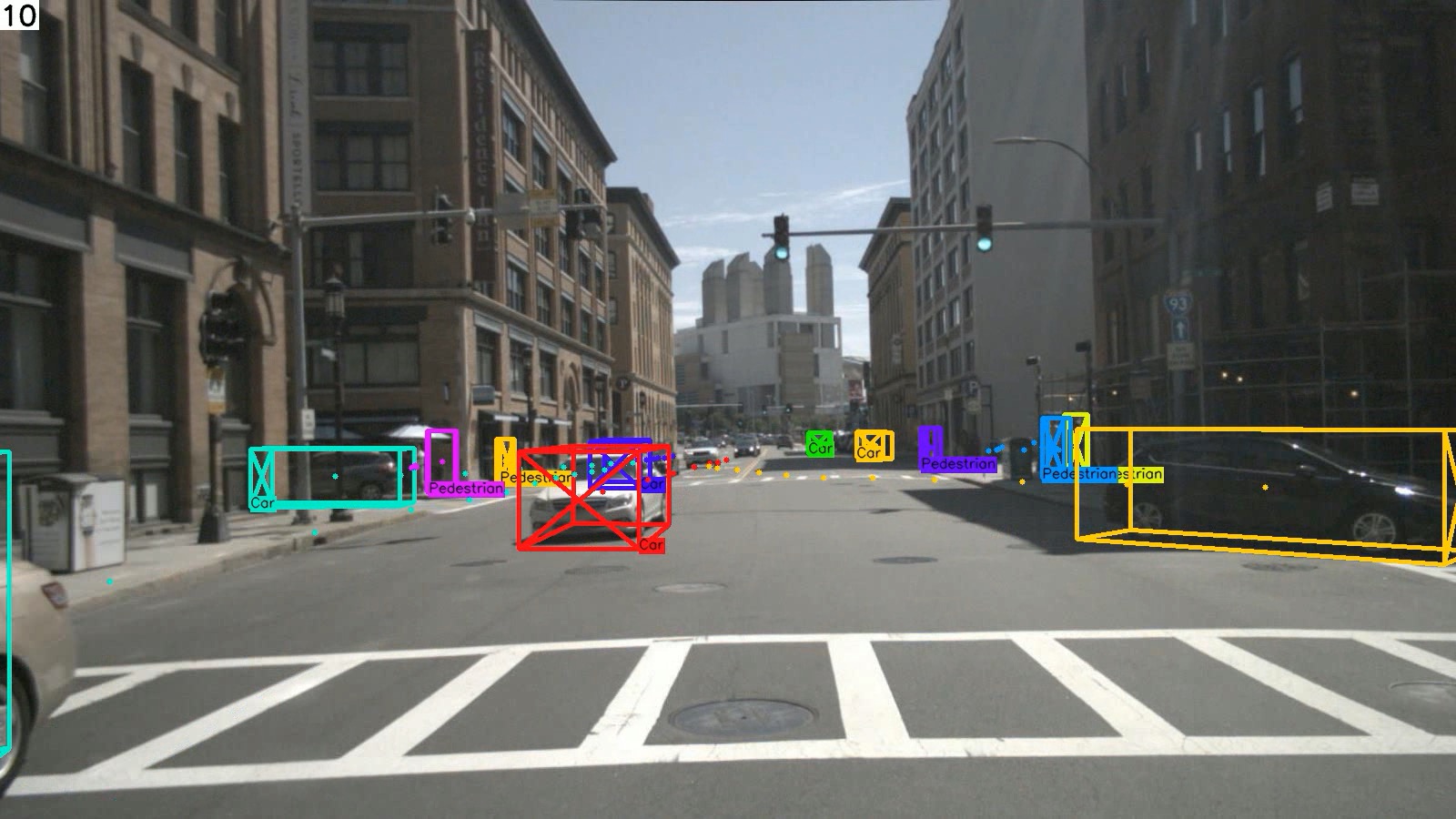}
    \includegraphics[width=1.0\linewidth, keepaspectratio]{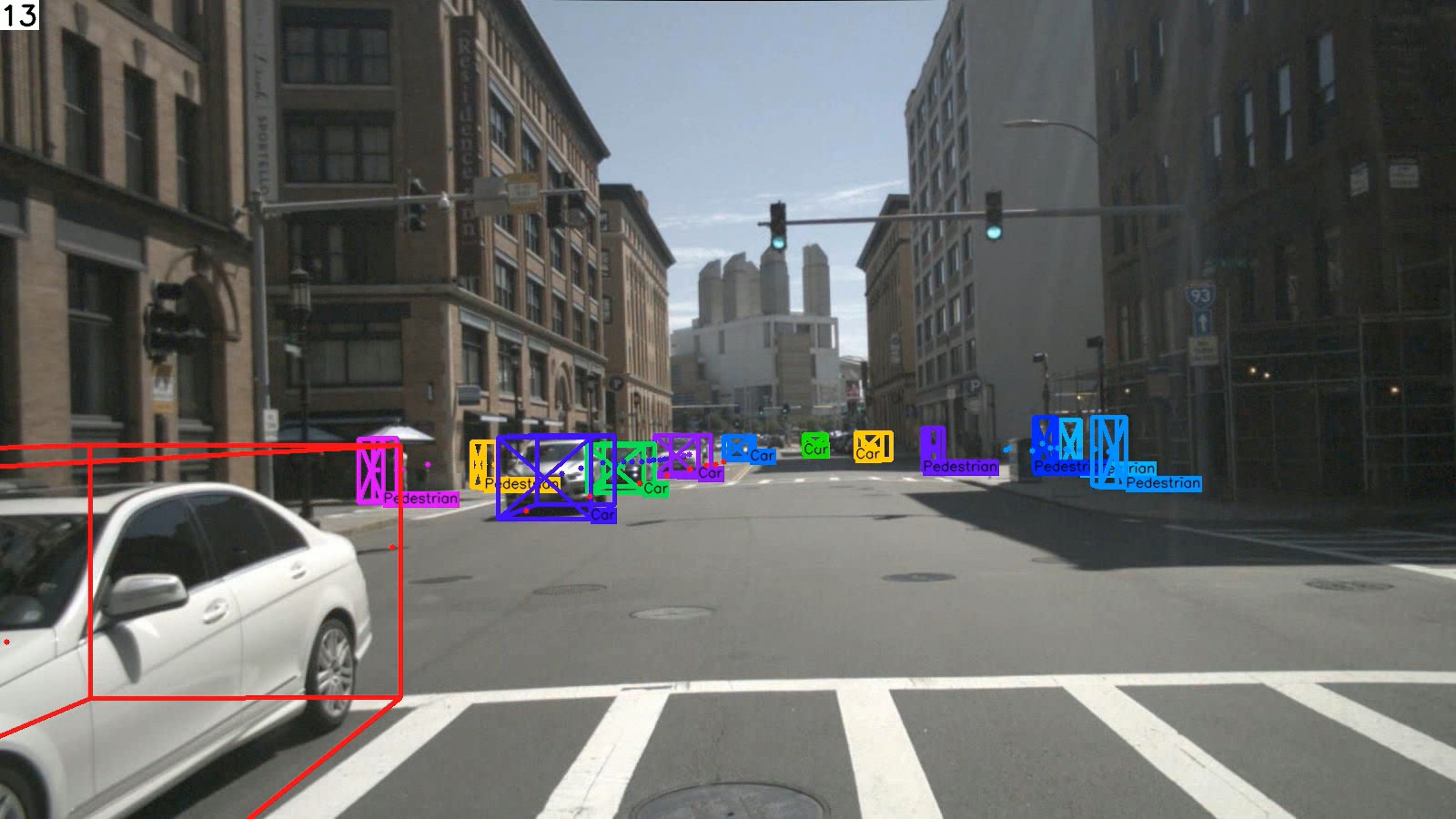}
    \endminipage
    \minipage{0.28\textwidth}
    \includegraphics[width=1.0\linewidth, keepaspectratio]{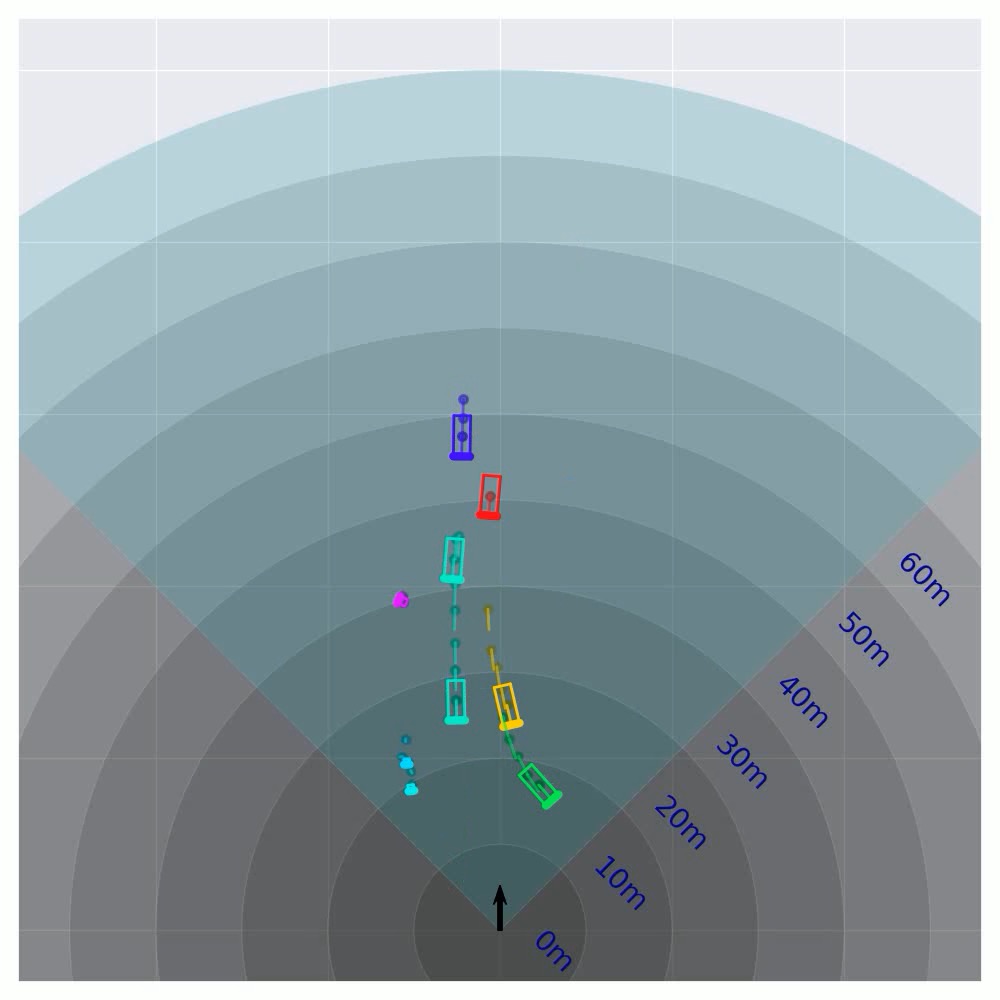}
    \includegraphics[width=1.0\linewidth, keepaspectratio]{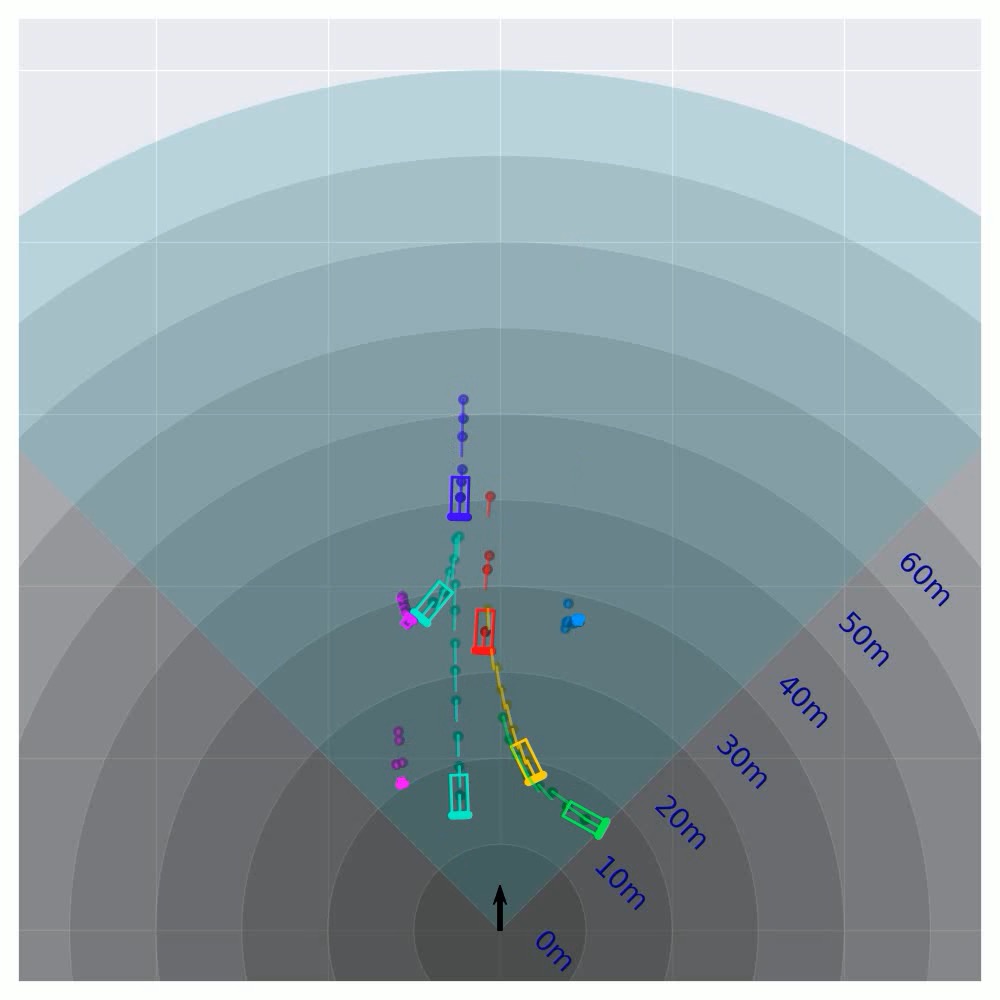}
    \includegraphics[width=1.0\linewidth, keepaspectratio]{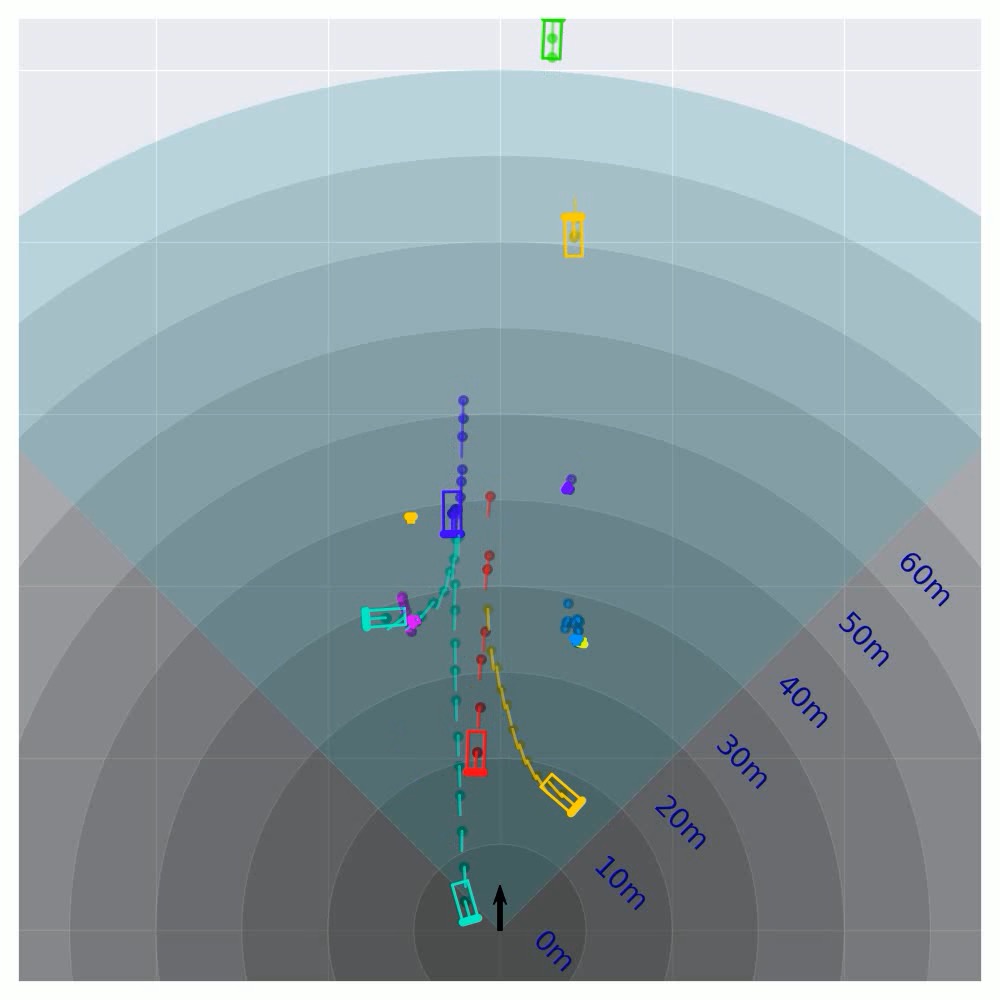}
    \includegraphics[width=1.0\linewidth, keepaspectratio]{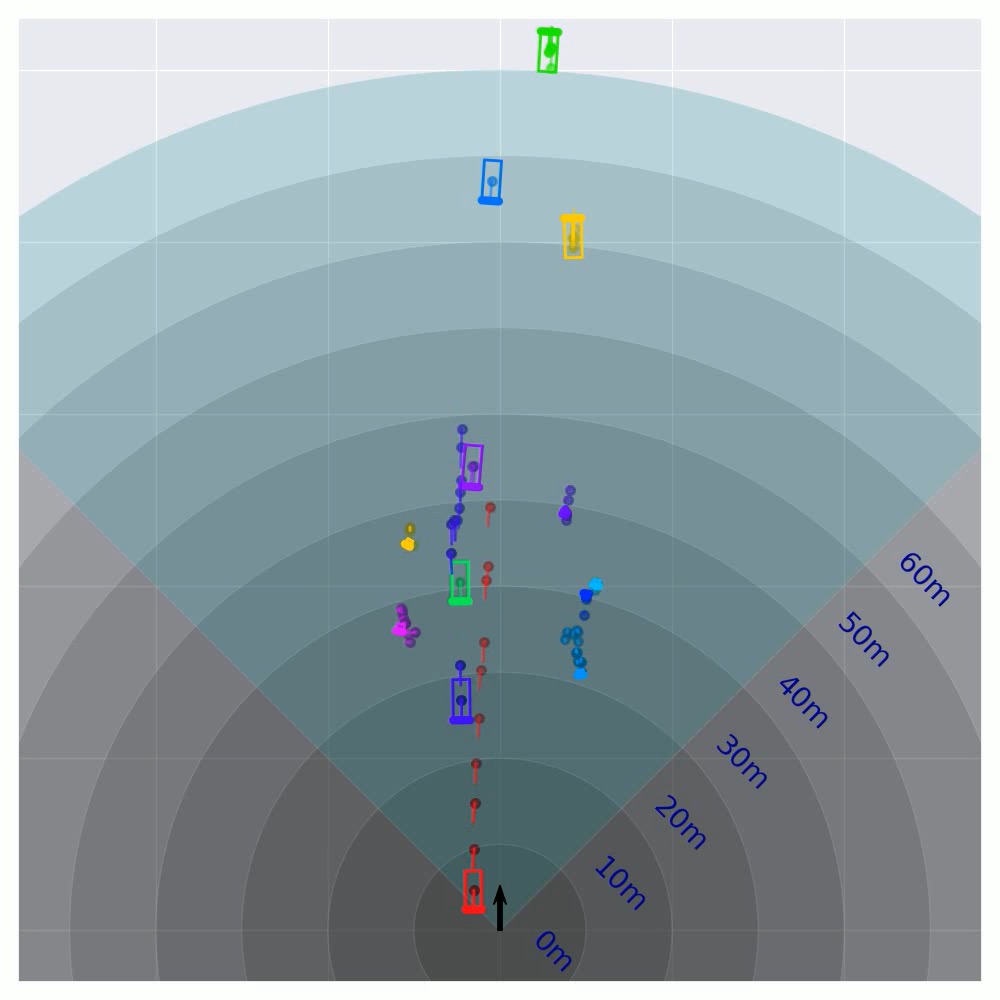}
    \endminipage
    \figcaption{Qualitative results on the testing set of nuScenes dataset.}{Our QD-3DT robustly tracks all observed objects and locates them in 3D. We show predicted 3D bounding boxes and trajectories colored with tracking IDs. Better visualization with color.}
    \label{fig:qualitative_nusc}
\end{figure*}

\begin{figure*}[htpb]
    \minipage{0.46\textwidth}
    \includegraphics[width=1.0\linewidth, keepaspectratio]{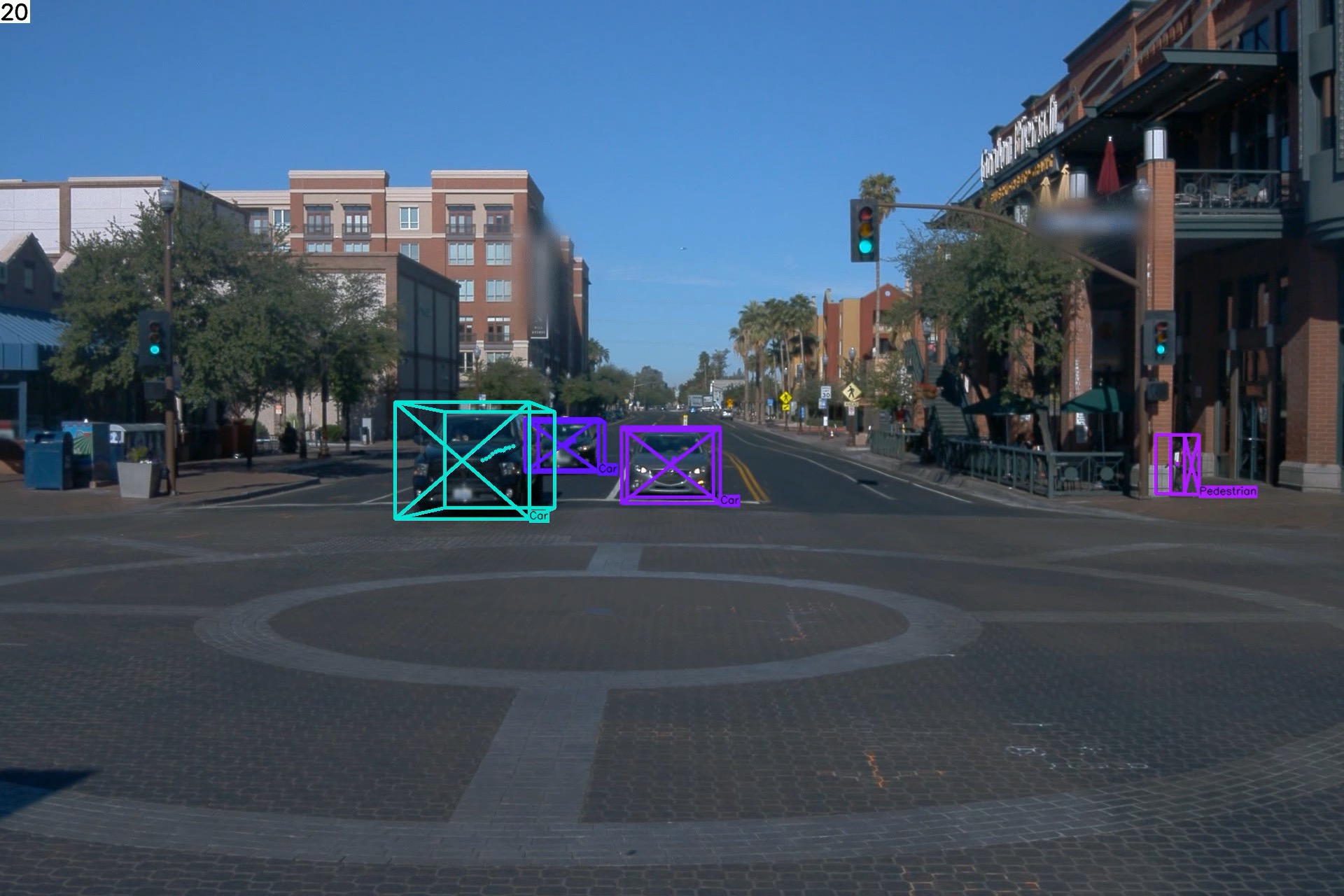}
    \includegraphics[width=1.0\linewidth, keepaspectratio]{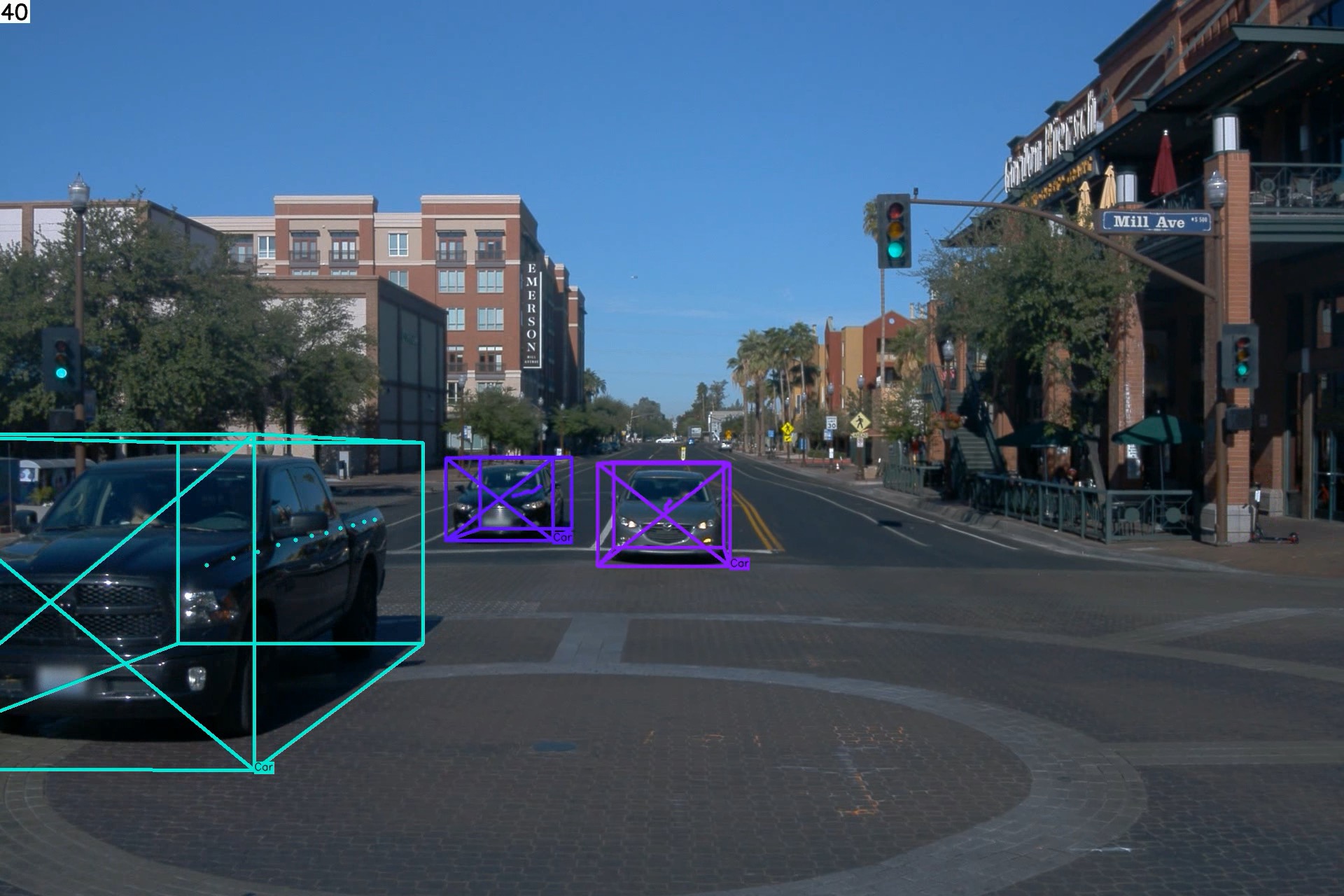}
    \includegraphics[width=1.0\linewidth, keepaspectratio]{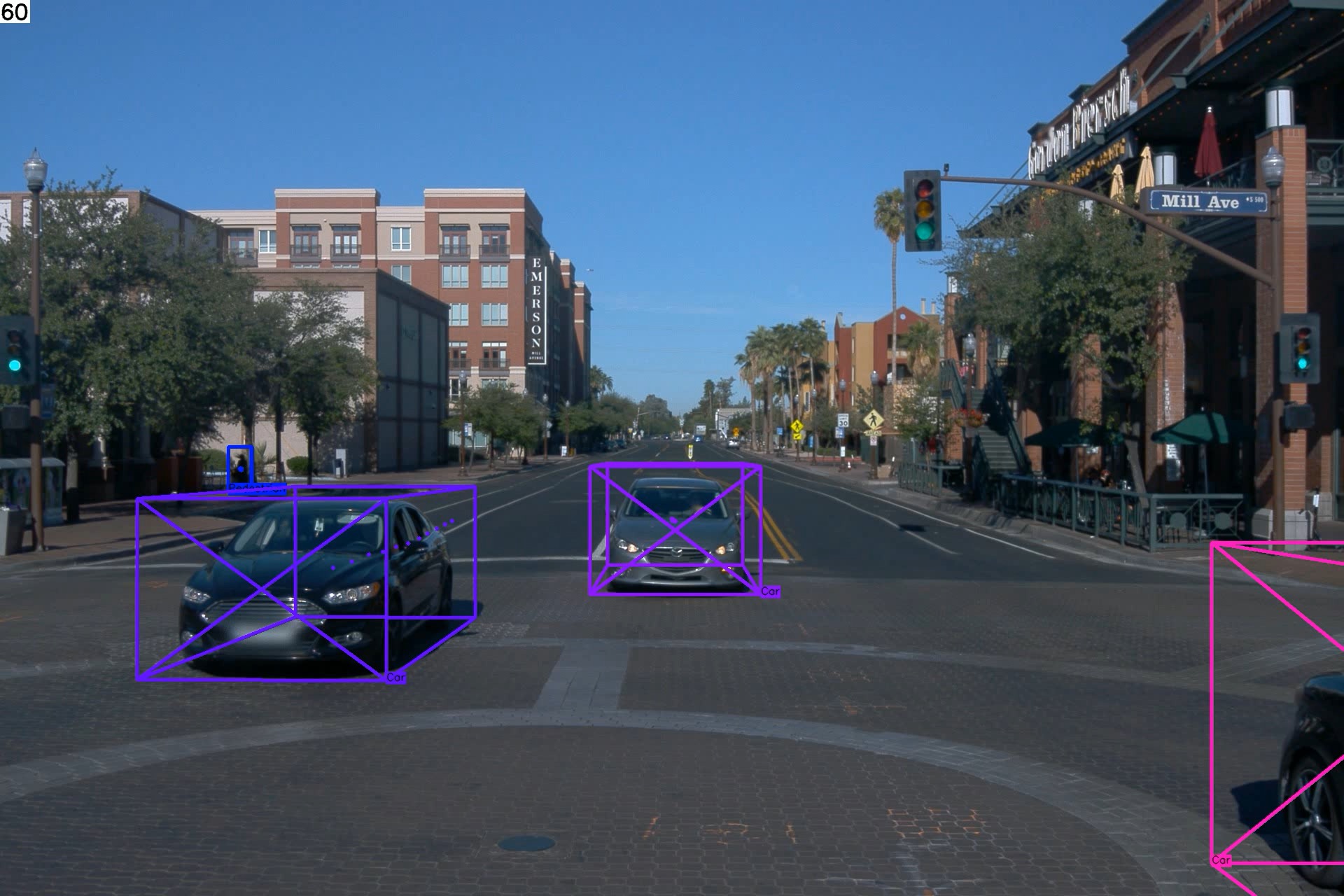}
    \includegraphics[width=1.0\linewidth, keepaspectratio]{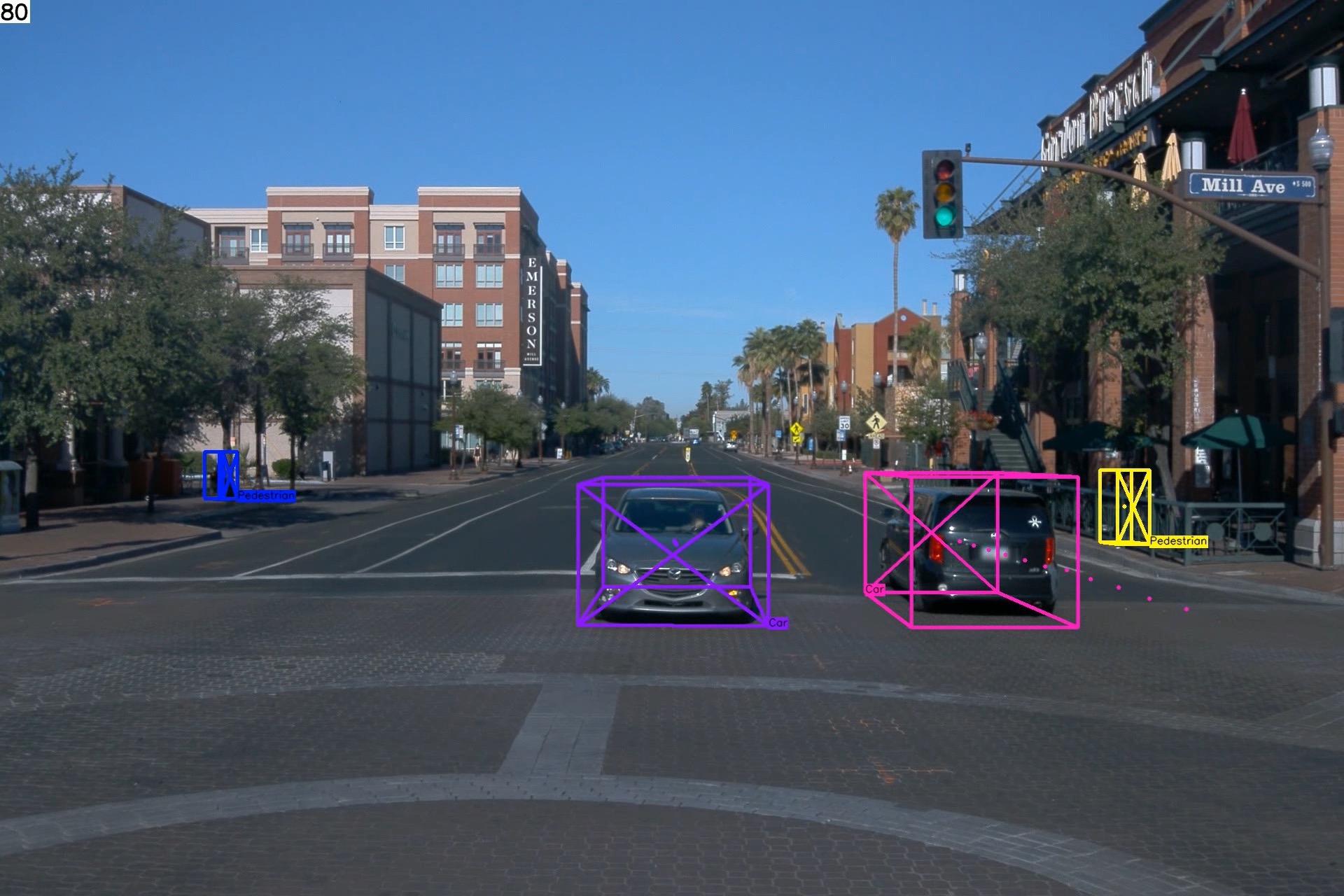}
    \endminipage
    \minipage{0.31\textwidth}
    \includegraphics[width=1.0\linewidth, keepaspectratio]{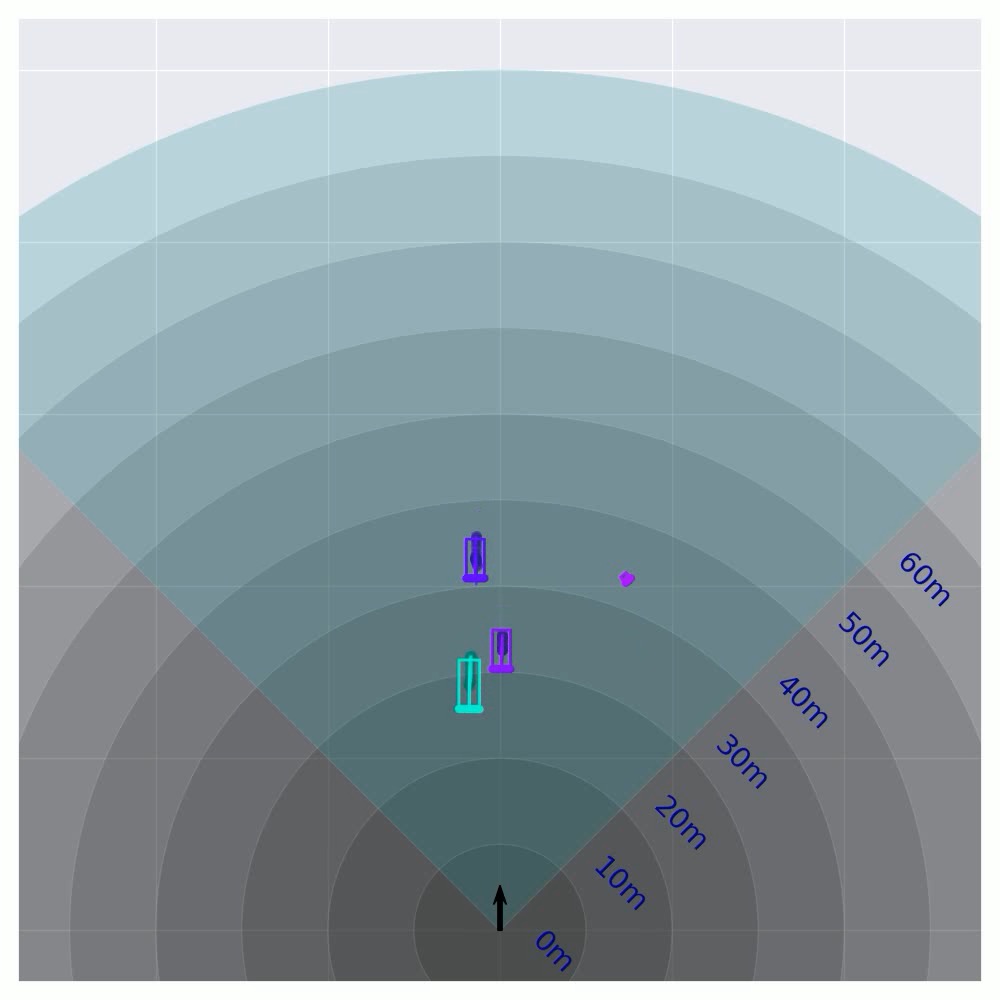}
    \includegraphics[width=1.0\linewidth, keepaspectratio]{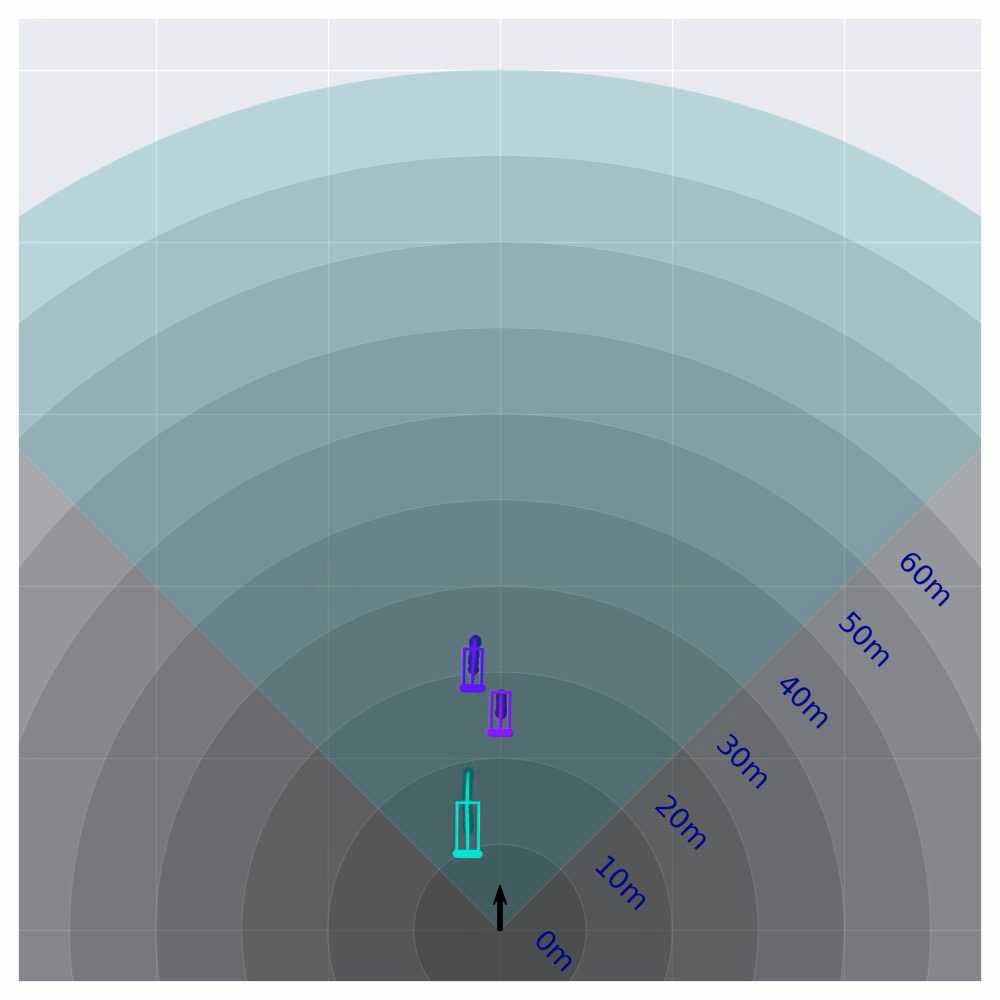}
    \includegraphics[width=1.0\linewidth, keepaspectratio]{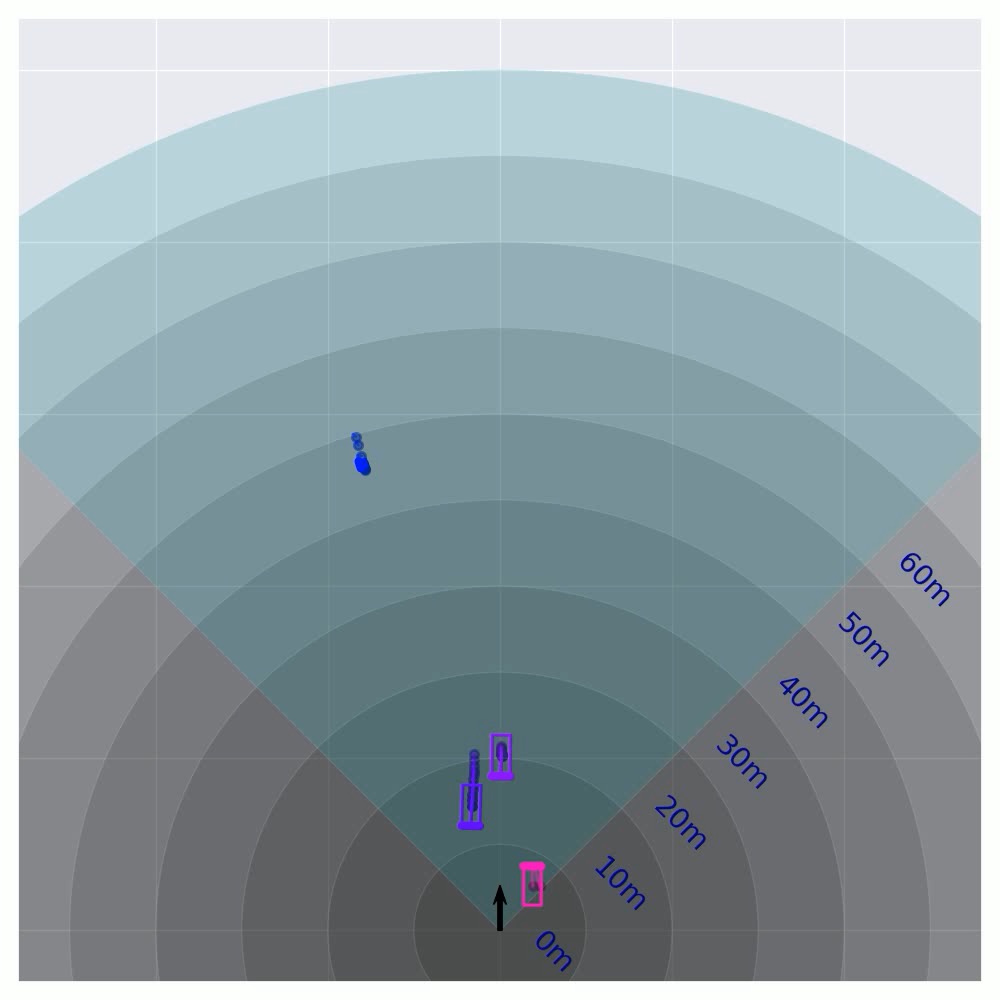}
    \includegraphics[width=1.0\linewidth, keepaspectratio]{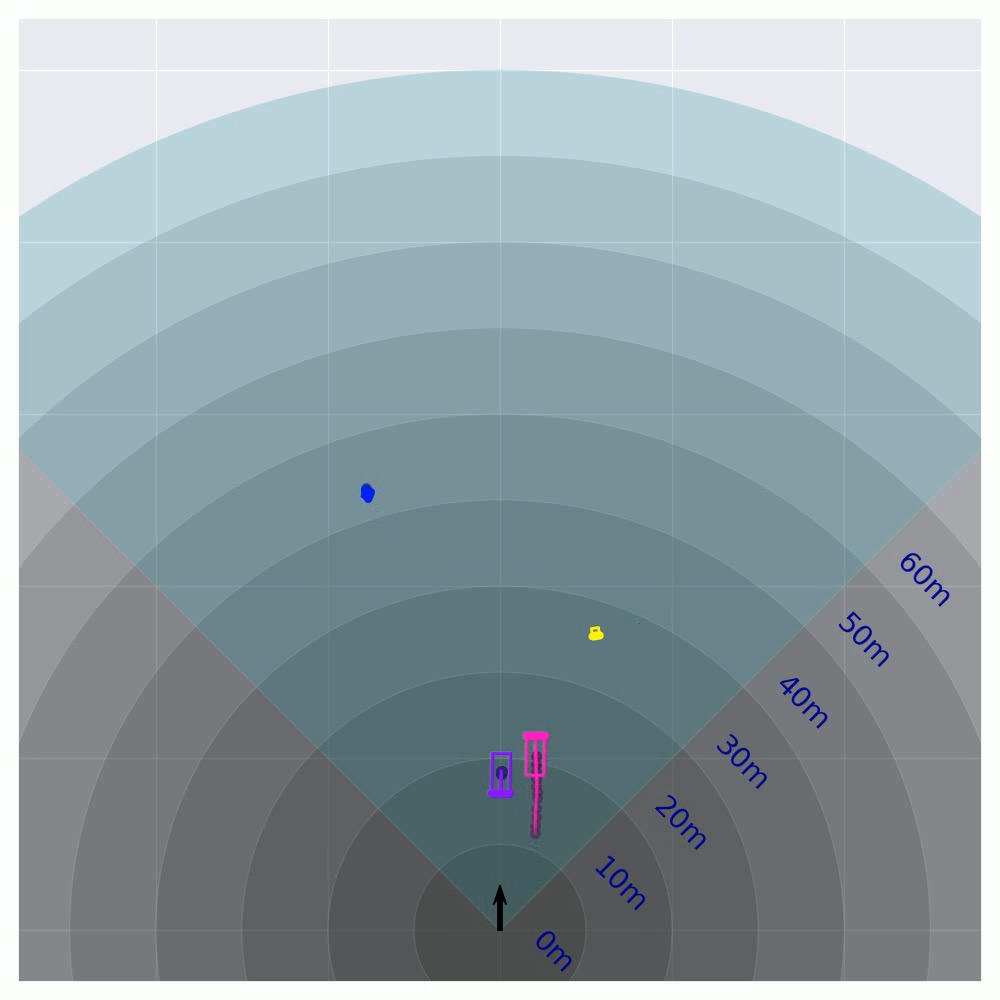}
    \endminipage
    \figcaption{Qualitative results on the testing set of Waymo Open dataset.}{Our QD-3DT accurately tracks all observed objects and locates them in 3D. We show predicted 3D bounding boxes and trajectories colored with tracking IDs. Better visualization with color.}
    \label{fig:qualitative_waymo}
\end{figure*}

\minisection{Qualitative results.}
We show our evaluation results in \figref{fig:qualitative_nusc} on the test set of nuScenes dataset and \figref{fig:qualitative_waymo} on the test set of Waymo Open Dataset.
The trajectories in~\figref{fig:qualitative_nusc} demonstrate our long-term tracking ability in the nuScenes dataset.
The results of Waymo show that our proposed 3D tracking pipeline establishes a strong baseline using monocular images.
The solid rectangular in bird's eye view stands for the predicted vehicle.
The 3D bounding boxes and the rectangular are colored with their tracking ID.
The figures are best viewed in color.

\minisection{Evaluation Video.}
We have uploaded a showcase video that demonstrates video inputs with estimated 3D bounding boxes in the camera view and tracked trajectories in bird's eye view on nuScenes, KITTI, and Waymo Open dataset. Please refer to the showcase video for more qualitative examples.

%% file: 09References.tex
\ifCLASSOPTIONcaptionsoff
  \newpage
\fi



\bibliographystyle{IEEEtran}
\bibliography{egbib}
%

